\documentclass[preprint,12pt,3p]{elsarticle}
\usepackage{amssymb}
\usepackage{amsmath}
\usepackage{relsize}
\usepackage[english]{babel}
\usepackage{hyperref}
\usepackage{enumitem}
\usepackage{graphicx}
\usepackage[ruled,vlined,linesnumbered]{algorithm2e}
\usepackage{blkarray}
\usepackage{array}
\usepackage{subcaption}
\usepackage{amssymb}
\usepackage{pifont}
\usepackage{array}
\usepackage{float}
\newcolumntype{P}[1]{>{\centering\arraybackslash}p{#1}}
\newcommand{\cmark}{\ding{51}}%
\newcommand{\xmark}{\ding{55}}%
\DeclareMathOperator*{\argmin}{\arg\!\min}
\graphicspath{ {./images/} }
\hypersetup{
    colorlinks=true,
    linkcolor=blue,
    filecolor=magenta,
    urlcolor=cyan,
}
\begin{document}
\begin{frontmatter}
\title{Unsupervised Hypergraph Feature Selection via a Novel Point-Weighting Framework and Low-Rank Representation}
\author{Ammar~Gilani\footnote{Ammar.Gilani@aut.ac.ir}}
\author{Maryam~Amirmazlaghani\footnote{Mazlaghani@aut.ac.ir}}
\address{Amirkabir University of Technology, Department of Computer Engineering and
Information Technology}
\begin{abstract}
Feature selection methods are widely used in order to solve the 'curse of dimensionality' problem. Many proposed feature selection frameworks, treat all data points equally; neglecting their different representation power and importance. In this paper, we propose an unsupervised hypergraph feature selection method via a novel point-weighting framework and low-rank representation that captures the importance of different data points. We introduce a novel soft hypergraph with low complexity to model data. Then, we formulate the feature selection as an optimization problem to preserve local relationships and also the global structure of data. Our approach for global structure preservation helps the framework overcome the problem of unavailability of data labels in unsupervised learning. The proposed feature selection method treats with different data points based on their importance in defining the data structure and representation power. Moreover, since the robustness of feature selection methods against noise and outlier is of great importance, we adopt low-rank representation in our model. Also, we provide an efficient algorithm to solve the proposed optimization problem. The computational cost of the proposed algorithm is lower than many state-of-the-art methods which is of high importance in feature selection tasks. We conducted comprehensive experiments with various evaluation methods on different benchmark data sets. These experiments indicate significant improvement, compared with state-of-the-art feature selection methods.
\end{abstract}
\end{frontmatter}

\section{Introduction}\label{introduction}
Nowadays confronting high dimensional data is expected in practical applications. The complexity of many algorithms in various fields like machine learning and pattern recognition highly depends on the dimensionality of data. Two types of dimensionality reduction algorithms can be used to solve the 'curse of dimensionality' problem: feature extraction and feature selection. Feature extraction methods try to define new features based on primary features and redescribe data based on them. In opposite, feature selection techniques aim to select the most descriptive subset of features. Unlike feature extraction techniques, feature selection algorithms preserve the primary representation of the data \cite{RFSTB}.
Feature selection methods are categorized into three groups: supervised, semi-supervised, and unsupervised. Supervised and semi-supervised methods like generalized fisher score (GFS) \cite{GFS}, hypergraph based information-theoretic feature selection (MII\_HG) \cite{MII_HG}, and hypergraph regularized Lasso (HLasso) \cite{HLasso} need labeled data to perform feature selection. Although these methods can be effective, collecting labels for data can be time-consuming or even impractical in many cases. Unlike supervised and semi-supervised methods, unsupervised methods such as feature selection via joint embedding learning and sparse regression (JELSR) \cite{JELSR} and self-representation hypergraph low-rank feature selection (SHLFS) \cite{SHLFS} need no labeled data for feature selection. Hence unsupervised methods are used more widely in many applications.\\
In unsupervised feature selection methods, providing an efficient representation of data is of great importance. Because the structure of data is preserved based on the constructed representation which without having an effective representation, preservation of the defined structure can be ineffective. In the last decade, various algorithms such as JELSR \cite{JELSR}, laplacian score (LapScore) \cite{LapScore}, similarity preserving feature selection (SPFS) \cite{SPFS}, and multi-cluster feature selection (MCFS) \cite{MCFS} used graphs to model data. In graphs each edge can only be connected to at most another edge; So, graph-based methods can be adopted to model only dual relationships among data points. On the other hand, in real applications, different data points can be related to each other as a group and as a result, there exist multiple relationships among them which should be modeled and in the next steps, preserved. To overcome this problem, in the last few years, some algorithms have been using hypergraphs instead to model interconnection among data points; For instance, JHLSR \cite{JHLSR}, MII\_HG \cite{MII_HG}, and SHLFS \cite{SHLFS}. In hypergraphs, hyperedges are used instead of conventional edges. Hyperedges can connect any number of vertices. Although hypergraphs improve ordinary graphs in many ways, they have an important limitation: there are only two states for a vertex and a hyperedge: a hyperedge completely consists of that vertex or not, but the vertex may need to have partial participation in that hyperedge; Because importance of other vertices in that hyperedge can be more or less critical. So, a generalized form of hypergraphs is needed in which amount of attendance of a vertex in a hyperedge can adopt continuous values instead of only two states.\\
Feature selection methods usually encounter the problem of inaccurate measurements and outlier in practical applications. So, it is of great importance for a feature selection method to overcome this problem. Proposing a model without handling noise and outlier can lead to focus on wrong data structure and as a result, inefficient feature selection. One way to overcome this problem is using low-rank representation in hypergraph-based feature selection methods proposed in \cite{HLR_FS, SHLFS}. In other words, hypergraph low-rank feature selection (HLR\_FS) \cite{HLR_FS} and self-representation hypergraph low-rank feature selection (SHLFS) \cite{SHLFS} use low-rank representation in order to deal with noise and outlier.\\
One common disadvantage of the mentioned methods is behaving different data points equally; While different data points can vary much in importance and role in defining the data structure.\\
Another important point in feature selection methods is computational cost. As it was clarified, it is awaited to meet high dimensional data in the task of feature selection. Therefore, the computational complexity of an algorithm is of high importance. An algorithm can be effective on small-sized data, but it may be infeasible to adopt it on a high dimensional practical data set.\\
In this paper, to overcome the weaknesses of the previously proposed methods, we introduce a novel low-rank feature selection method using soft hypergraphs which captures the importance of different data points. In the proposed point-weighting framework, we treat different data points based on their representation power and role in defining the data structure. We propose using soft hypergraphs in order to resolve the limitation of ordinary hypergraphs. Unlike ordinary hypergraphs, in soft hypergraphs, different vertices can have different participation in a hyperedge which leads to more accurate modeling. Moreover, the proposed approach uses cluster centroids to build the hypergraph which results in reducing the computational cost, and also better handling of noise, outlier, and redundancy. The reason for this can be described as follows: Usually, in real situations, some data points can have similar properties. So in order to preserve data structure, there is no need to use all data points. Points with analogous properties can be grouped as a cluster and a representative data point can be used instead of the cluster to represent the properties of that group. In this way, computational complexity decreases and the effect of noise and outlier reduces; Because the small differences in properties of some data points can be only the result of noise; Moreover, redundant information of almost similar data points will be discarded. The best choice for this representative point is the centroid of that cluster; Since it has a good representation power. We use the constructed hypergraph to preserve local relationships between centroids.\\
To preserve the complete structure of data, the global structure needs to be preserved along with local relationships between centroids. For this purpose, we try to maintain the global structure of data by adding a term which also helps resolve the problem of data labels unavailability in unsupervised learning.\\
So, in brief, in this paper, by weighting points, we focus on data points with high representation power; Especially cluster centroids. We also adopt low-rank representation and propose a novel term for global structure preservation. We have tested our proposed framework using various evaluation methods on different benchmark datasets which indicates significant improvement, compared with state-of-the-art feature selection methods.
The innovative aspects of the proposed method can be summarized as follows: i) Introducing a new soft hypergraph to model data for unsupervised feature selection. ii) Proposing a point-weighting framework. iii) Adopting low-rank representation in the point-weighting framework. iv) Introducing a new approach to preserve the global structure of data. v) Providing an efficient algorithm to solve the proposed optimization problem.\\
The notations used in this paper and their definitions are summarized in table \ref{table:notations}. Here the organization of the rest of the paper is explained. In Section \ref{hypergraph-based feature selection}, we review concepts of hypergraph-based feature selection. In Section \ref{proposed method for feature selection}, we explain our proposed framework for feature selection. In section \ref{optimization}, we provide an efficient algorithm to solve the proposed optimization problem. In Section \ref{computational cost and convergence study}, we calculate the computational cost of our proposed algorithm. Section \ref{experiments} exhibits our experimental results. Finally, in Section \ref{conclusion}, we conclude our paper.

\begin{table}
\begin{tabular}{ |P{3cm}|l| }
  \hline
  \multicolumn{2}{|c|}{Important Notations} \\
  \hline
  $m_i^j$ & the element in ith row and jth column of matrix M (a typical matrix)\\
  tr(M) & the trace of M\\
  $||M||_F$ & the Frobenius norm of M\\
  $M^T$ & the transpose of M\\
  $M^{-1}$ & the inverse of M\\
  X $\in$ $R^{n\times d}$ & the data matrix; Each row is a data point\\
  n & the number of data points\\
  m & the number of centroids, which is also the number of vertices and hyperedges\\
  C $\in$ $R^{m\times d}$ & centroids matrix; Each row is a centroid\\
  $\zeta_i$ & the cluster number of the ith data point\\
  H $\in$ $R^{m\times m}$ & the incidence matrix of the soft hypergraph\\
  d(v) & the degree of vertex v\\
  $D_v$ $\in$ $R_+^{m\times m}$ & the diagonal matrix of vertices degrees\\
  $\delta(e)$ & the degree of hyperedge e\\
  $D_e$ $\in$ $R_+^{m\times m}$ & the diagonal matrix of hyperedges degrees\\
  w(e) & the weight function which assigns each hyperedge like e, a non-negative weight\\
  W $\in$ $R_+^{m\times m}$ & the diagonal matrix of hyperedges weights\\
  diag(W) $\in$ $R_+^m$ & the vector of hyperedges weights (the main diagonal of W)\\
  T $\in$ $R^{d\times k}$ & the linear transformation matrix used to embed data\\
  $\Delta $ & the hypergraph laplacian\\
  $I_n$ $\in$ $R^{n\times n}$ & the identity matrix\\
  k & the dimensionality of embedding\\
  d & the initial number of features\\
  \hline
\end{tabular}
\caption{The list of important notations used in this paper and their brief explanation.}
\label{table:notations}
\end{table}

\section{Hypergraph-based feature selection}\label{hypergraph-based feature selection}
An important step in feature selection algorithms is modeling data. Graphs can be used in order to model local relationships between data points. But they can model only dual relationships. To overcome this problem, in recent studies, hypergraphs has been used instead \cite{SHLFS, JHLSR, HLasso}. In this section, first, we explain the fundamentals of hypergraphs. Afterward, principals of soft hypergraphs are explained. Then, we review hypergraph-based feature selection methods quickly. 

\subsection{Hypergraph fundamentals}\label{hypergraph fundamentals}
An ordinary graph can be defined as the triplet: G = (V, E, w); Where V is the set of vertices, E is the set of edges, and w is a function which assigns each edge a weight. An edge consists of at most two vertices; So it can only model dual relationships among vertices. As it was explained earlier, in Section \ref{introduction}, in many applications, there exist multiple relationships among data points. So a conventional graph is not appropriate to thoroughly model relationships among them. To tackle this limitation of conventional graphs, hypergraphs can be used instead. Hypergraphs can be considered as the generalized form of simple graphs. They can be shown using the following notation: $G_H$ = (V, E, w). Similar to graphs, V and w are the set of vertices and the weight function respectively. Although instead of simple edges, hyperedges are adopted in hypergraphs to show interconnection among vertices; They are denoted by E. Hyperedges are the extended mode of traditional edges and can include multiple vertices; As a result, they can model multiple relationships amid vertices. In order to determine the relationship between different vertices and hyperedges, the incidence matrix H is defined as follows:

\begin{equation}\label{eq:H}
    h_v^{e}=
    \begin{cases}
      1, & \text{if}\ vertex~v \in hyperedge~e \\
      0, & \text{elsewhere}
    \end{cases}
\end{equation}

Where $h_v^{e}$ is the element in vth row and eth column of matrix H. Akin to conventional graphs, degree is defined for vertices and hyperedges. Degree of a vertex is the total weights of hyperedges which it is comprised in, i.e.
\begin{equation}\label{eq:dv} d(v)=\sum_{e~=~1}^{|E|} w(e)h_v^{e}\end{equation}
Where $|E|$ is the number of hyperedges. Degree of a hyperedge is the total amount of contributions of different vertices in it which in conventional hypergraphs is equal to the number of vertices it embraces, i.e.
\begin{equation}\label{eq:delta_e} \delta(e)=\sum_{v~=~1}^{|V|} h_v^{e}\end{equation}
Where $|V|$ is the number of vertices, $h_{v}^{e}$ is the element in vth row and eth column of matrix H.

\subsection{Soft hypergraph}
\begin{figure}[ht]
\centering
\begin{subfigure}[b]{0.325\textwidth}
\includegraphics[scale=0.25]{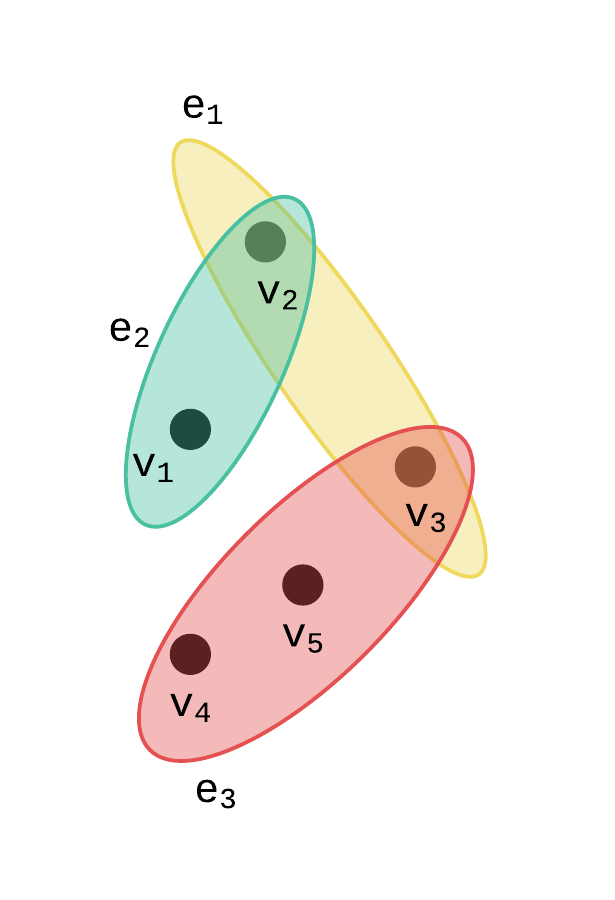}
\caption{Visualization}
\end{subfigure}
\begin{subfigure}[b]{0.325\textwidth}
\[
H=~
\begin{blockarray}{cccc}
e_1 & e_2 & e_3\\
\begin{block}{(ccc)c}
  0 & 0.38 & 0 & v_1\\
  1 & 0.15 & 0 & v_2\\
  0.73 & 0 & 0.18 & v_3\\
  0 & 0 & 0.59 & v_4\\
  0 & 0 & 0.42 & v_5\\
\end{block}
\end{blockarray}
\]
\caption{Incidence matrix}
\end{subfigure}
\caption{An example of soft hypergraph.}
\label{figure:hypergraph example}
\end{figure}

Even though hypergraphs cover the main limitation of ordinary graphs, they have the problem of equal participation of different vertices in their hyperedges. In hyperedges, all vertices are treated equally, but different vertices have different importance and contributions in hyperedges. In addition, there is no need to define the incidence matrix binary, While it can be continuous; At least in our case. This type of hypergraph is called soft hypergraph \cite{IVFH}. An example of soft hypergraph is provided in figure \ref{figure:hypergraph example}; $E=\{e_1, e_2, e_3\}$ is the set of hyperedges and $V=\{v_1, v_2, v_3, v_4, v_5\}$ is the set of vertices. Hyperedge $e_3$ models multiple relationship between vertices $v_3$, $v_4$, and $v_5$. Since the incidence matrix is not binary, these vertices could have different participation in that hyperedge, i.e. 0.18, 0.59, and 0.42 respectively.

\subsection{A review on hypergraph-based feature selection methods}\label{A review on hypergraph-based feature selection methods}
In this section, we review some state-of-the-art feature selection methods which use hypergraphs to perform data modeling. We list related algorithms and provide a brief explanation about them.\\
JHLSR \cite{JHLSR}: This framework adopts sparse representation \cite{RFRSR} to construct its hypergraph. Sparse representation for a data point can be described as the minimal subset of linearly dependent vertices. JHLSR needs one hyperedge for each data point; Moreover, Because of not using soft hypergraph, they have to repeat the process of finding the sparse representation of each data point several times with different sparsification parameters. As a result, the constructed hypergraph becomes very large which increases the computational complexity of the proposed model. As it was explained earlier, encountering noise and outlier is inevitable, but JHLSR does not provide any solution to handle this problem.\\
SHLFS \cite{SHLFS}: Some supervised feature selection methods such as \cite{HLR_FS, ASFS} try to generate the response vector using the linear combination of features and in the last step, report features with the most contribution in generating the response vector, as output. SHLFS modifies this method to be applicable in unsupervised learning and tries to generate each feature, using the linear combination of other features. SHLFS also adopts low-rank representation to handle noise and outlier.\\
HLasso \cite{HLasso}: In order to construct the hypergraph, HLasso connects every data point to its k nearest neighbors which results in a large hypergraph which imposes much computational cost; Moreover, since it adopts ordinary hypergraphs, instead of soft hypergraphs, the nearest neighbors of each data point are treated equally, while they can have different distances from that point. After modeling data, HLasso uses hypergraph laplacian to preserve the distances of each point from its nearest neighbors. HLasso does not handle noise and outlier.\\
HLR\_FS \cite{HLR_FS}: This method constructs an ordinary hypergraph by connecting data points with the same label. HLR\_FS also uses a low-rank constraint to make the model robust against noise and outlier.\\
The above methods overlook the importance of treating different data points based on their representation power and role in defining the data structure. Moreover, none of them adopt soft hypergraphs in order to model data.

\section{Proposed method for feature selection}\label{proposed method for feature selection}
In this paper, we propose an unsupervised hypergraph feature selection method via a novel point-weighting framework and low-rank representation (referred to as HPWL). Since each centroid is the indicator of its cluster and its corresponding points, it has good representational power. So at the first step, we should find the centroids. In this way, any general clustering method can be adopted. We use the k-means algorithm in our experiments. the k-means algorithm is as follows:

\begin{algorithm}[H]
\SetAlgoLined
\textbf{Input:} X, m.\\
\textbf{Output:} $\mu$, $\zeta$.\\
Step 1: Initialization:\\
\Indp
    Initialize cluster centers $\mu_1$, $\mu_2$, ..., $\mu_m$ randomly;\\
\Indm
Step 2: Updating clusters and their centers:\\
\Indp
    \textbf{repeat}\\
    \Indp
        $\zeta_i = \argmin_j{||x_i - \mu_j||^2}$, for i = 1, 2, ..., n\\
        $\mu_j = \frac{\sum_{i=1}^m 1\{\zeta_i=j\}x_i}{\sum_{i=1}^m 1\{\zeta_i=j\}}$, for j = 1, 2, ..., m\\
    \Indm
    \textbf{until} convergence;\\
\Indm
\caption{k-means clustering algorithm.}
\label{algorithm:kmeans}
\end{algorithm}

Where X is the data matrix, $x_i$ is the ith data point, m is the number of clusters, n is the number of data points, $\mu_i$ is the center of the ith cluster, and $\zeta_i$ indicates the cluster number of the ith data point. Using k-means, we determine m clusters. Since in our proposed method each centroid needs to be one of the data points, we consider our ith centroid, i.e. $c_i$ the nearest data point to $\mu_i$; For i = 1, 2, ..., m. We use obtained centroids for hypergraph construction. Treating all data points equally instead of focusing on centroids, results in useless computational complexity and redundant information; Moreover, it makes the algorithm sensitive to noise and outlier which leads to an ineffective selection of features.
\subsection{Soft hypergraph construction}
We construct our soft hypergraph by focusing on the defined centroids. By using soft hypergraph, we can model and then preserve multiple relationships among different centroids precisely. Each centroid is considered as a vertex in the hypergraph. To generate hyperedges, we connect every centroid to its $l$ nearest centroids. To construct the H matrix, suppose that we are generating the hyperedge corresponding to the centroid i, i.e. $h^i$. Its jth element is computed as follows:

\begin{equation}\label{eq:e}
    h_j^i=
    \begin{cases}
      a_j^i, & \text{if}\ c_j \in \text{\{$l$ nearest centroids of } c_i\}\\
      0, & \text{elsewhere}
    \end{cases}
\end{equation}

Where $h_j^i$ is the element in jth row and ith column of matrix H, $a_j^i = {exp}{\Big(-\frac{||c_i - c_j||_{2}^{2}}{\sigma^{2}}\Big)}$; $\sigma$ is average Euclidean distance among centroids and $c_i$, $c_j$ are two centroids. $H~=~[h^1,~h^2,...~,~h^m]$ where $h^i$ is the ith column of H, i.e. the ith hyperedge which is the hyperedge corresponding to the ith centroid. Our algorithm is capable of differentiating the importance of different centroids. We use equal values to initialize the weights of hyperedges. To select the most informative hyperedges and assign a higher weight to them, hyperedges weights should be updated. So the number of hyperedges directly affects the computational complexity. Hence a large number of hyperedges as used in \cite{JHLSR} imposes too much computational complexity. In the proposed method, the number of vertices and the number of hyperedges are equal to the number of centroids which is much lower than the number of data points. So it does not generate too many hyperedges and its computational cost is much lower than other hypergraph-based methods such as \cite{JHLSR, EMHG, HLasso, HSIR}.

\subsection{Data embedding}\label{data embedding}
To perform the task of feature selection, we use a transformation matrix (T) to project data into a low dimensional space. If $x_i$, $x_j$ are close to each other, we want $x_i T$, $x_j T$ to be close to each other too. After finding the optimal T, we use $\ell_{2}$ norm to find primary features which have the most role in setting up this transformation matrix.\\
As it was explained earlier in Section \ref{introduction}, real-world data sets generally are corrupted and include noise and outlier; As a result, the rank of matrices is high in practical applications \cite{SRRDR, MCNE}; Moreover, it has been demonstrated that high-dimensional data sets usually have low-dimensional representations, i.e. they can be redescribed in a low-dimensional space \cite{DGAHNP}. In order to remedy the aforementioned corruption and robust selection of features, we can use low-rank representation. Hence, we suppose that the transformation matrix is low-rank and we consider the following low-rank constraint for it:

\text{rank(T) $\leq$  r}

Since features are finally ranked according to matrix T, in order not to use uninformative and redundant features in data embedding, matrix T also needs to be sparse; Moreover, some elements of T may have small values which can be only the effect of errors. Trying to make matrix T sparse, also helps to set the values of these elements to zero.

\subsection{Local relationships between centroids}\label{local relationships between centroids}
Importance of preservation of local neighborhood relationships among data points has been well discussed in recent works \cite{LapScore, JELSR}. In order to maintain data local relationships, we focus on the centroids and then we try to preserve local neighborhood relationships among them. It has been demonstrated that using Gaussian kernel as similarity function can model local neighborhoods \cite{SpecCl}. So, we try to preserve the Gaussian similarity between different centroids. In other words, we try to maintain the distances between different centroids. To achieve this goal, we need to minimize the following objective function:

\begin{equation}\label{eq:local1}
    \Psi^{(1)}(W,~T)~=~\frac{1}{2}{\mathlarger{\sum}_{c_{i},c_{j}\in V}}\alpha_{i,j}\times||(c_iT)^T~-~(c_jT)^T||_2^2
\end{equation}

where $c_iT$ and $c_jT$ are transformed centroids, $||.||_2$ is $\ell_2$ norm of a vector, and $\alpha_{i,j}$ is calculated using the following formula:

\begin{equation}\label{eq:local2}
    \alpha_{i,j}~=~\mathlarger{\sum}_{e \in E}~\frac{w(e)h(c_i,~e)h(c_j,~e)}{\delta(e)d(c_i)}
\end{equation}

Where $c_i$ is the ith centroid, $d(c_i)$ is its degree in hypergraph, w(e) is the weight of hyperedge e, and $\delta(e)$ is its degree. $\alpha_{i,j}$ determines the importance of closing $c_iT$ and $c_jT$ which is zero in case centroids i and j are not connected to each other; Because h($c_i$, e)h($c_j$, e) = 0. In the other case, namely h($c_i$, e)h($c_j$, e) $>$ 0, the higher value of this factor means these two centroids are closer which leads to a larger $\alpha_{i,j}$; Moreover, $\frac{w(e)}{\delta(e)d(c_i)}$ determines the normalized weight of that hyperedge for $c_i$. So the larger $\alpha_{i,j}$ is, the higher the importance of closing $c_iT$ and $c_jT$ is. As it was explained, $\alpha_{i,j}$ also indicates importance of closing $c_i$ and $c_j$. So minimizing $\Psi^{(1)}$ leads to preservation of distances among different centroids. We have:

\begin{equation}\label{eq:local3}
    \Psi^{(1)}(W,~T)~=~\frac{1}{2}{\mathlarger{\sum}_{c_{i},c_{j}\in V}}\alpha_{i,j}\times||(c_iT)^T~-~(c_jT)^T||_2^2~=~tr(T^TC^T\Delta CT)
\end{equation}

Where C is matrix of centroids and $\Delta$ is the unnormalized hypergraph laplacian matrix defined by \cite{LHC}:

\begin{equation}\label{eq:local4}
    \Delta ~=~I_{|V|}~-~D_v^{-1}HWD_e^{-1}H^T
\end{equation}

$D_v$, $D_e$, and W are diagonal matrices of vertices degrees, hyperedges degrees, and hyperedges weights respectively and $I_{|V|}$ is a $|V| \times |V|$ identity matrix.
It is worth noting that generally, minimizing $tr\big(F(C)^T\Delta F(C)\big)$, where F(C) is a function (in our case CT), leads to closing $f(c_i)$ and $f(c_j)$ where $f(c_i)=F(C)_i$, and $c_i$ and $c_j$ are in at least one common hyperedge; In addition, the amount of effort to close $f(c_i)$ and $f(c_j)$ is determined by contributions of $c_i$ and $c_j$ in their common hyperedge and the normalized hyperedge weight. This analysis clarifies the importance of using soft hypergraphs where different vertices are eligible to have different participation in a hyperedge because elements of the incidence matrix can adopt continuous values.\\
Considering the low-rank constraint for transformation matrix, we have:

\begin{equation}\label{eq:local5}
    \begin{aligned}
        \min_{W,~T}\Psi^{(1)}(W,~T)~=~tr(T^TC^T\Delta CT)\\
        \text{s.t. rank(T) $\leq$ r}
    \end{aligned}
\end{equation}

Where C is the matrix of centroids and considering low-rank constraint is helpful in handling noise and outlier. We also add $\ell2,1$ norm of matrix T as a regularizer to try to make it sparse and as a result, not to use uninformative features in data embedding and reducing the effect of errors:

\begin{equation}\label{eq:W1}
            \Theta^{(1)}(T)~=~||T||_{2, 1}
\end{equation}

Although $||T||_{2, 1}$ is convex, its derivative does not exist when at least one of its rows is zero. Hence we adopt the definition in \cite{JELSR}:

\begin{equation}\label{eq:W2}
            \Theta(T)~=~tr(T^TBT)
\end{equation}

Where B is a $d\times d$ diagonal matrix which its ith element is calculated as follows: 

\begin{equation}\label{eq:B}
            b_i^i~=~\frac{1}{2||t_i||_2}
\end{equation}

Where $t_i$ is the ith row of T. Adding this term, changes the optimization problem to:

\begin{equation}\label{eq:local5_2}
    \begin{aligned}
        \min_{W,~T}\Psi^{(1)}(W,~T)~+~\rho\Theta(T)=~tr(T^TC^T\Delta CT)~+~\rho\times tr(T^TBT)\\
        \text{s.t. rank(T) $\leq$  r}
    \end{aligned}
\end{equation}

Where $\rho$ is a regularization parameter. In the next step, in order to control model complexity and setting the weight of uninformative or redundant hyperedges to zero, two constraints are added to the optimization problem (\ref{eq:local5_2}) which results:

\begin{equation}\label{eq:local6}
    \begin{aligned}
            \min_{W,~T}\Psi(W,~T)~+~\rho\Theta(T)~=~tr(T^TC^T\Delta CT)~+~\rho\times tr(T^TBT)~+~\kappa||diag(W)||_2^2\\
            s.t.~\sum_{i~=~1}^mW_i=1,~W_i\geq0~for~i=0,1,...,m,~\text{rank(T) $\leq$ r}
        \end{aligned}
\end{equation}

Where diag(W) denotes the main diagonal of W, i.e. the vector of hyperedges weights, $W_i = w_i^i$ which is the weight of the ith hyperedge, $\kappa$ is a regularization parameter, and $||.||_2$ is $\ell_{2}$ norm of a vector.

Although the low-rank constraint added in Section \ref{data embedding} helps our proposed framework become robust to noise and outlier, it makes the optimization problem non-convex and NP-hard \cite{CSCTR}. An appropriate way to apply this constraint is to produce T by multiplying two low-rank matrices, i.e. T = PQ where P $\in$ $R^{d\times r}$ and Q $\in$ $R^{r\times k}$:

\begin{equation}\label{eq:local7}
    \begin{aligned}
            \min_{W,~P,~Q}~[\Psi(W,~P,~Q)+\rho\Theta(P,~Q)]~=~tr(Q^TP^TC^T\Delta CPQ)~+~\rho\times tr(Q^TP^TBPQ)~+\\
            \kappa||diag(W)||_2^2~~~s.t.~\sum_{i~=~1}^mW_i=1,~W_i\geq0~for~i=0,1,...,m
        \end{aligned}
\end{equation}

\subsection{Global structure}
As it was discussed, our proposed framework preserves local relationships between centroids by maintaining the distances among them. But in order to conserve the complete structure of data, the global structure of data should be preserved along with local relationships. Since data labels have a significant role in defining the structure of data, many supervised methods such as \cite{mRMR, PattC, MII_HG, GFS} focus on them to preserve data structure. But in unsupervised learning, data labels are not available. This motivates us to introduce an alternative way of preserving global structure in our unsupervised framework. It has been demonstrated that data points within the same class have high linear dependency and correlation \cite{RFRSR}. So, trying to preserve the correlation between data points, can be used as an alternative to focusing on data labels for unsupervised learning. Hence, in the proposed framework, we try to maintain the correlation between data points for global structure preservation; Especially data points with more representation power and role in defining the data structure. Since hypergraph construction and using it have high computational cost, we proposed using cluster centroids in Section \ref{local relationships between centroids}. However, using only centroids leads to inevitable loss of some information. In order to use all the information with minor computational cost, we use all data points in the equation (\ref{eq:global1}).
In this way, our goal can be minimizing the following objective function:

\begin{equation}\label{eq:global1}
    \Upsilon^{(1)}(T)~=~||(XT)(XT)^T~-~XX^T||_F^2
\end{equation}

Where XT is the transformed data and $||.||_F$ is the Frobenius norm of a matrix. In order to normalize the above correlation matrices, we divide each matrix by the length of its generating vectors, i.e. k and d respectively:

\begin{equation}\label{eq:global2}
    \Upsilon^{(2)}(T)~=~||\frac{XT(XT)^T}{k}~-~\frac{XX^T}{d}||_F^2
\end{equation}

Which can be reformulated as follows:

\begin{equation}\label{eq:global3}
    \Upsilon^{(3)}(T)~=~||XT(XT)^T~-~\frac{k}{d}~\times~XX^T||_F^2
\end{equation}

Where k is the dimensionality of embedding and d is the initial number of features.
As in local relationships preservation, more representative data points should be emphasized on, especially the centroids. As a data point gets further from its corresponding centroid, not only its importance and representation power reduces, but it gets more likely an outlier. So it is not of high importance to preserve correlation for these data points. For this purpose, we define diagonal matrix D which its ith element is calculated as follows:

\begin{equation}\label{eq:D}
    d_i^i=
    \begin{cases}
      1, & \text{if}\ x_i \text{ is one of the centroids}\\
      affinity(x_i, c_{\zeta_i}), & \text{elsewhere}
    \end{cases}
\end{equation}

Where $c_{\zeta_i}$ is the centroid of the cluster which $x_i$ belongs to. So, $affinity(x_i, c_{\zeta_i})$ is the Gaussian similarity between the data point i and its corresponding centroid. Note that $affinity(x_i, c_{\zeta_i})~\leq~1$. The further a data point gets from its corresponding centroid, the smaller its corresponding value in D will be. To focus on data points with better representation power and neglect outliers we change the equation (\ref{eq:global3}) to:

\begin{equation}\label{eq:global23}
    \Upsilon^{(4)}(T)~=~||D^\frac{1}{2}[XT(XT)^T~-~\frac{k}{d}~\times~XX^T]D^\frac{1}{2}||_F^2
\end{equation}

We are actually multiplying the changes of correlations between different data points by their importance, prior to calculating the Frobenius norm which is equivalent to emphasizing on more representative points, especially the centroids. The problem is that the defined term is not convex with respect to T. So we need to modify its representation:

\begin{equation}\label{eq:global4}
    \begin{aligned}
    \Upsilon^{(4)}(T)~=~||D^\frac{1}{2}[XT(XT)^T~-~\frac{k}{d}~\times~XX^T]D^\frac{1}{2}||_F^2\\
        =~||D^\frac{1}{2}XT(XT)^TD^\frac{1}{2}~-~\frac{k}{d}~\times~D^\frac{1}{2}XX^TD^\frac{1}{2}||_F^2\\
        =~||D^\frac{1}{2}XTT^TX^TD^\frac{1}{2}~-~\frac{k}{d}~\times~D^\frac{1}{2}XX^TD^\frac{1}{2}||_F^2
    \end{aligned}
\end{equation}

And in the last step, to make it convex, we use an approximation provided in \cite{SPFS} and we have:

\begin{equation}\label{eq:global5}
    \Upsilon(T)~=~||D^\frac{1}{2}XT~-~Z_k||_F^2
\end{equation}

Where $Z_k = \Gamma_k\Xi_k^{1/2}$, $\Xi_k$ is the diagonal matrix of k eigenvalues of $[\frac{k}{d}~\times~D^\frac{1}{2}XX^TD^\frac{1}{2}]$ sorted descendingly, and $\Gamma_k$ is the matrix of their corresponding eigenvectors put together column-wise.
So, minimizing $\Upsilon(T)$ results in preserving the correlation between data points.

\subsection{Final optimization problem}
To find appropriate solutions for W and T, we should minimize $\Psi(W, T)$, $\Theta(T)$, and $\Upsilon(T)$ simultaneously, and T needs to be replaced with PQ. In this way, our optimization problem becomes as follows:

\begin{equation}\label{eq:main_final}
    \begin{aligned}
            \min_{W,~P,~Q}~[\Psi(W,~P,~Q)+\tau\Upsilon(P,~Q)+\rho\Theta(P,~Q)]~=~tr(Q^TP^TC^T\Delta CPQ)~+~\tau||D^\frac{1}{2}XPQ~-~Z_k||_F^2\\
            +~\kappa||diag(W)||_2^2~+~\rho\times tr(Q^TP^TBPQ)~~~s.t.~\sum_{i~=~1}^mW_i=1,~W_i\geq0~for~i=0,1,...,m
        \end{aligned}
\end{equation}

The optimization problem (\ref{eq:main_final}) is the final optimization problem we aim to solve.

\section{Optimization}\label{optimization}
Our proposed objective function consists of three factors: W, P, and Q. Even though our optimization problem is generally non-convex and complicated, if we fix two factors and consider the other factor a variable, the problem becomes much simpler. So, we use coordinate descent algorithm to solve it which is summarized in algorithm \ref{algorithm:main}. Each outer iteration includes three steps as follows:
\begin{enumerate}
  \item Fix Q, W, optimize P.
  \item Fix P, W, find the optimal value of Q.
  \item Solve for W by fixing P and Q.
\end{enumerate}

\begin{algorithm}[H]
\SetAlgoLined
\textbf{Input:} X and parameters $\tau$, $\rho$, $\kappa$, and r.\\
\textbf{Output:} features scores.\\
Step 1: Hypergraph construction.\\
\Indp
    1.1 cluster X using k-means algorithm;\\
    1.2 generate matrix H based on the equation (\ref{eq:e});\\
\Indm
Step 2: Updating P, Q:\\
\Indp
    \textbf{repeat}\\
    \Indp
    2.1 Update P, Q using the equations (\ref{eq:opt_p3}), (\ref{eq:opt_q3}) respectively;\\
    2.2 Update B according to the equation (\ref{eq:B});\\
    \Indm
    \textbf{until} convergence;\\
\Indm
Step 3: Updating W:\\
\Indp
    Update W using the coordinate descent algorithm explained in Section \ref{updating w};\\
\Indm
Step 4: Updating $\Delta $:\\
\Indp
    Update $\Delta $ based on the equation (\ref{eq:local4});\\
\Indm
Step 5: Checking convergence and calculating features scores if necessary:\\
\Indp
    If P, Q have converged, calculate features scores using $\ell2$ norm of T rows, otherwise go to Step 2;\\
\Indm
\caption{novel hypergraph point-weighting framework via low-rank representation (HPWL)}
\label{algorithm:main}
\end{algorithm}

\subsection{Updating P}

By considering Q and W as constants, we need to solve the following optimization problem:

\begin{equation}\label{eq:opt_p1}
    \begin{aligned}
            \min_P~tr(Q^TP^TC^T\Delta CPQ)~+~\tau||D^\frac{1}{2}XPQ~-~Z_k||_F^2\\
            +~\rho\times tr(Q^TP^TBPQ)
    \end{aligned}
\end{equation}

Since the objective function in (\ref{eq:opt_p1}) is convex with respect to P, it suffices to take its derivative with respect to P and set it to zero:

\begin{equation}\label{eq:opt_p2}
    \begin{aligned}
            C^T(\Delta + \Delta^T)CPQQ^T~+~2\tau X^TD^\frac{1}{2}(D^\frac{1}{2}XPQ-Z_k)Q^T+~2\rho BPQQ^T~=~0
    \end{aligned}
\end{equation}

Which results:

\begin{equation}\label{eq:opt_p3}
    \begin{aligned}
            P~=~\tau[C^T\Delta^\prime C~+~\tau X^TDX~+~\rho B]^{-1}X^TD^\frac{1}{2}Z_kQ^{-1}
    \end{aligned}
\end{equation}

Where $\Delta^\prime = (\Delta + \Delta^T) / 2$.

\subsection{Updating Q}

Similarly, we need to solve the following optimization problem to update Q:

\begin{equation}\label{eq:opt_q1}
    \begin{aligned}
            \min_Q~tr(Q^TP^TC^T\Delta CPQ)~+~\tau||D^\frac{1}{2}XPQ~-~Z_k||_F^2\\
            +~\rho\times tr(Q^TP^TBPQ)
    \end{aligned}
\end{equation}

Setting its derivative with respect to Q, to zero results:

\begin{equation}\label{eq:opt_q2}
    \begin{aligned}
            P^TC^T(\Delta + \Delta^T)CPQ~+~2\tau P^TX^TD^\frac{1}{2}(D^\frac{1}{2}XPQ-Z_k)+~2\rho P^TBPQ~=~0
    \end{aligned}
\end{equation}

Which leads to:

\begin{equation}\label{eq:opt_q3}
    \begin{aligned}
            Q~=~\tau P^{-1}[C^T\Delta^\prime C~+~\tau X^TDX~+~\rho B]^{-1}X^TD^\frac{1}{2}Z_k
    \end{aligned}
\end{equation}

Where $\Delta^\prime = (\Delta + \Delta^T) / 2$.

\subsection{Updating W}\label{updating w}

Since P and Q are fixed, in order to update W, we need to solve the following optimization problem:

\begin{equation}\label{eq:opt_w1}
    \begin{aligned}
        \min_W~tr(Q^TP^TC^T\Delta CPQ)~+~\kappa||diag(W)||_2^2~~~s.t.~\sum_{i~=~1}^mW_i=1,~W_i\geq0~for~i=0,1,...,m
    \end{aligned}
\end{equation}

Based on the equation (\ref{eq:local4}), $\Delta ~=~I_{|V|}~-~D_v^{-1}HWD_e^{-1}H^T$. The optimization problem (\ref{eq:opt_w1}) can be simplified to:

\begin{equation}\label{eq:opt_w2}
    \begin{aligned}
        \min_W~tr(Q^TP^TC^TD_v^{-1}HWD_e^{-1}H^TCPQ)~+~\kappa||diag(W)||_2^2\\
        s.t.~\sum_{i~=~1}^mW_i=1,~W_i\geq0~for~i=0,1,...,m    
    \end{aligned}
\end{equation}

Let $R=Q^TP^TC^TD_v^{-1}H$, $S=D_e^{-1}H^TCPQ$. for simplification. We have:

\begin{equation}\label{eq:opt_w3}
    \begin{aligned}
        \min_W~tr(RWS)~+~\kappa||diag(W)||_2^2\\
        s.t.~\sum_{i~=~1}^mW_i=1,~W_i\geq0~for~i=0,1,...,m    
    \end{aligned}
\end{equation}

We have the following equalities:

\small
\begin{equation}\label{eq:opt_w4}
    \begin{aligned}
        tr(RWS)~+~\kappa||diag(W)||_2^2~=~tr\Bigg(\begin{bmatrix}r^1~r^2~\hdots~r^m\end{bmatrix}\begin{bmatrix}W_1&&\\ &\ddots&\\&&W_m\end{bmatrix}\begin{bmatrix}s_1\\ s_2\\ \vdots\\ s_m\end{bmatrix}\Bigg)\\+~\kappa||diag(W)||_2^2~=~tr\bigg(\sum_{i=1}^m W_i r^i s_i\bigg)+~\kappa||diag(W)||_2^2~=~\sum_{i=1}^m\bigg[\sum_{j=1}^k r_j^i s_i^j W_i\bigg]\\
        +~\kappa||diag(W)||_2^2~=~\sum_{i=1}^m\bigg[\Omega_iW_i~+~\kappa W_i^2\bigg]
    \end{aligned}
\end{equation}
\normalsize

Where $\Omega_i=\sum_{j=1}^k r_j^i s_i^j$, $s_i$ is the ith row of S, and $r^i$ is the ith column of R. So the optimization problem (\ref{eq:opt_w3}) can be simplified to:

\begin{equation}\label{eq:opt_w4.5}
    \begin{aligned}
        \min_W~\sum_{i=1}^m\bigg[\Omega_iW_i~+~\kappa W_i^2\bigg]\\
        s.t.~\sum_{i~=~1}^mW_i=1,~W_i\geq0~for~i=0,1,...,m    
    \end{aligned}
\end{equation}

Where $\Omega_i=\sum_{j=1}^k r_j^i s_i^j$. We use the coordinate descent algorithm to solve the optimization problem (\ref{eq:opt_w4.5}); Because it has the potential to provide a sparse solution, i.e. setting weights of uninformative hyperedges to zero. In each iteration of the coordinate descent algorithm, we select two consecutive hyperedges, like $W_i$ and $W_{i+1}$, to update their weights. Since $\sum_{i~=~1}^mW_i=1$, $(W_i + W_{i+1})$ will not change after each inner iteration. So we have:

\begin{equation}\label{eq:opt_w4.6}
\begin{aligned}
      W_i^*, W_{i+1}^*~=~\argmin_{W_i, W_{i+1}}{\bigg[\Omega_iW_i~+~\Omega_{i+1}W_{i+1}~+~\kappa (W_i^2+W_{i+1}^2)\bigg]}\\
      s.t.~W_i, W_{i+1} \geq 0,~~W_i + W_{i+1} = c
\end{aligned}
\end{equation}

Where $W_i^*$, $W_{i+1}^*$ are the optimal values of $W_i$, $W_{i+1}$ respectively. Replacing $W_{i+1}$ with $(c-W_i)$ results in:

\begin{equation}\label{eq:opt_w4.7}
\begin{aligned}
      W_i^*~=~\argmin_{W_i}{\bigg[\Omega_iW_i~+~\Omega_{i+1}(c-W_i)~+~\kappa (W_i^2+(c-W_i)^2)\bigg]}\\
      s.t.~0 \leq W_i \leq c
\end{aligned}
\end{equation}

Which can be simplified to:

\begin{equation}\label{eq:opt_w4.8}
\begin{aligned}
      W_i^*~=~\argmin_{W_i}{\bigg[2\kappa W_i^2 + (\Omega_i-\Omega_{i+1}-2\kappa c)W_i\bigg]}\\
      s.t.~0 \leq W_i \leq c
\end{aligned}
\end{equation}

Since the objective function in (\ref{eq:opt_w4.8}) is convex, we have the following equations:

\begin{equation}\label{eq:opt_w5}
    \begin{cases}
      W_i^* = 0,~W_{i+1}^* = c & if~2\kappa c \leq (\Omega_i - \Omega_{i + 1})\\
      W_{i+1}^* = 0,~W_i^* = c & if~2\kappa c \leq (\Omega_{i + 1} - \Omega_i)\\
      W_i^* = \frac{-\Omega_i+\Omega_{i+1}+2\kappa c}{4\kappa},~W_{i+1}^*=c - W_i^* & \text{elsewhere}
    \end{cases}
\end{equation}

Where $W_i^*$, $W_{i+1}^*$ are the updated values of $W_i$, $W_{i+1}$ respectively, $\Omega_i=\sum_{j=1}^k r_j^i s_i^j$, and $c = (W_i + W_{i+1})$. Using this method, the value of the objective function in (\ref{eq:main_final}) reduces after each inner iteration of updating W. As it can be seen in the equation (\ref{eq:opt_w5}), the provided solution by the coordinate descent algorithm is potentially sparse. After finding the optimal values of P and Q, matrix T is constructed using T=PQ. Each row of T corresponds to a feature and score of each feature is calculated based on $\ell2$ norm of its corresponding row in T. Finally, top-ranked features are selected.

\section{Computational cost and convergence study}\label{computational cost and convergence study}
\subsection{Computational cost}\label{computational cost}
The first step of algorithm \ref{algorithm:main} is hypergraph construction. To construct the hypergraph, data points should be clustered using the k-means method. its computational cost is $\mathcal{O}(t_0mn)$; Where $t_0\ll n$, $m\ll n$, and n are the number of iterations, clusters, and data points respectively. In step 2, matrices P and Q need to be updated. In Section \ref{local relationships between centroids}, because only centroids participated in the hypergraph, the incidence matrix H was defined $\in R^{m \times m}$. But, if needed, H can be defined $\in R^{n \times m}$ by adding zero valued rows for non-centroid data points, without any change in the hypergraph. Updating P, Q requires calculating the inverse of a $d\times d$ matrix where d is the number of features. Since in feature selection problems, usually the number of data points is much fewer than the number of features, based on \cite{l21RFS}, by defining matrix H $\in R^{n \times m}$, the equations (\ref{eq:opt_p3}), (\ref{eq:opt_q3}) can be reformulated as follows:

\begin{equation}\label{eq:opt_p_f}
    P~=~\frac{\tau}{\rho}B^{-1}X^T[\frac{1}{\rho}(\Delta^\prime +\tau D)XB^{-1}X^T+I_n]^{-1}D^\frac{1}{2}Z_kQ^{-1}
\end{equation}

\begin{equation}\label{eq:opt_q_f}
    Q~=~\frac{\tau}{\rho} P^{-1}B^{-1}X^T[\frac{1}{\rho}(\Delta^\prime +\tau D)XBX^T + I_n]^{-1}D^\frac{1}{2}Z_k
\end{equation}

By this approach the inverses of $n\times n$ matrices can be calculated instead which yields the time complexity of $\mathcal{O}(min(d,n)^3)$ for the second step. The third step consists of m inner iterations and computational cost of each iteration is k; Where k is the dimensionality of embedding. As a result, the time complexity of this step is $\mathcal{O}(mk)$. Since in equation (\ref{eq:local4}), after each outer iteration, the values of H and $D_e$ does not change, there is no need to compute $\Delta $ from the ground up. So, the computational complexity of this step is $m^2\log m$. Since $m\ll n$, in almost all cases its lower than step 2. So The computational cost of each step of algorithm \ref{algorithm:main} is $max\{\mathcal{O}(min(d,n)^3), \mathcal{O}(mk), \mathcal{O}(t_0mn)\}$,~~$m,t_0\ll n$; Which is less than several state-of-the-art algorithms such as \cite{JHLSR, LapScore, JELSR, MCFS} in most cases.

\subsection{Convergence study}\label{convergence study}
In Section \ref{updating w}, it was proved that the value of the objective function decreases after each inner iteration. The proposed objective function has lower bound of zero; Moreover, we have the following proposition:

\textbf{Proposition 1.} \textit{By using coordinate descent method, value of the objective function in (\ref{eq:main_final}) reduces after each iteration of step 2 of algorithm \ref{algorithm:main}.}

\textit{Proof:} let

\begin{equation}\label{eq:convergence1}
    P^{(i)}, Q^{(i)} =\argmin_{P, Q}{\Psi(P, Q, W) + \Upsilon(P, Q) + \rho \times tr(Q^TP^TB^{(i-1)}PQ)}
\end{equation}

Where $P^{(i)}, Q^{(i)}$ are the values of P, Q after the ith iteration and $B^{(i-1)}$ is the value of B after iteration (i-1). As a result:

\begin{equation}\label{eq:convergence2}
\begin{aligned}
    \Psi(P^{(i)}, Q^{(i)}, W) + \Upsilon(P^{(i)}, Q^{(i)}) + \rho \times tr((Q^{(i)})^T(P^{(i)})^TB^{(i-1)}P^{(i)}Q^{(i)}) \leq\\
    \Psi(P^{(i-1)}, Q^{(i-1)}, W) + \Upsilon(P^{(i-1)}, Q^{(i-1)}) + \rho \times tr((Q^{(i-1)})^T(P^{(i-1)})^TB^{(i-1)}P^{(i-1)}Q^{(i-1)})
\end{aligned}
\end{equation}

Since based on the definition in (\ref{eq:B}), $b_i^i~=~\frac{1}{2||t_i||_2}$, equation (\ref{eq:convergence2}) can be rewritten as follows:

\begin{equation}\label{eq:convergence3}
\begin{aligned}
    \Psi(P^{(i)}, Q^{(i)}, W) + \Upsilon(P^{(i)}, Q^{(i)}) + \rho \times \sum_{j=1}^d\frac{||t_j^{(i)}||_2^2}{||t_j^{(i-1)}||_2} \leq\\
    \Psi(P^{(i-1)}, Q^{(i-1)}, W) + \Upsilon(P^{(i-1)}, Q^{(i-1)}) + \rho \times \sum_{j=1}^d\frac{||t_j^{(i-1)}||_2^2}{||t_j^{(i-1)}||_2}
\end{aligned}
\end{equation}

Based on a lemma provided in \cite{l21RFS}, we have:

\begin{equation}\label{eq:convergence4}
\begin{aligned}
    \sum_{j=1}^d ||t_j^i||_2 - \sum_{j=1}^d \frac{||t_j^i||_2^2}{||t_j^{(i-1)}||_2} \leq \sum_{j=1}^d ||t_j^{(i-1)}||_2 - \sum_{j=1}^d \frac{||t_j^{(i-1)}||_2^2}{||t_j^{(i-1)}||_2}
\end{aligned}
\end{equation}

According to the equations (\ref{eq:convergence3}), (\ref{eq:convergence4}), the following result holds:

\begin{equation}\label{eq:convergence5}
\begin{aligned}
    \Psi(P^{(i)}, Q^{(i)}, W) + \Upsilon(P^{(i)}, Q^{(i)}) + \rho \times \sum_{j=1}^d ||t_j^{(i)}||_2^2 \leq\\
    \Psi(P^{(i-1)}, Q^{(i-1)}, W) + \Upsilon(P^{(i-1)}, Q^{(i-1)}) + \rho \times \sum_{j=1}^d ||t_j^{(i-1)}||_2^2
\end{aligned}
\end{equation}

Which is equivalent to:

\begin{equation}\label{eq:convergence_final}
\begin{aligned}
    \Psi(P^{(i)}, Q^{(i)}, W) + \Upsilon(P^{(i)}, Q^{(i)}) + \rho \times \Theta(P^{(i)}, Q^{(i)}) \leq\\
    \Psi(P^{(i-1)}, Q^{(i-1)}, W) + \Upsilon(P^{(i-1)}, Q^{(i-1)}) + \rho \times \Theta(P^{(i-1)}, Q^{(i-1)})
\end{aligned}
\end{equation}

This completes the proof. Algorithm \ref{algorithm:main} generally converges in two iterations; See figures \ref{figure:convergence_T}, \ref{figure:convergence_obj}. This number of iterations is fewer than several other feature selection methods such as \cite{JHLSR, JELSR}.

\subsection{Comparison with other methods}\label{comparison with other methods}
The computational cost of some state-of-the-art feature selection methods which mentioned their computational complexity, along with the number of outer iterations which they need to converge, is provided in table \ref{table:computational_cost}. As it can be seen, focusing on centroids helps our proposed method need a fewer number of iterations to converge and have lower computational complexity in most cases.

\begin{table}
\centering
\begin{tabular}{ |P{1.5cm}|P{10cm}|P{4cm}| }
  \hline
  method & computational cost & number of iterations\\
  \hline
  JHLSR & $max\{\mathcal{O}(min(d,n)^3),~\mathcal{O}(mk),~\mathcal{O}(r^3+n^2)\}$, $m \gg n$ & 4\\
  LapScore & $\mathcal{O}(n^2d)$ & -\\
  JELSR & $max\{\mathcal{O}(n^2d), \mathcal{O}(n^3), \mathcal{O}(knd)\}$ & 10\\
  MCFS & $max\{\mathcal{O}(n^2d), \mathcal{O}(n^3), \mathcal{O}(knd)\}$ & -\\
  proposed & $max\{\mathcal{O}(min(d,n)^3), \mathcal{O}(mk), \mathcal{O}(t_0mn)\}$,~~$m,t_0\ll n$ & 2\\
  \hline
\end{tabular}
\caption{Computational cost of state-of-the-art methods beside number of outer iterations each one needs to converge.}
\label{table:computational_cost}
\end{table}

\subsection{Parameter determination}\label{parameter determination}
Our proposed objective function includes three parameters that need to be optimized manually: $\tau$, $\kappa$, and $\rho$. $\kappa$ and $\rho$ are regularization parameters which adjust sparsity of W and T. Parameter $\tau$ is used to set the importance of global structure preservation versus maintaining local neighborhood relationships among different centroids. We tuned these parameters empirically by grid search. In our experiments, we used $m = \left \lfloor{\frac{n}{10}}\right \rfloor$ for the number of centroids.

\section{Experiments}\label{experiments}
\subsection{Evaluation methods}\label{evaluation methods}
In order to exhibit the effectiveness of our proposed method, we tested it using classification methods. Since our proposed method does not collaborate with any specific classifier, it can be tested using any classification method. We evaluated the proposed method using four general classification methods: K-Nearest Neighbours (KNN), Support Vector Machine with linear kernel (L-SVM), Support Vector Machine with radial basis kernel (RB-SVM), and Na\"ive Bayes (NB). These classifiers have been used to evaluate feature selection methods widely in recent studies \cite{mRMR, RSR, JELSR, MCFS, trace-ratio, ELMS}.

\subsection{Data sets}\label{data sets}
Six publicly available benchmark data sets were used in our experiments: Gene-Expression \cite{CGAP}, Smoke-Cancer \cite{AEGE}, Various-Cancers \cite{RSCTC}, Burkitt-Lymphoma \cite{RSCTC}, Mouse-Type \cite{RSCTC}, and Hepatitis-C \cite{RSCTC}. The number of classes in these datasets ranges from 2 to 10. These datasets also have a different number of samples: from 187 to 801. Moreover, these data sets are from different areas. The details of these data sets are provided in table \ref{table:data sets}.

\begin{table}
\centering
\begin{tabular}{ |P{4cm}|P{2cm}|P{2cm}|P{1.5cm}| }
  \hline
  Data set & Samples & Features & Classes\\
  \hline
  Gene-Expression & 801 & 20,531 & 2\\
  Smoke-Cancer & 187 & 19,993 & 2\\
  Various-Cancers & 383 & 54,676 & 10\\
  Burkitt-Lymphoma & 220 & 22,284 & 3\\
  Mouse-Type & 214 & 45,102 & 7\\
  Hepatitis-C & 283 & 54,622 & 3\\
  \hline
\end{tabular}
\caption{Details of benchmark data sets used in our experiments.}
\label{table:data sets}
\end{table}

\subsection{Comparison methods}\label{comparison methods}
We compared the accuracy of our proposed method with seven state-of-the-art feature selection frameworks. These methods treat all data points equally and do not focus on more representative points, Also the importance of handling noise, redundancy and outlier is underestimated in them. Here we provide a brief explanation of each one of them:

\begin{description}[font=$\bullet$\scshape\bfseries]
\item JHLSR \cite{JHLSR} Uses sparse representation for hypergraph construction.
\item FScore \cite{PattC} This supervised framework scores the features one by one. This method tries to select the features which have the same value for data points within the same class, and different values for data points from different classes.
\item l21RFS \cite{l21RFS} Tries to preserve similarity between data points and measures regression loss by $\ell_{2, 1}$ norm. This framework is supervised with class labels.
\item TraceRatio \cite{trace-ratio} The main idea of this framework is similar to FScore. But it evaluates features jointly. We used the supervised form of this method in our experiments.
\item LapScore \cite{LapScore} Selects features which can best preserve local structure of data.
\item JELSR \cite{JELSR} Adopts a graph to model data structure. This method performs feature selection by jointly learning embedding and sparse regression.
\item MCFS \cite{MCFS} First carries out manifold learning and then spectral regression.
\end{description}

From the above methods, FScore, l21RFS, and TraceRatio are supervised with class labels and other methods are unsupervised. The differences between the unsupervised methods are provided in table \ref{table:differences}.

\begin{table}
\centering
\scalebox{0.85}{
\begin{tabular}{ |P{2.5cm}|P{2.5cm}|P{2.5cm}|P{2.5cm}|P{2.5cm}|P{2.5cm}|P{2.5cm}| }
  \hline
  Method & Noise and redundancy & Local relationships & Global structure & Weighting points$^a$ & Multiple relationships$^b$ & Joint learning$^c$\\
  \hline
  JHLSR \cite{JHLSR} & \xmark & \cmark & \cmark & \xmark & \cmark & \cmark\\
  LapScore \cite{LapScore} & \xmark & \cmark & \xmark & \xmark & \xmark & \xmark\\
  JELSR \cite{JELSR} & \xmark & \cmark & \xmark & \xmark & \xmark & \cmark\\
  MCFS \cite{MCFS} & \xmark & \cmark & \xmark & \xmark & \xmark & \xmark\\
  proposed & \cmark & \cmark & \cmark & \cmark & \cmark & \cmark\\
  \hline
\end{tabular}
}
\caption{Differences between the proposed method and other state-of-the-art unsupervised feature selection methods.\\
$^a$treating data points differently based on their importance.\\
$^b$adopting hypergraph for data representation.\\
$^c$unifying two learning instead of performing them consecutively.}
\label{table:differences}
\end{table}

\subsection{Experimental settings}\label{Experimental settings}
Since the optimal number of features is unknown, we performed comprehensive experiments with 10, 20, 30, ..., 200 number of features on each benchmark dataset. As in \cite{JHLSR, JELSR} we randomly select 50\% of data as training data and the other 50\% as test data. We repeated this procedure five times and reported average and standard deviation of accuracies. Training data is given to feature selection methods as input. When the selection of features is conducted, reported features of test data are selected and along with corresponding labels are passed to a classification algorithm. Finally, classification accuracy is reported.

\subsection{Experiments Results}\label{experiments results}
\begin{figure}[!htbp]
\centering
\begin{subfigure}[b]{0.325\textwidth}
\includegraphics[scale=0.42]{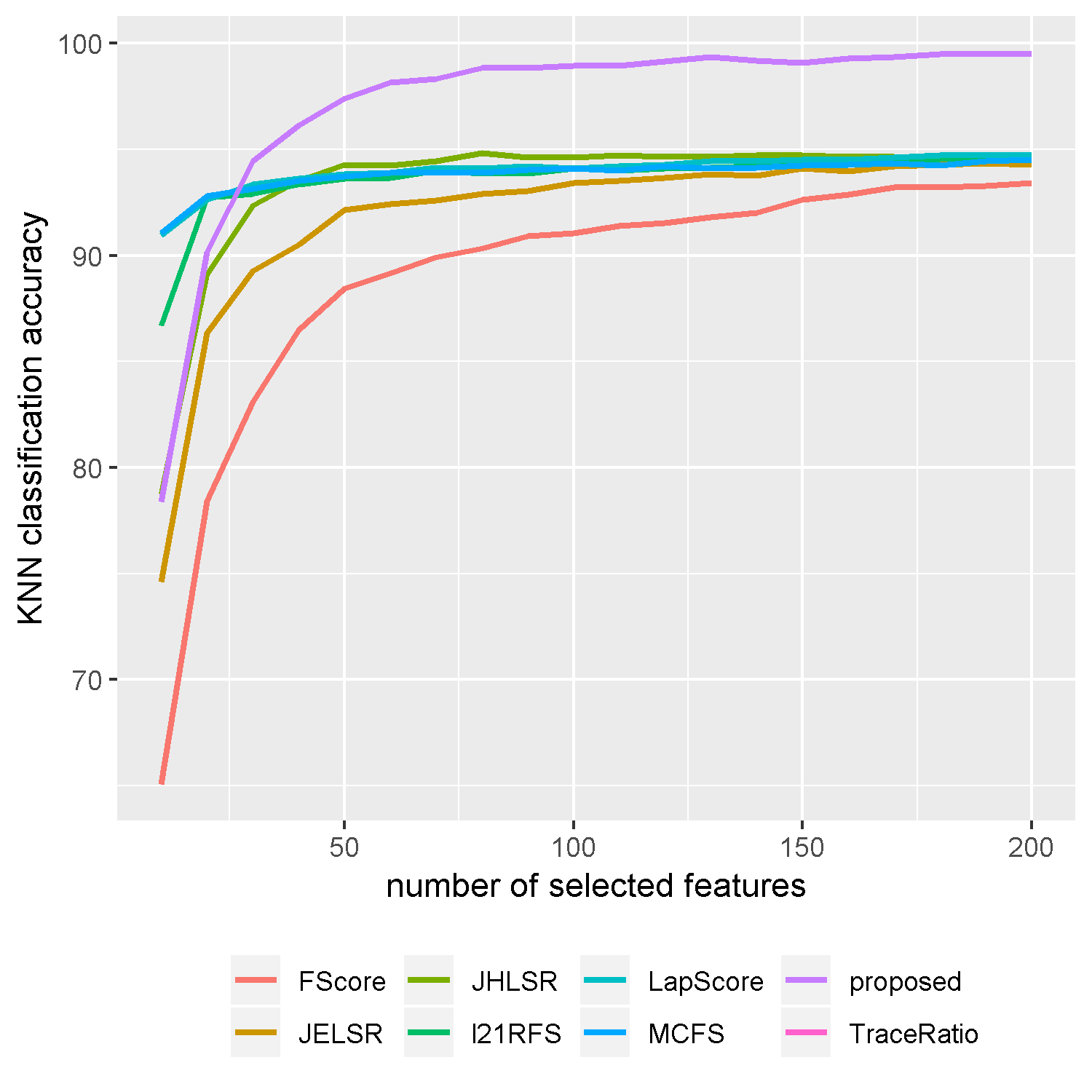}
\caption{Gene-Expression}
\end{subfigure}
\begin{subfigure}[b]{0.325\textwidth}
\includegraphics[scale=0.42]{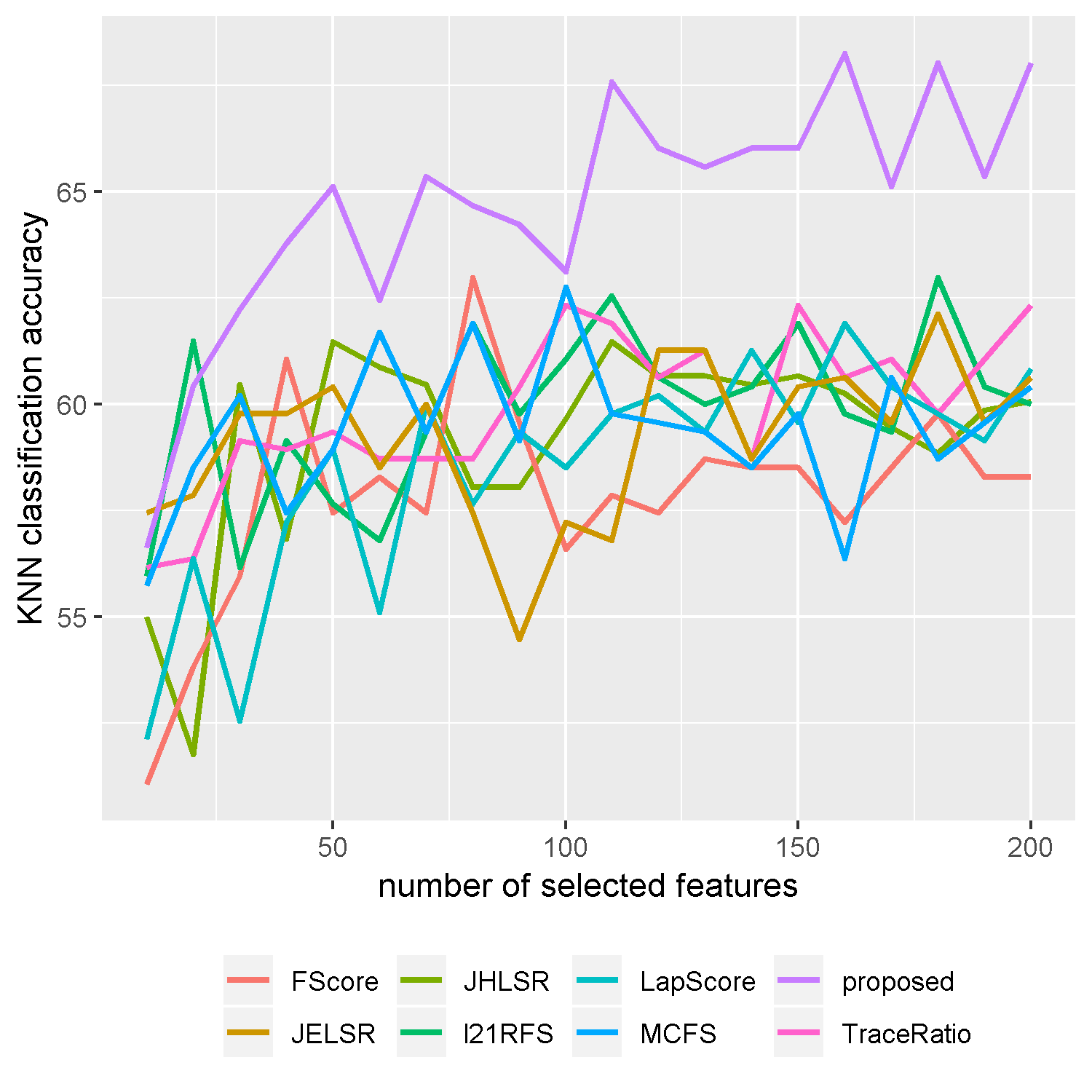}
\caption{Smoke-Cancer}
\end{subfigure}
\begin{subfigure}[b]{0.325\textwidth}
\includegraphics[scale=0.42]{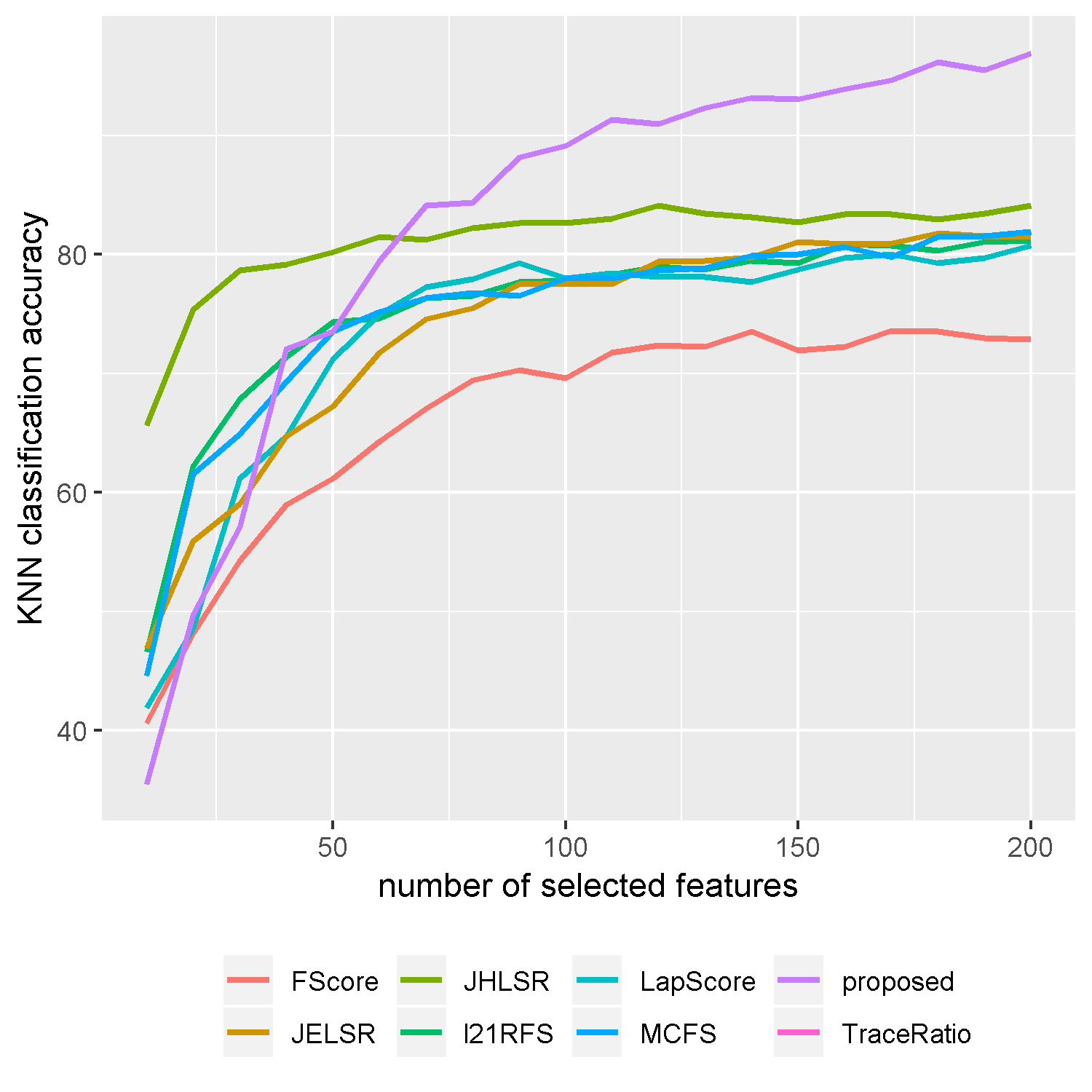}
\caption{Various-Cancers}
\end{subfigure}
\begin{subfigure}[b]{0.325\textwidth}
\includegraphics[scale=0.42]{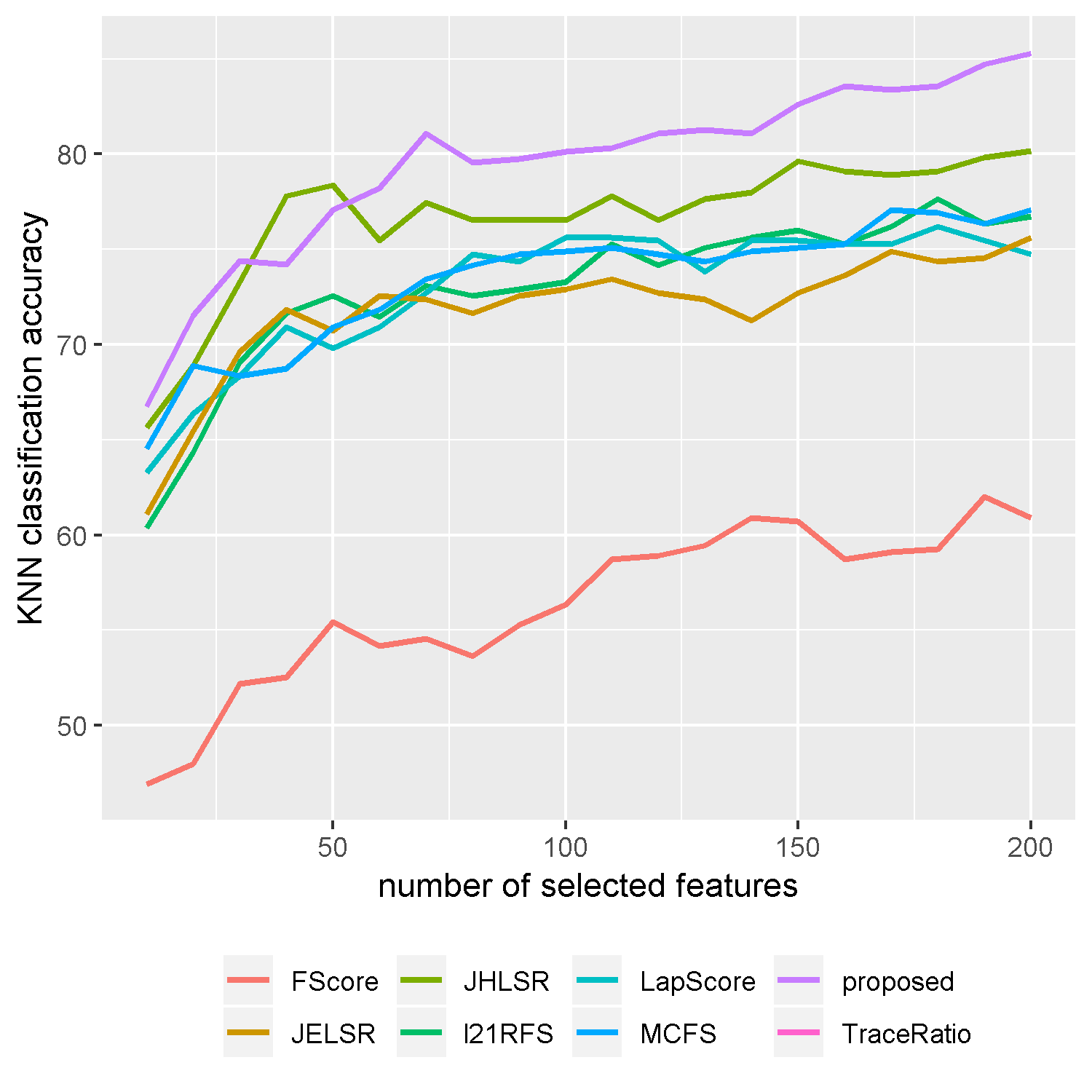}
\caption{Burkitt-Lymphoma}
\end{subfigure}
\begin{subfigure}[b]{0.325\textwidth}
\includegraphics[scale=0.42]{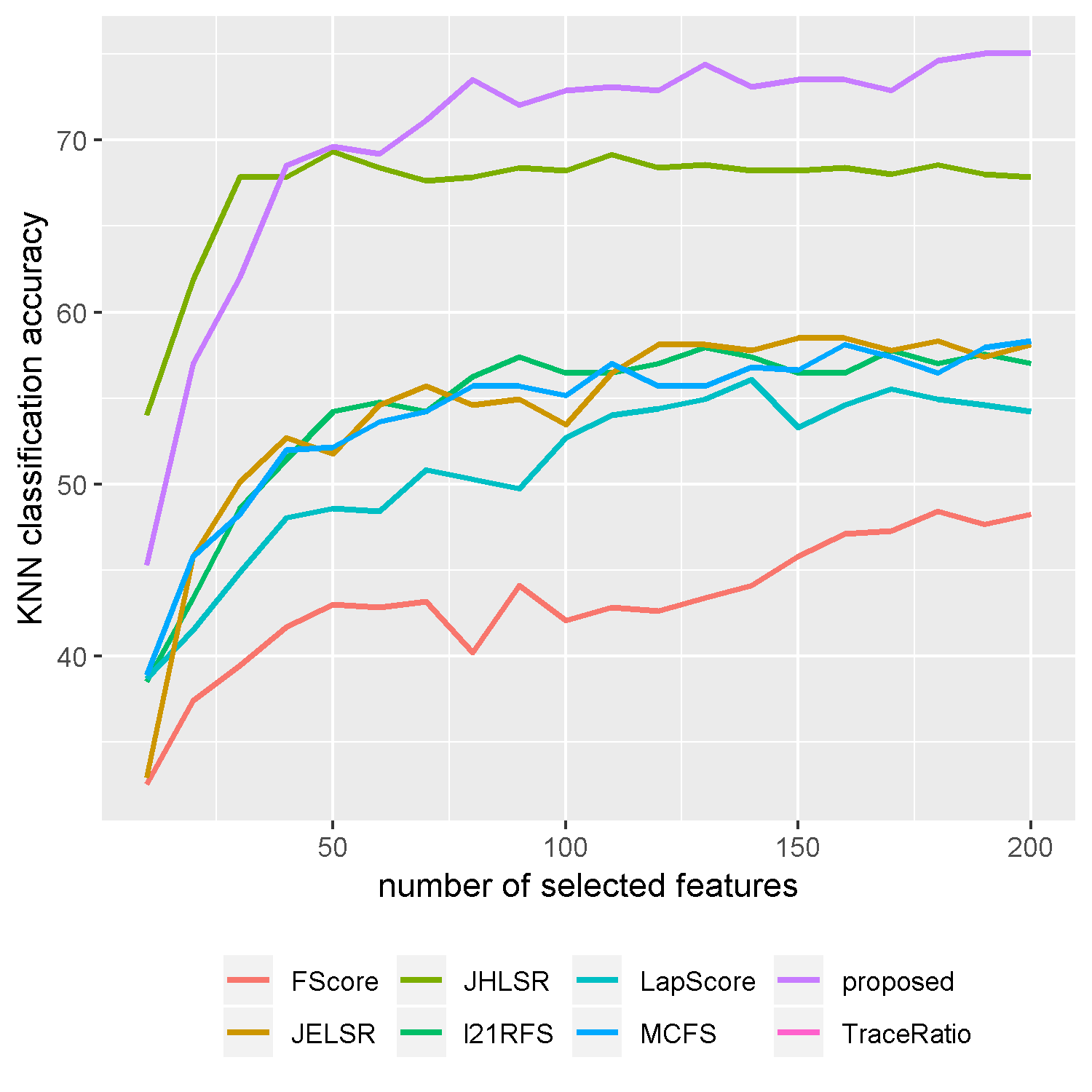}
\caption{Mouse-Type}
\end{subfigure}
\begin{subfigure}[b]{0.325\textwidth}
\includegraphics[scale=0.42]{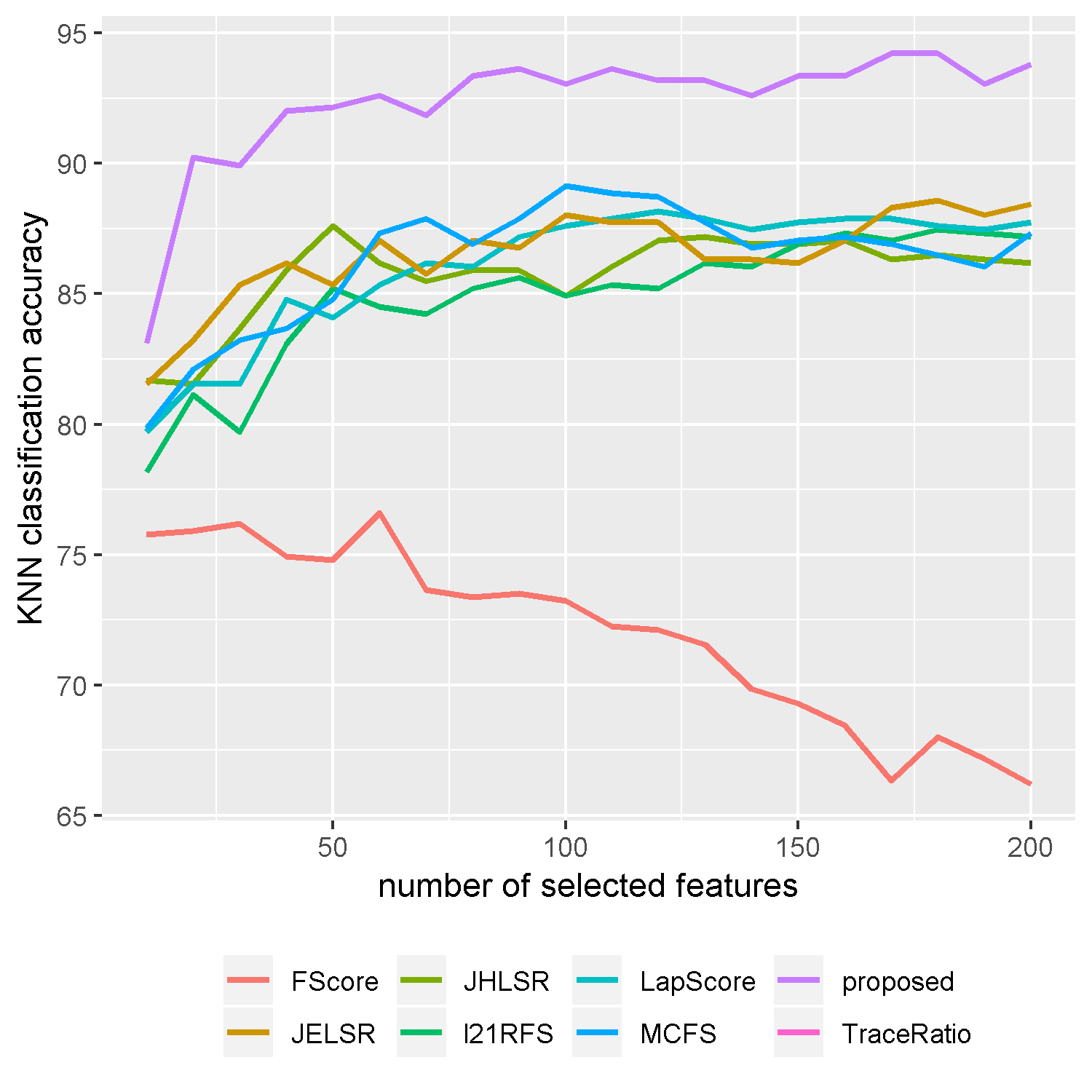}
\caption{Hepatitis-C}
\end{subfigure}
\caption{KNN classification accuracy vs. number of selected features on benchmark data sets.}
\label{figure:results_knn}
\end{figure}

\begin{figure}[!htbp]
\centering
\begin{subfigure}[b]{0.325\textwidth}
\includegraphics[scale=0.42]{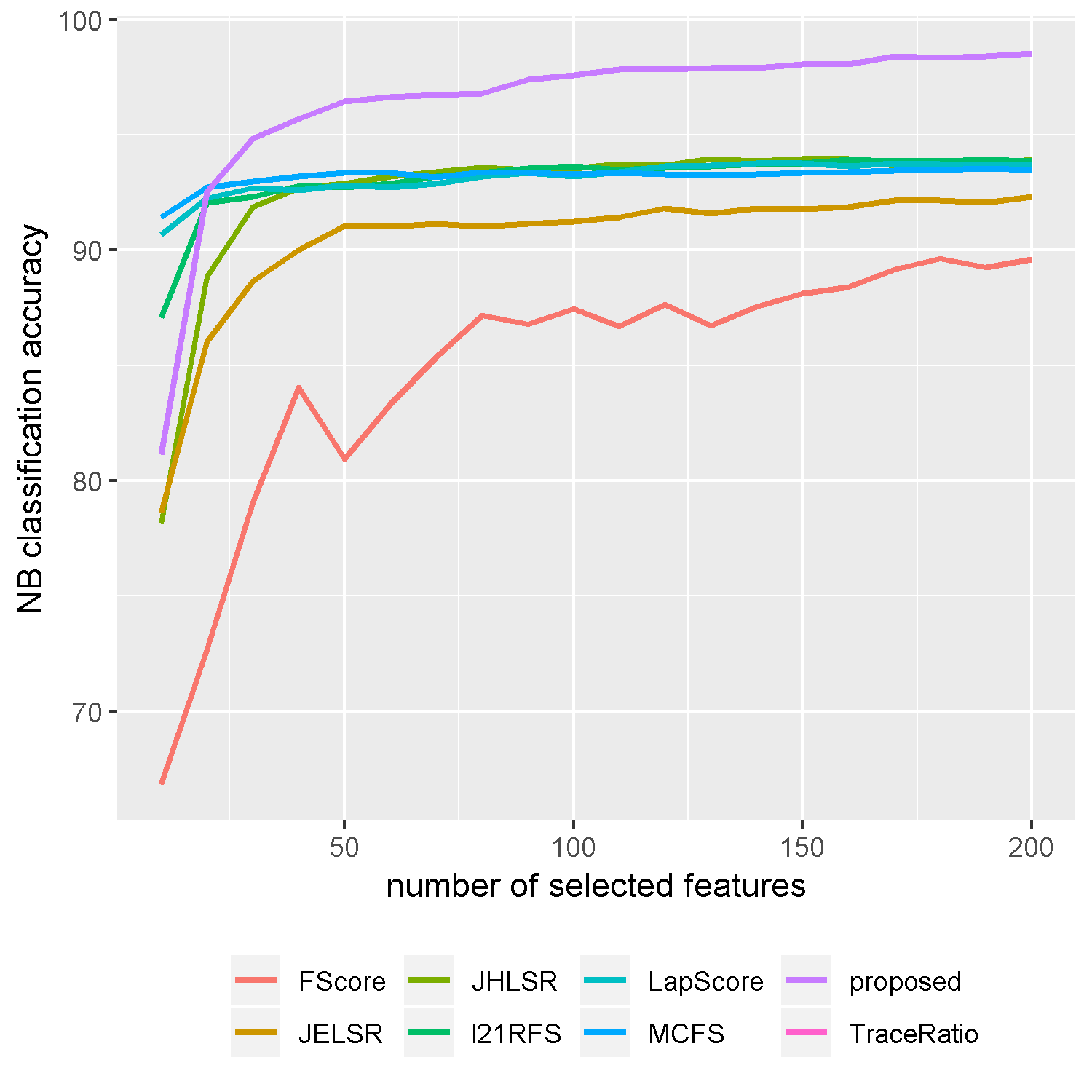}
\caption{Gene-Expression}
\end{subfigure}
\begin{subfigure}[b]{0.325\textwidth}
\includegraphics[scale=0.42]{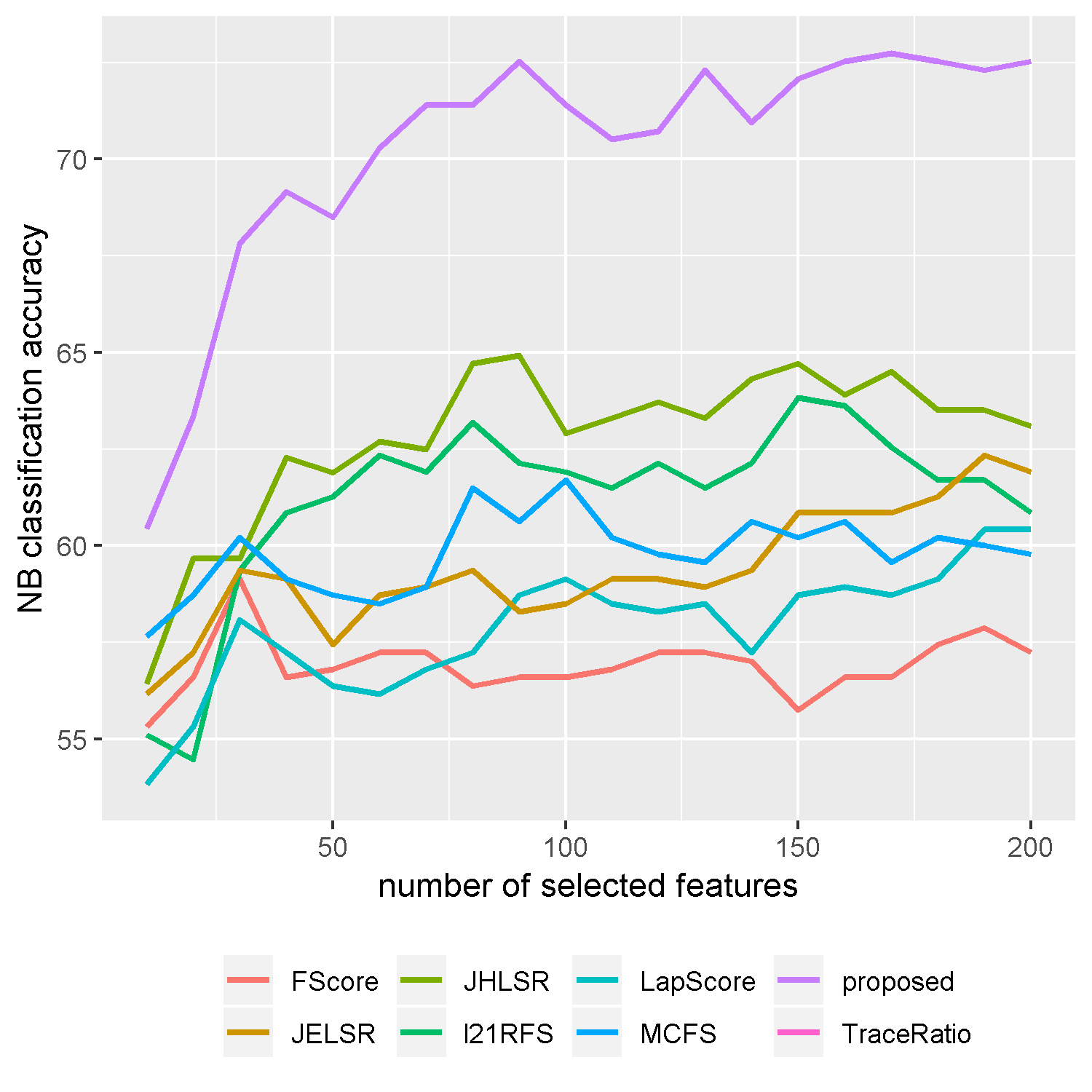}
\caption{Smoke-Cancer}
\end{subfigure}
\begin{subfigure}[b]{0.325\textwidth}
\includegraphics[scale=0.42]{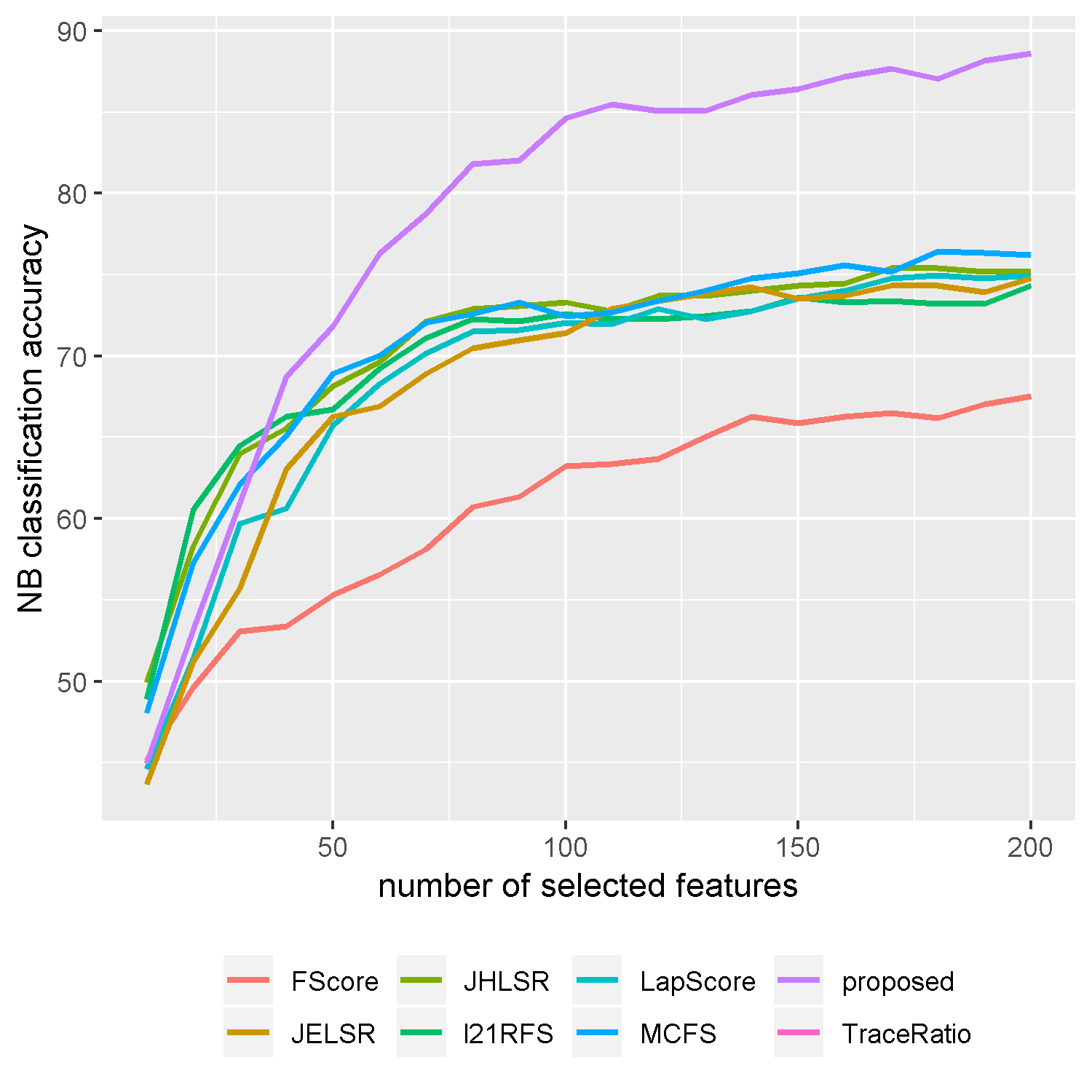}
\caption{Various-Cancers}
\end{subfigure}
\begin{subfigure}[b]{0.325\textwidth}
\includegraphics[scale=0.42]{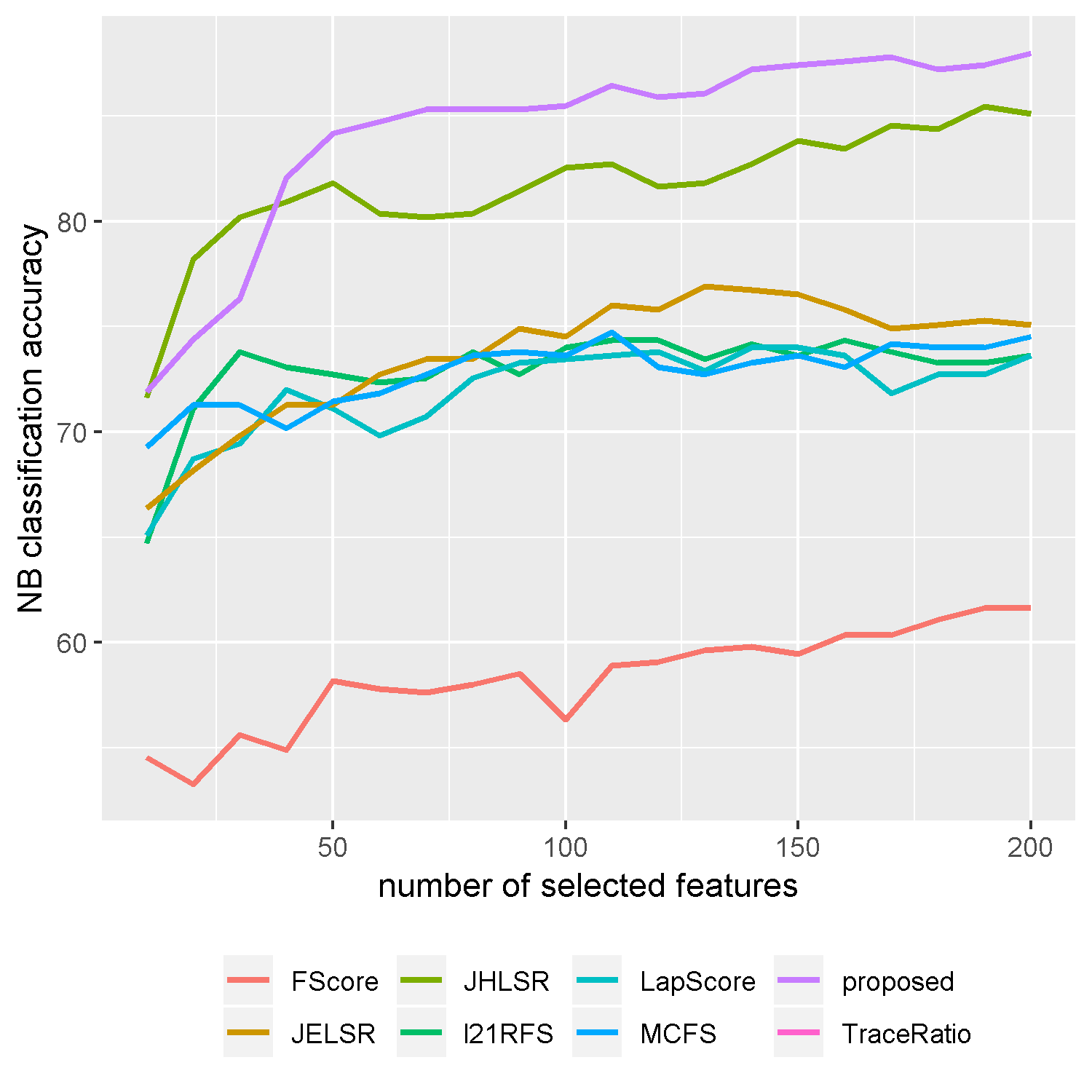}
\caption{Burkitt-Lymphoma}
\end{subfigure}
\begin{subfigure}[b]{0.325\textwidth}
\includegraphics[scale=0.42]{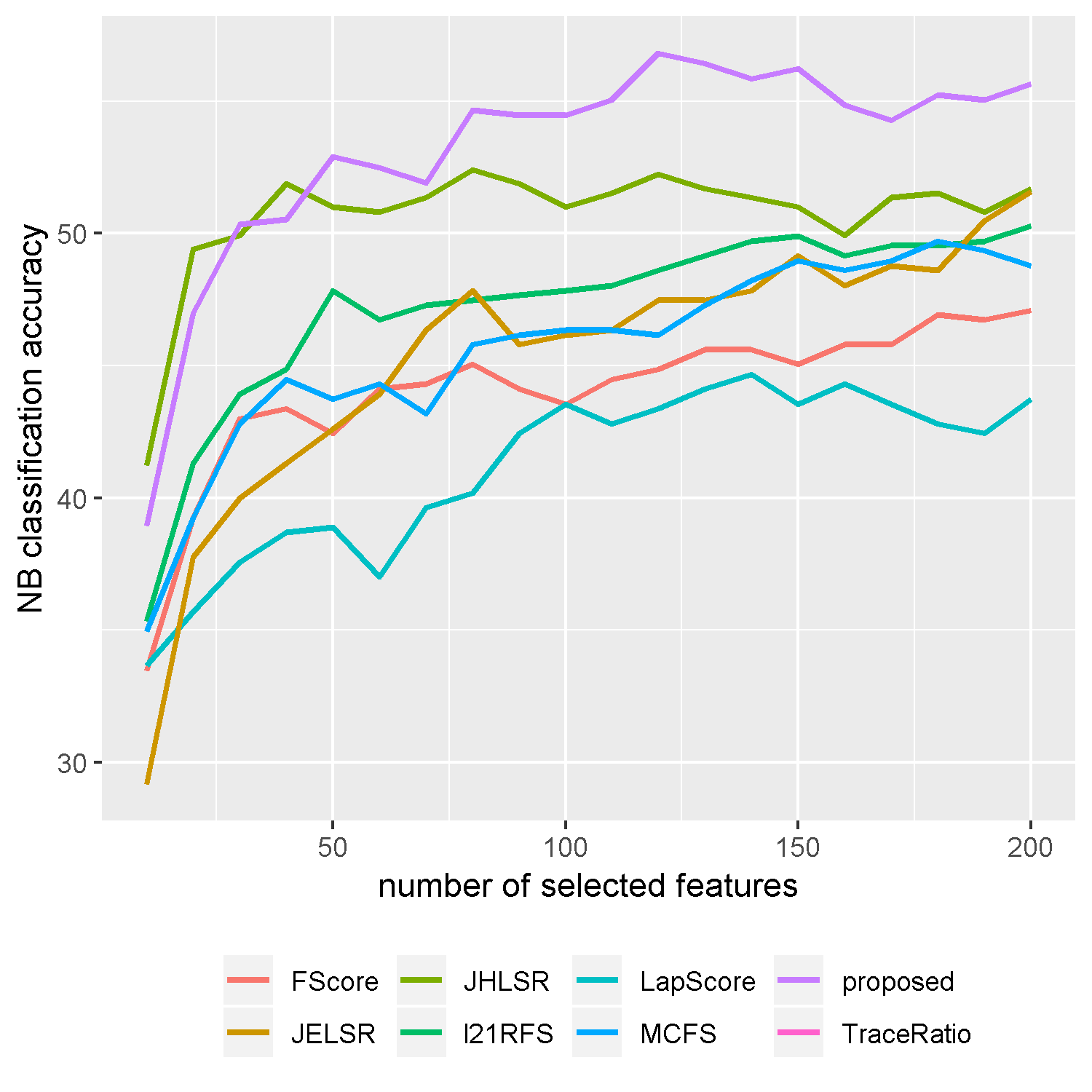}
\caption{Mouse-Type}
\end{subfigure}
\begin{subfigure}[b]{0.325\textwidth}
\includegraphics[scale=0.42]{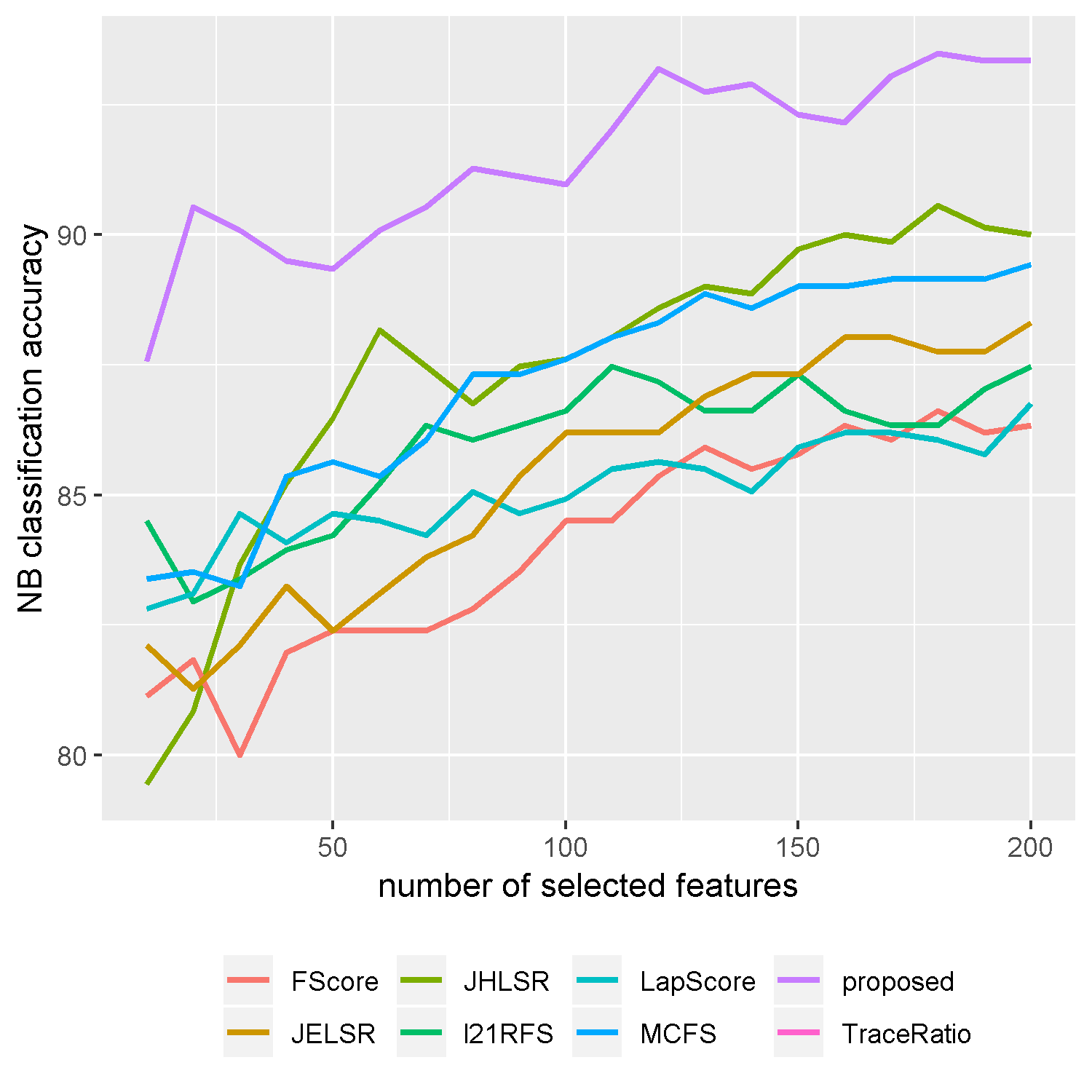}
\caption{Hepatitis-C}
\end{subfigure}
\caption{NB classification accuracy vs. number of selected features on benchmark data sets.}
\label{figure:results_nb}
\end{figure}

The result of the exhaustive experiments conducted on various data sets prove efficiency of our proposed method. Although our proposed unsupervised framework, has lower computational complexity in most cases, it outperforms state-of-the-art supervised and unsupervised feature selection methods. Average and standard deviation of accuracies of each method on every data set are reported in tables \ref{table:results_knn}, \ref{table:results_lsvm}, \ref{table:results_rbsvm}, \ref{table:results_nb}. The highest accuracies are boldfaced. We enhance average classification accuracy by 3.79\% (JHLSR), 16.88\% (FScore), 7.56\% (l21RFS), 7.3\% (TraceRatio), 8.38\% (LapScore), 8.12\% (JELSR), and 7.38\% (MCFS) by KNN classification method. Using L-SVM classifier, we improve average classification accuracy by 3.47\% (JHLSR), 16.75\% (FScore), 8.98\% (l21RFS), 8.12\% (TraceRatio), 9.13\% (LapScore), 8.98\% (JELSR), and 8.22\% (MCFS). By classifying using RB-SVM, we improve average classification accuracy by 3.93\% (JHLSR), 18.86\% (FScore), 8.1\% (l21RFS), 8.15\% (TraceRatio), 9.02\% (LapScore), 8.93\% (JELSR), and 8.29\% (MCFS). Using NB classification method, we enhance average classification accuracy by 4.75\% (JHLSR), 14.19\% (FScore), 7.33\% (l21RFS), 7.46\% (TraceRatio), 9.37\% (LapScore), 8.57\% (JELSR), and 7.46\% (MCFS). Accuracies of different methods for different number of features is also compared in figures \ref{figure:results_knn}, \ref{figure:results_lsvm}, \ref{figure:results_rbsvm}, \ref{figure:results_nb}.

\begin{table}
\centering
\scalebox{0.75}{
\begin{tabular}{ |P{4cm}|P{2cm}|P{2cm}|P{2cm}|P{2cm}|P{2cm}|P{2cm}|P{2cm}|P{2cm}| }
  \hline
  Data set & JHLSR & FScore & l21RFS & TraceRatio & LapScore & JELSR & MCFS & proposed\\
  \hline
  Gene-Expression & 93.38$\pm{0.37}$ & 88.92$\pm{1.1}$ & 93.58$\pm{0.57}$ & 93.84$\pm{0.53}$ & 94.01$\pm{0.5}$ & 91.87$\pm{0.87}$ & 93.84$\pm{0.53}$ & \textbf{97.13}$\pm{0.6}$\\
  Smoke-Cancer & 59.26$\pm{2.91}$ & 57.87$\pm{1.11}$ & 59.87$\pm{1.36}$ & 59.94$\pm{2.79}$ & 58.51$\pm{2.13}$ & 59.2$\pm{3.08}$ & 59.43$\pm{2.47}$ & \textbf{64.71}$\pm{1.44}$\\
  Various-Cancers & 81.14$\pm{1.83}$ & 66.54$\pm{1.75}$ & 75.21$\pm{5.09}$ & 74.87$\pm{4.72}$ & 73.27$\pm{5.33}$ & 73.72$\pm{5.24}$ & 74.87$\pm{4.72}$ & \textbf{82.54}$\pm{2.38}$\\
  Burkitt-Lymphoma & 76.66$\pm{1.42}$ & 56.39$\pm{6.93}$ & 72.98$\pm{5.19}$ & 73.37$\pm{4.37}$ & 72.99$\pm{5.42}$ & 71.82$\pm{2.44}$ & 73.37$\pm{4.37}$ & \textbf{79.49}$\pm{3.84}$\\
  Mouse-Type & 67.25$\pm{4.7}$ & 43.19$\pm{4.63}$ & 54.31$\pm{8.95}$ & 54.07$\pm{4.2}$ & 51.01$\pm{4.52}$ & 54.28$\pm{7.45}$ & 54.07$\pm{4.2}$ & \textbf{69.99}$\pm{3.36}$\\
  Hepatitis-C & 85.77$\pm{2.92}$ & 71.96$\pm{11.63}$ & 84.89$\pm{3.14}$ & 86.3$\pm{2.41}$ & 86.09$\pm{1.52}$ & 86.56$\pm{1.17}$ & 86.3$\pm{2.41}$ & \textbf{92.33}$\pm{2.75}$\\
  \hline
  AVERAGE & 77.24 & 64.15 & 73.47 & 73.73 & 72.65 & 72.91 & 73.65 & \textbf{81.03}\\
  \hline
\end{tabular}
}
\caption{Average KNN classification accuracy of the methods on benchmark data sets.}
\label{table:results_knn}
\end{table}

\begin{table}
\centering
\scalebox{0.75}{
\begin{tabular}{ |P{4cm}|P{2cm}|P{2cm}|P{2cm}|P{2cm}|P{2cm}|P{2cm}|P{2cm}|P{2cm}| }
  \hline
  Data set & JHLSR & FScore & l21RFS & TraceRatio & LapScore & JELSR & MCFS & proposed\\
  \hline
  Gene-Expression & 93.64$\pm{0.24}$ & 91.3$\pm{1.12}$ & 94.04$\pm{0.41}$ & 94.41$\pm{0.32}$ & 94.43$\pm{0.27}$ & 92.86$\pm{0.67}$ & 94.4$\pm{0.33}$ & \textbf{97.72}$\pm{0.2}$\\
  Smoke-Cancer & 60.1$\pm{2.89}$ & 49.33$\pm{3.79}$ & 59.45$\pm{3.87}$ & 57.54$\pm{5.9}$ & 54.51$\pm{2.53}$ & 59.35$\pm{3.33}$ & 57.44$\pm{5.81}$ & \textbf{65.47}$\pm{3.66}$\\
  Various-Cancers & 76.46$\pm{1.5}$ & 61.66$\pm{1.79}$ & 72.06$\pm{5.87}$ & 72.28$\pm{3.53}$ & 69.91$\pm{4.69}$ & 70.33$\pm{5.37}$ & 72.26$\pm{3.46}$ & \textbf{78.27}$\pm{2.09}$\\
  Burkitt-Lymphoma & 83.01$\pm{1.8}$ & 63.25$\pm{7.84}$ & 76.57$\pm{4.15}$ & 78.17$\pm{3.28}$ & 77.32$\pm{5.54}$ & 77.49$\pm{4.36}$ & 78.22$\pm{3.21}$ & \textbf{85.33}$\pm{3.11}$\\
  Mouse-Type & 70.86$\pm{4.28}$ & 41.18$\pm{3.46}$ & 54.17$\pm{9.34}$ & 55.01$\pm{6.68}$ & 55.05$\pm{3.52}$ & 51.91$\pm{9.55}$ & 54.64$\pm{6.47}$ & \textbf{73.82}$\pm{3.77}$\\
  Hepatitis-C & 90.33$\pm{1.36}$ & 88.1$\pm{2.64}$ & 89.14$\pm{0.81}$ & 89.18$\pm{0.69}$ & 89.28$\pm{1.24}$ & 89.46$\pm{0.54}$ & 89.04$\pm{0.68}$ & \textbf{94.7}$\pm{1.66}$\\
  \hline
  AVERAGE & 79.07 & 65.8 & 73.57 & 74.43 & 73.42 & 73.57 & 74.33 & \textbf{82.55}\\
  \hline
\end{tabular}
}
\caption{Average L-SVM classification accuracy of the methods on benchmark data sets.}
\label{table:results_lsvm}
\end{table}

\begin{table}
\centering
\scalebox{0.75}{
\begin{tabular}{ |P{4cm}|P{2cm}|P{2cm}|P{2cm}|P{2cm}|P{2cm}|P{2cm}|P{2cm}|P{2cm}| }
  \hline
  Data set & JHLSR & FScore & l21RFS & TraceRatio & LapScore & JELSR & MCFS & proposed\\
  \hline
  Gene-Expression & 93.59$\pm{0.25}$ & 91.19$\pm{1.18}$ & 93.59$\pm{0.68}$ & 93.77$\pm{0.69}$ & 93.8$\pm{0.7}$ & 92.53$\pm{0.95}$ & 93.76$\pm{0.71}$ & \textbf{97.74}$\pm{0.22}$\\
  Smoke-Cancer & 65.78$\pm{3.88}$ & 44.83$\pm{12.44}$ & 65.56$\pm{5.79}$ & 67.49$\pm{4.82}$ & 66.22$\pm{4.52}$ & 65.15$\pm{4.28}$ & 67.16$\pm{4.53}$ & \textbf{71.56}$\pm{3.33}$\\
  Various-Cancer & 82.04$\pm{2.42}$ & 65.53$\pm{2.53}$ & 76.82$\pm{5.5}$ & 75.27$\pm{4.09}$ & 74.21$\pm{3.01}$ & 75$\pm{4.22}$ & 74.77$\pm{3.88}$ & \textbf{84.5}$\pm{2.69}$\\
  Burkitt-Lymphoma & 84.22$\pm{2}$ & 61$\pm{8.4}$ & 77.71$\pm{4.73}$ & 77.83$\pm{1.91}$ & 76.69$\pm{6.1}$ & 77.54$\pm{2.88}$ & 77.62$\pm{2.26}$ & \textbf{86.06}$\pm{1.99}$\\
  Mouse-Type & 67.35$\pm{2.8}$ & 43.62$\pm{5.15}$ & 55.32$\pm{8.48}$ & 53.83$\pm{6.12}$ & 52.87$\pm{5.64}$ & 53.58$\pm{7.91}$ & 54.04$\pm{5.82}$ & \textbf{71.4}$\pm{4.14}$\\
  Hepatitis-C & 90.32$\pm{1.93}$ & 87.56$\pm{2.55}$ & 89.29$\pm{1.15}$ & 89.77$\pm{1.08}$ & 88.94$\pm{1.14}$ & 89.51$\pm{0.81}$ & 89.77$\pm{1.04}$ & \textbf{95.44}$\pm{2.14}$\\
  \hline
  AVERAGE & 80.55 & 65.62 & 76.38 & 76.33 & 75.46 & 75.55 & 76.19 & \textbf{84.48}\\
  \hline
\end{tabular}
}
\caption{Average RB-SVM classification accuracy of the methods on benchmark data sets.}
\label{table:results_rbsvm}
\end{table}

\begin{table}
\centering
\scalebox{0.75}{
\begin{tabular}{ |P{4cm}|P{2cm}|P{2cm}|P{2cm}|P{2cm}|P{2cm}|P{2cm}|P{2cm}|P{2cm}| }
  \hline
  Data set & JHLSR & FScore & l21RFS & TraceRatio & LapScore & JELSR & MCFS & proposed\\
  \hline
  Gene-Expression & 92.49$\pm{0.49}$ & 84.82$\pm{3.45}$ & 93.06$\pm{0.68}$ & 93.21$\pm{1.38}$ & 93.16$\pm{1.07}$ & 90.45$\pm{1.27}$ & 93.21$\pm{1.38}$ & \textbf{96.36}$\pm{0.8}$\\
  Smoke-Cancer & 62.78$\pm{1}$ & 56.91$\pm{2.84}$ & 61.2$\pm{3.43}$ & 59.82$\pm{3.27}$ & 57.89$\pm{4.24}$ & 59.39$\pm{3.8}$ & 59.82$\pm{3.27}$ & \textbf{70.27}$\pm{1.85}$\\
  Various-Cancers & 70.56$\pm{3.17}$ & 60.71$\pm{3.73}$ & 69.75$\pm{4.6}$ & 70.58$\pm{4.42}$ & 68.63$\pm{3.66}$ & 68.38$\pm{2.11}$ & 70.58$\pm{4.42}$ & \textbf{78.5}$\pm{3.67}$\\
  Burkitt-Lymphoma & 81.66$\pm{3.7}$ & 58.35$\pm{5.81}$ & 72.96$\pm{7.04}$ & 72.82$\pm{4.43}$ & 71.95$\pm{4.99}$ & 73.71$\pm{3.86}$ & 72.82$\pm{4.43}$ & \textbf{84.3}$\pm{1.97}$\\
  Mouse-Type & 50.7$\pm{3.29}$ & 44.03$\pm{6.61}$ & 47.2$\pm{5.19}$ & 45.67$\pm{3.37}$ & 41.13$\pm{7.98}$ & 45.34$\pm{5.49}$ & 45.67$\pm{3.37}$ & \textbf{53.16}$\pm{3.38}$\\
  Hepatitis-C & 87.39$\pm{0.86}$ & 84.08$\pm{2.44}$ & 85.93$\pm{1.89}$ & 87.18$\pm{2.2}$ & 85.06$\pm{5}$ & 85.38$\pm{1.16}$ & 87.18$\pm{2.2}$ & \textbf{91.48}$\pm{1.91}$\\
  \hline
  AVERAGE & 74.26 & 64.82 & 71.68 & 71.55 & 69.64 & 70.44 & 71.55 & \textbf{79.01}\\
  \hline
\end{tabular}
}
\caption{Average NB classification accuracy of the methods on benchmark data sets.}
\label{table:results_nb}
\end{table}

\begin{figure}[!htbp]
\centering
\begin{subfigure}[b]{0.325\textwidth}
\includegraphics[scale=0.42]{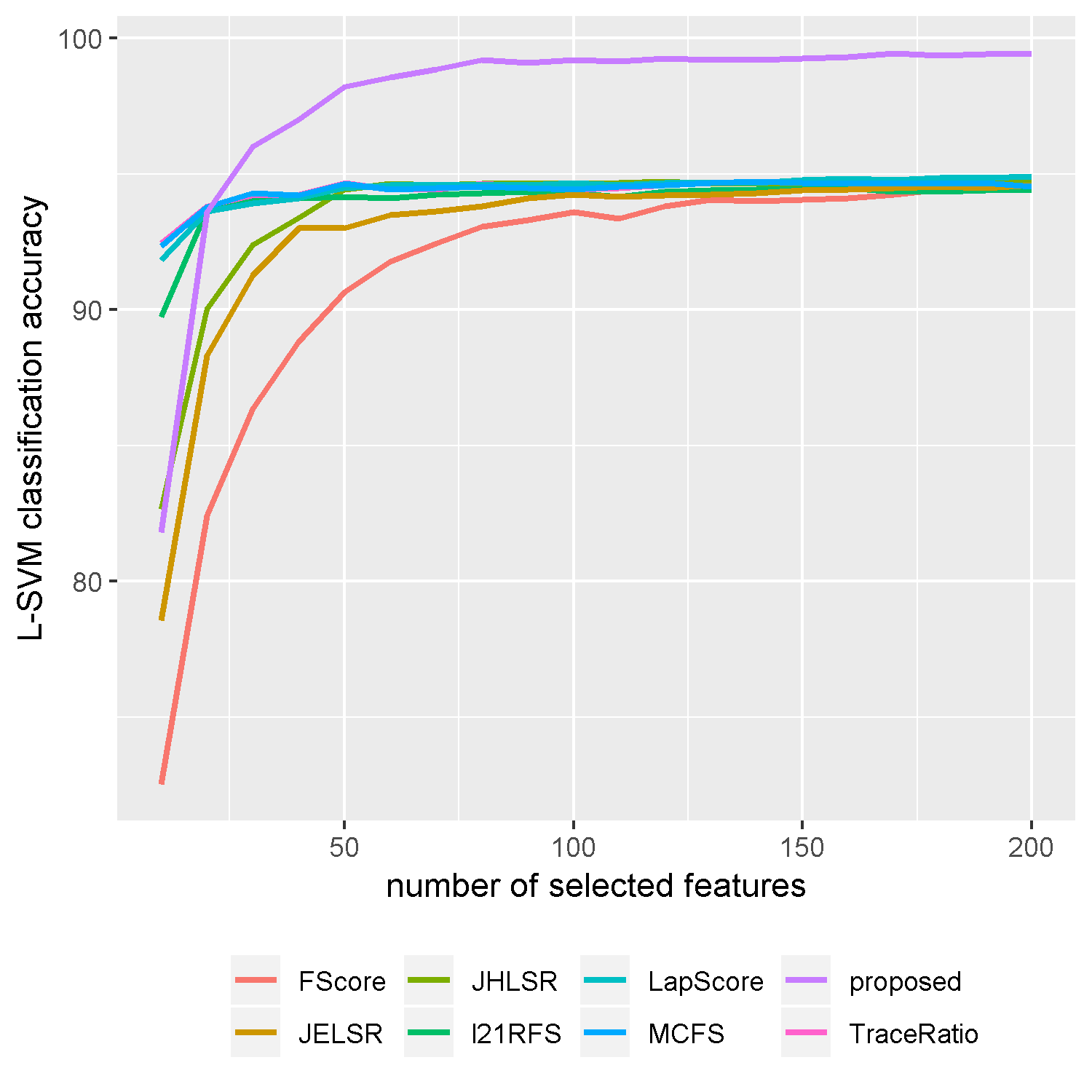}
\caption{Gene-Expression}
\end{subfigure}
\begin{subfigure}[b]{0.325\textwidth}
\includegraphics[scale=0.42]{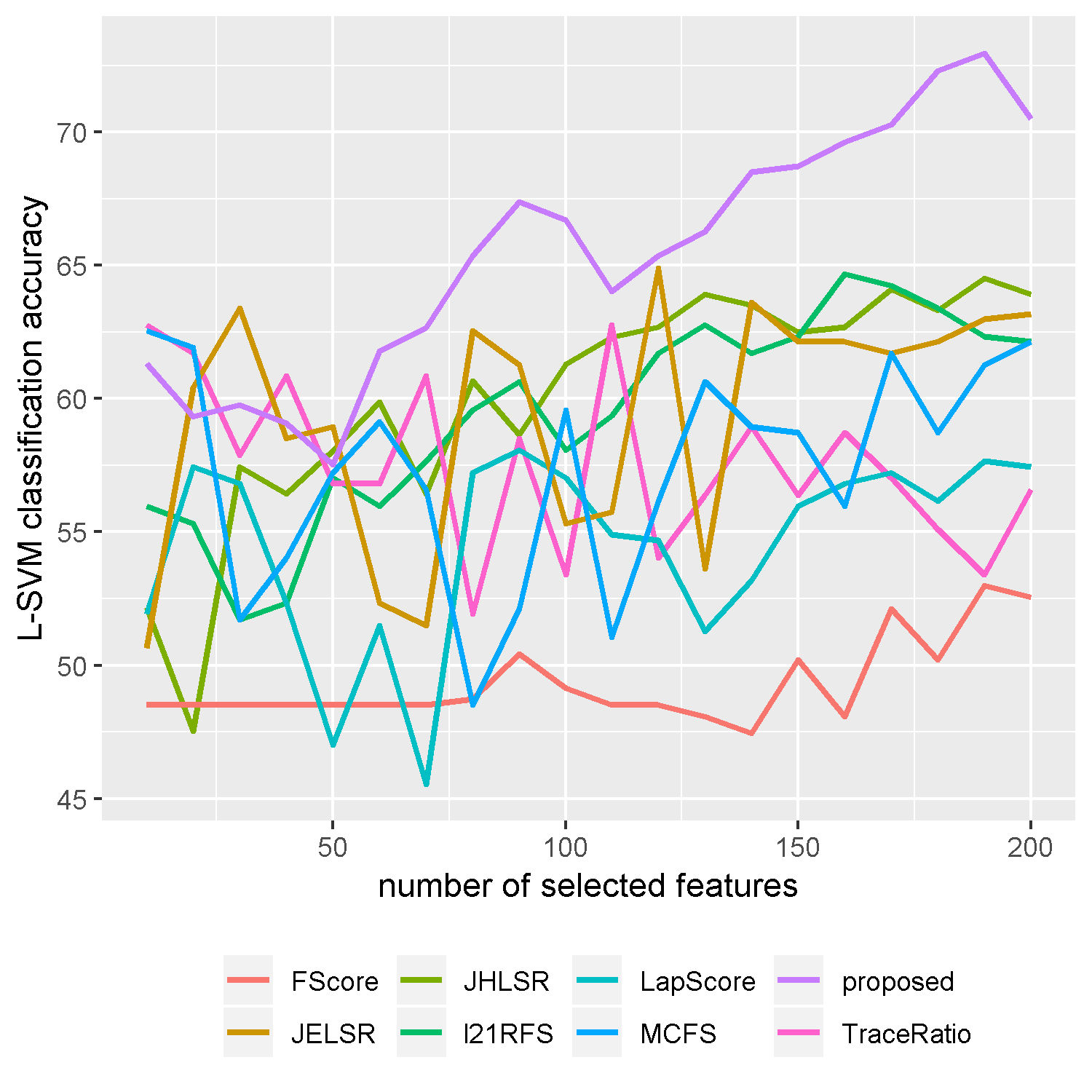}
\caption{Smoke-Cancer}
\end{subfigure}
\begin{subfigure}[b]{0.325\textwidth}
\includegraphics[scale=0.42]{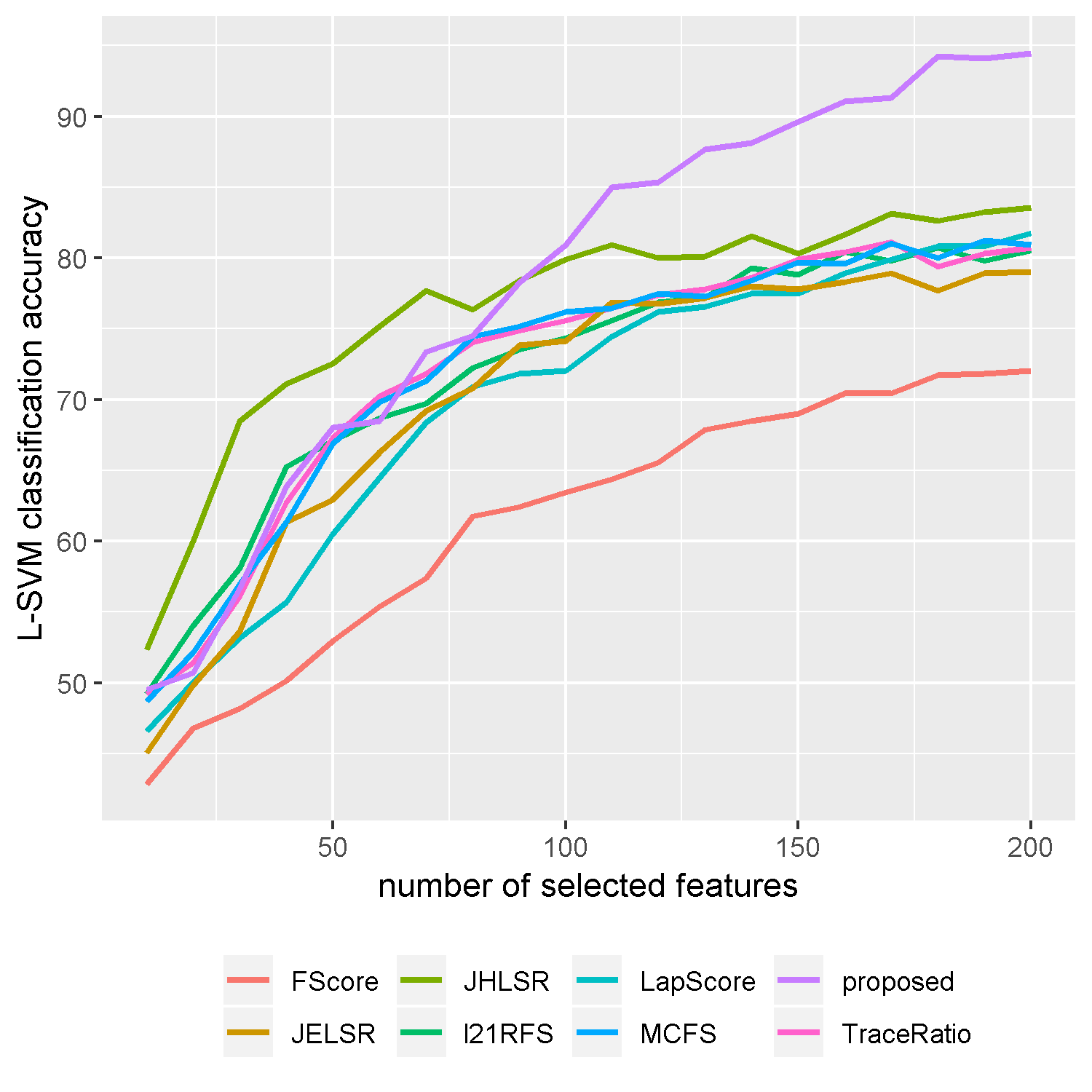}
\caption{Various-Cancers}
\end{subfigure}
\end{figure}
\begin{figure}[!htbp]
\ContinuedFloat
\centering
\begin{subfigure}[b]{0.325\textwidth}
\includegraphics[scale=0.42]{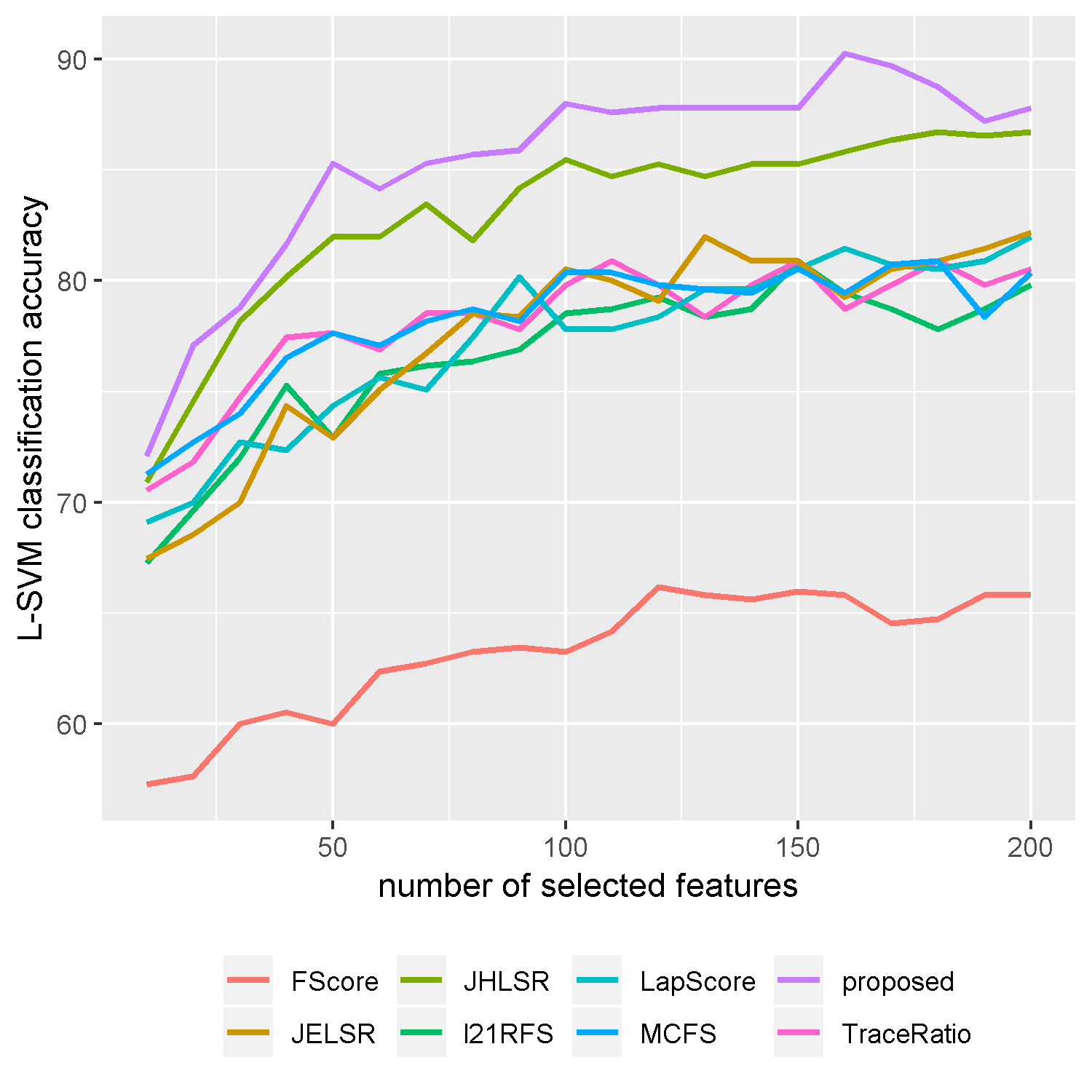}
\caption{Burkitt-Lymphoma}
\end{subfigure}
\begin{subfigure}[b]{0.325\textwidth}
\includegraphics[scale=0.42]{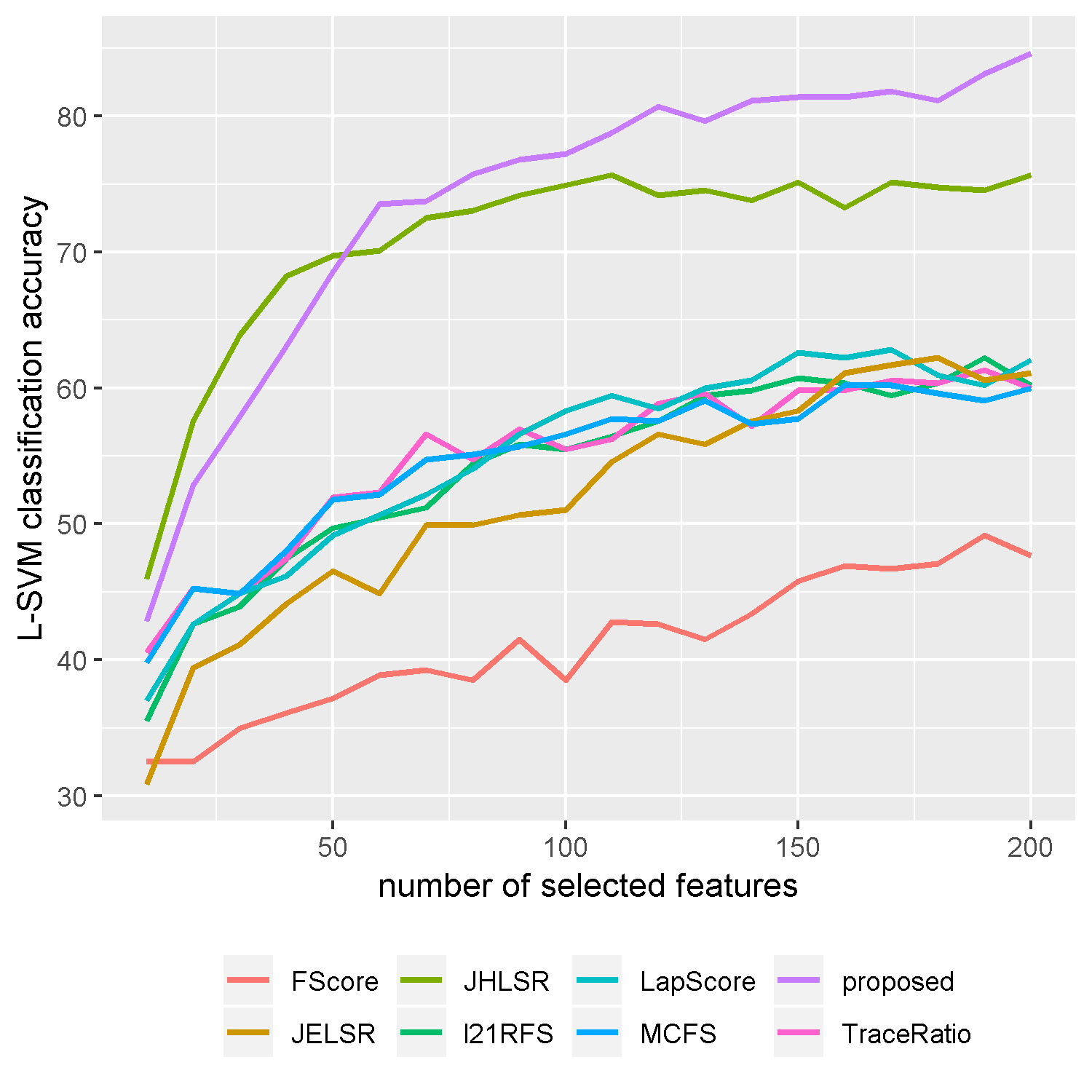}
\caption{Mouse-Type}
\end{subfigure}
\begin{subfigure}[b]{0.325\textwidth}
\includegraphics[scale=0.42]{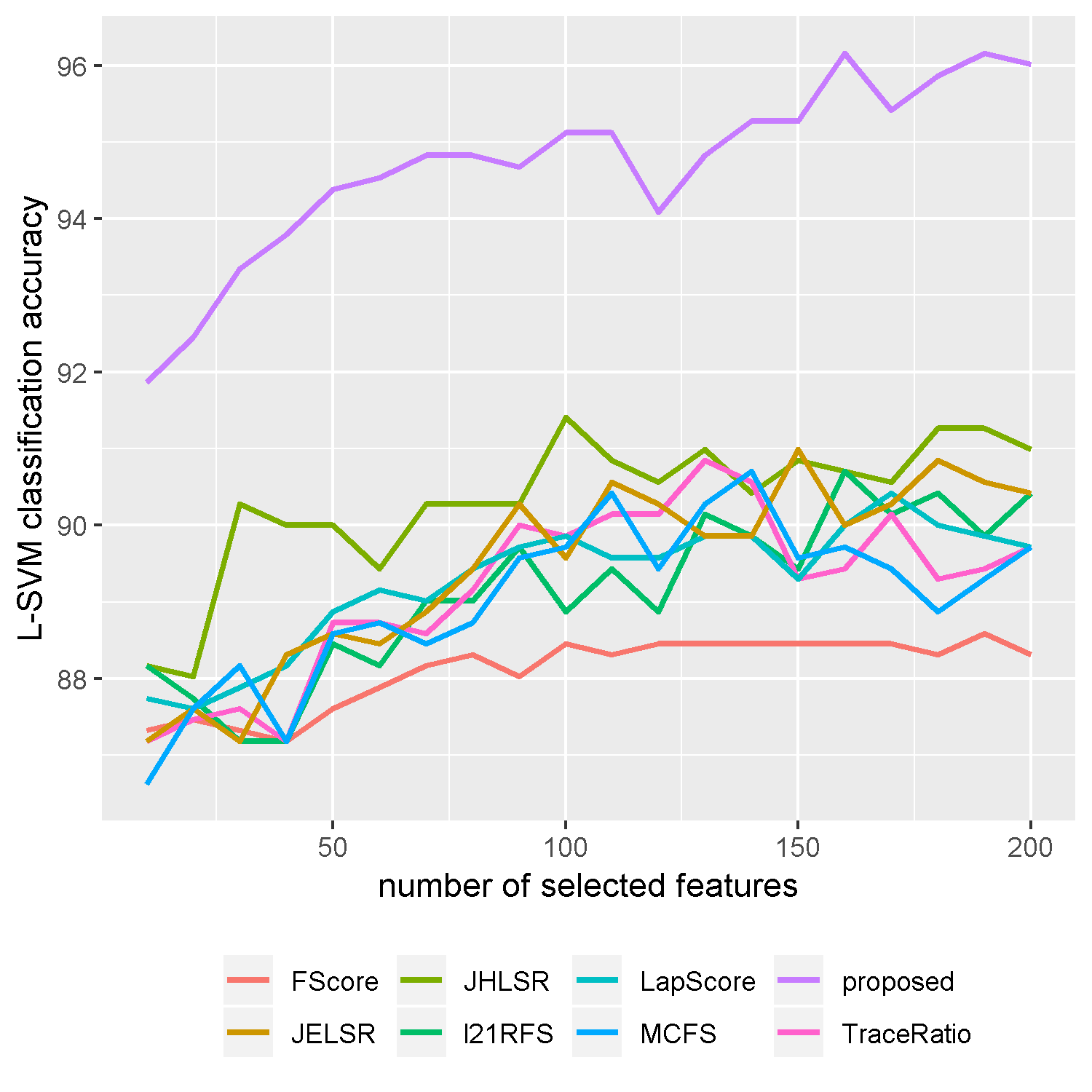}
\caption{Hepatitis-C}
\end{subfigure}
\caption{L-SVM classification accuracy vs. number of selected features on benchmark data sets.}
\label{figure:results_lsvm}
\end{figure}

\begin{figure}[!htbp]
\centering
\begin{subfigure}[b]{0.325\textwidth}
\includegraphics[scale=0.42]{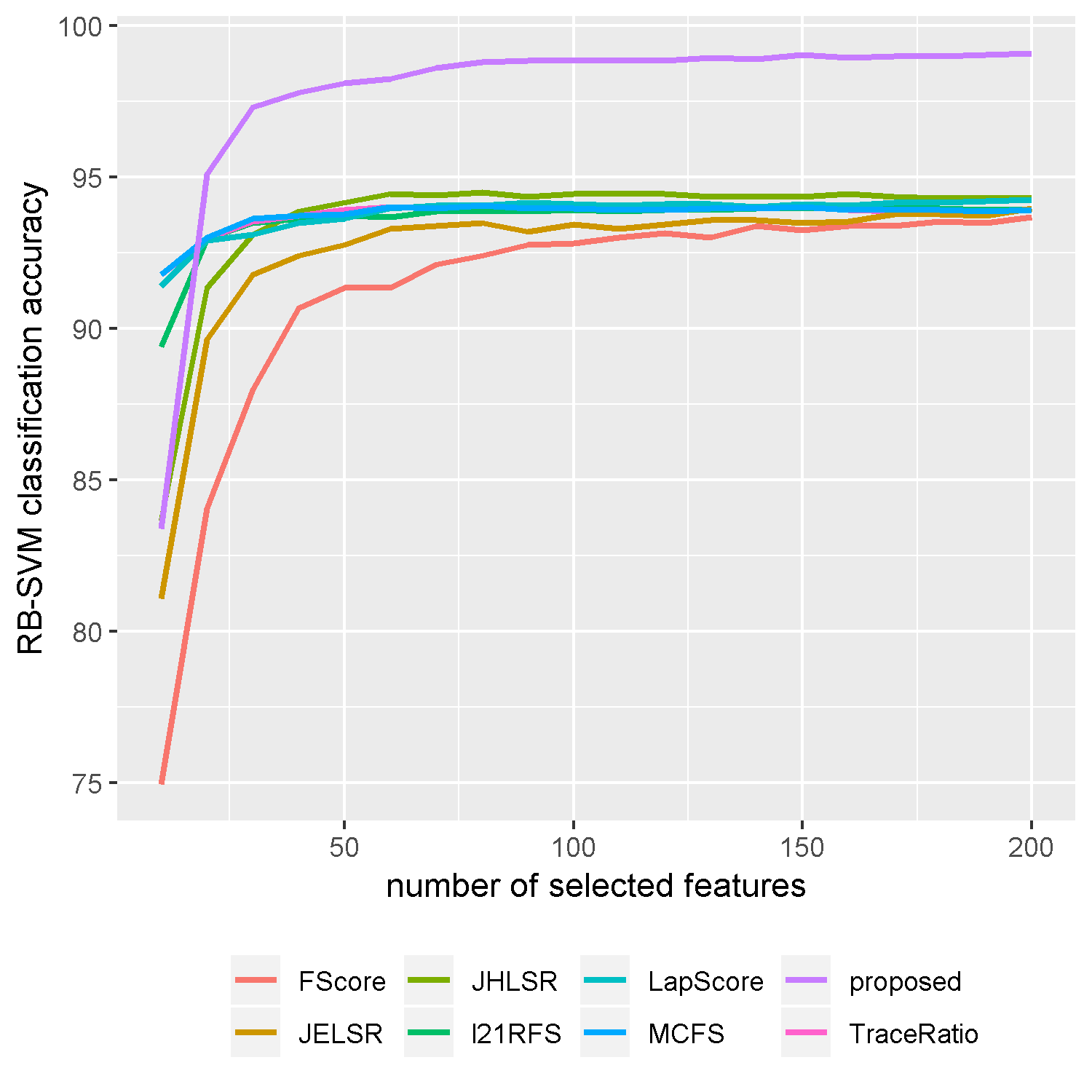}
\caption{Gene-Expression}
\end{subfigure}
\begin{subfigure}[b]{0.325\textwidth}
\includegraphics[scale=0.42]{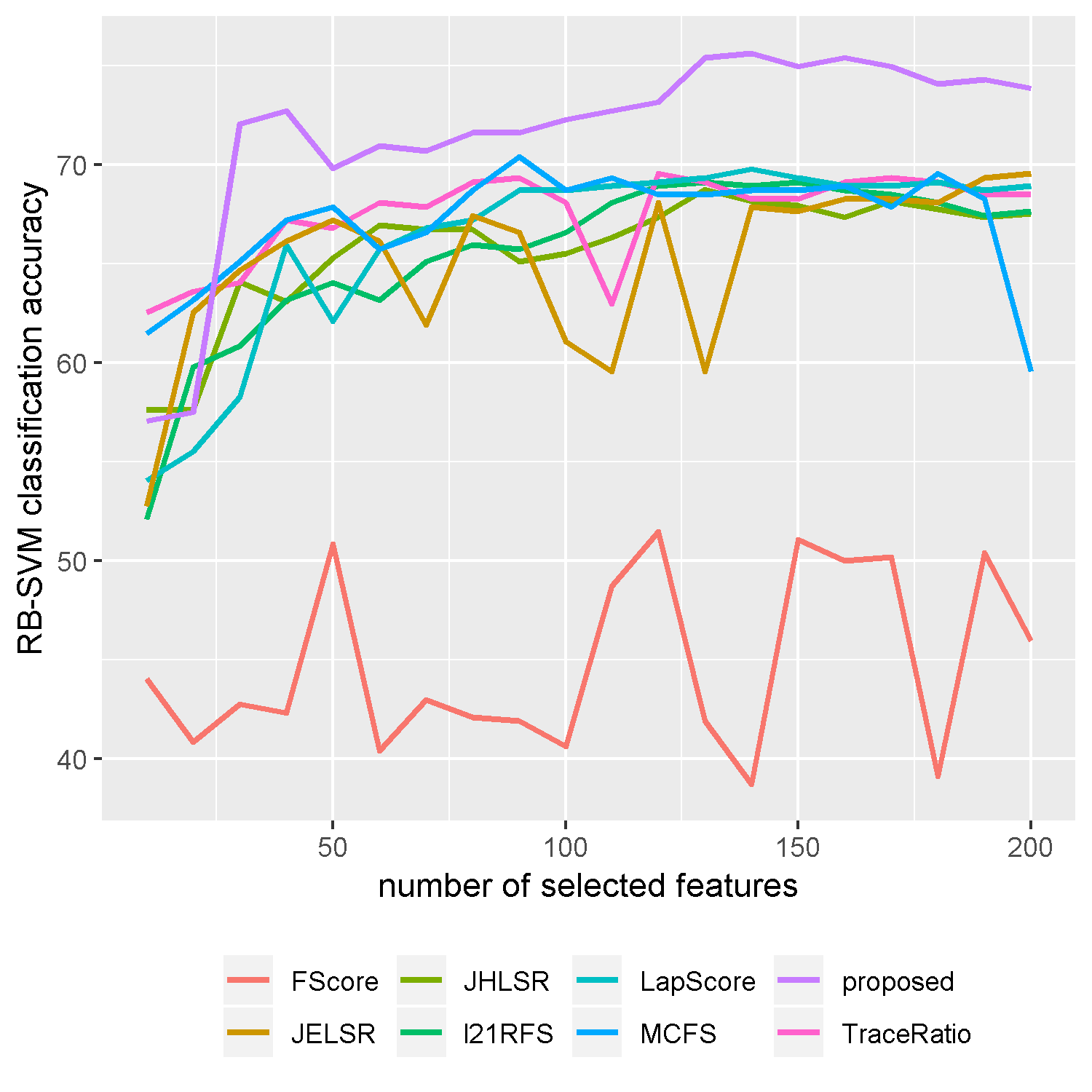}
\caption{Smoke-Cancer}
\end{subfigure}
\begin{subfigure}[b]{0.325\textwidth}
\includegraphics[scale=0.42]{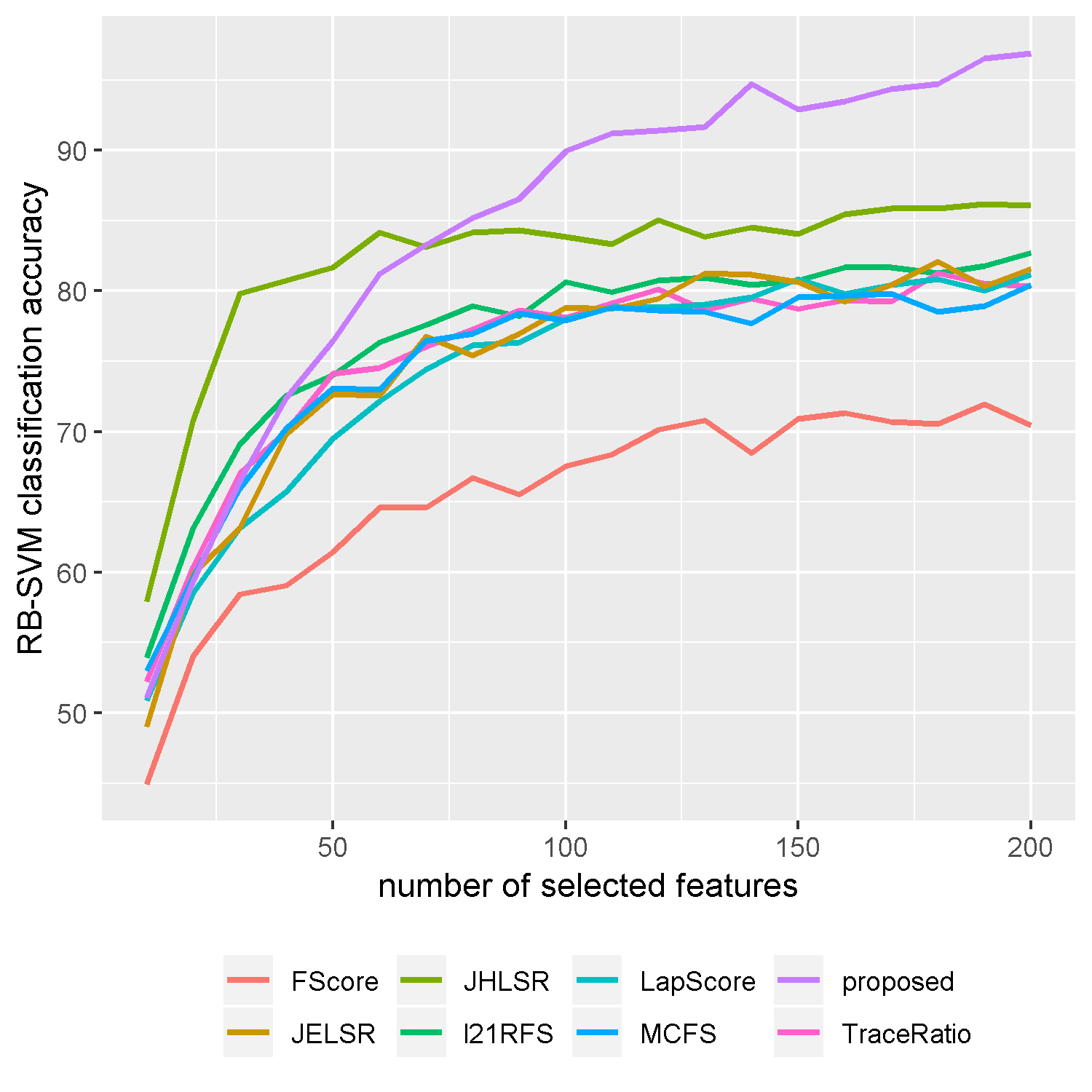}
\caption{Various-Cancers}
\end{subfigure}
\end{figure}
\begin{figure}[!htbp]
\ContinuedFloat
\centering
\begin{subfigure}[b]{0.325\textwidth}
\includegraphics[scale=0.42]{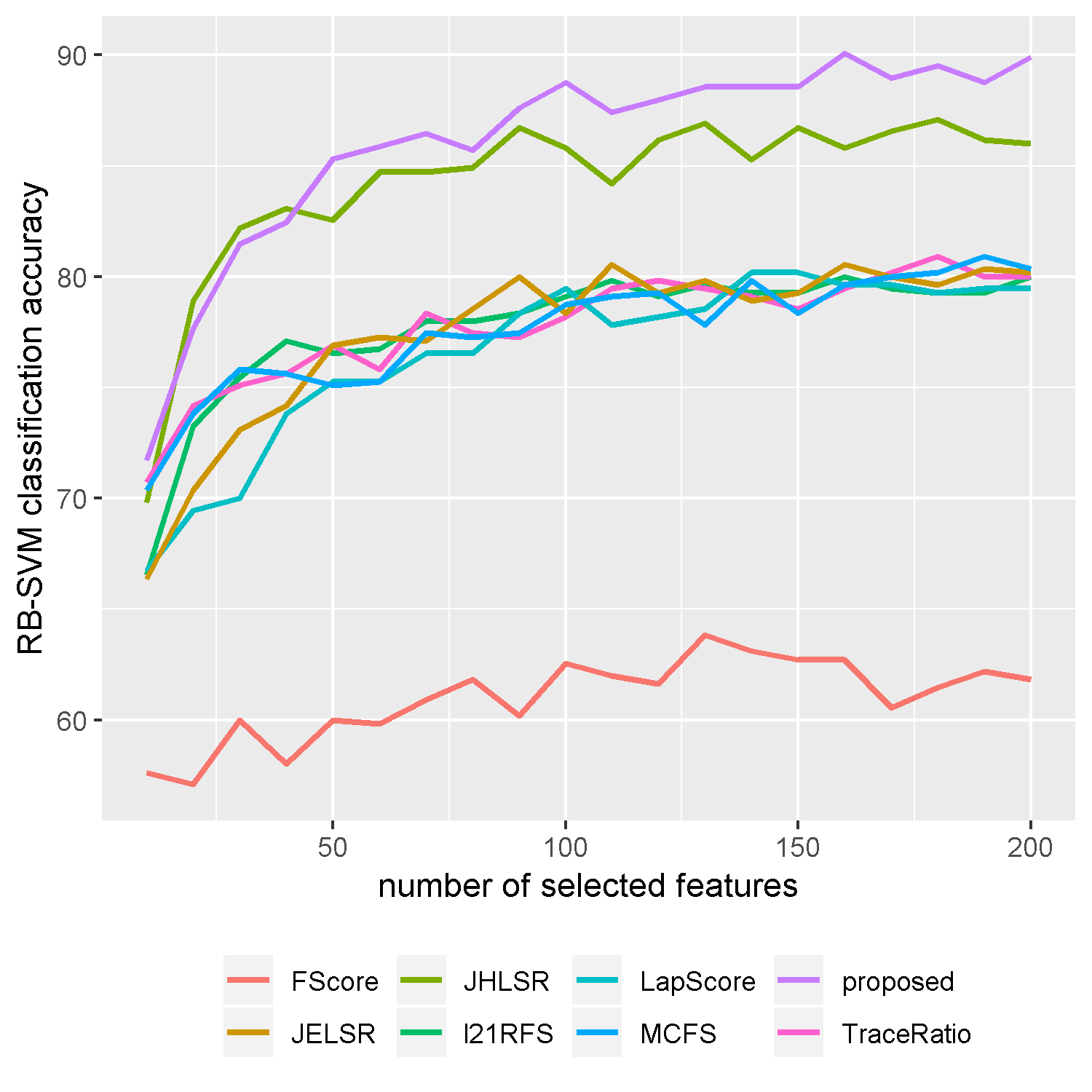}
\caption{Burkitt-Lymphoma}
\end{subfigure}
\begin{subfigure}[b]{0.325\textwidth}
\includegraphics[scale=0.42]{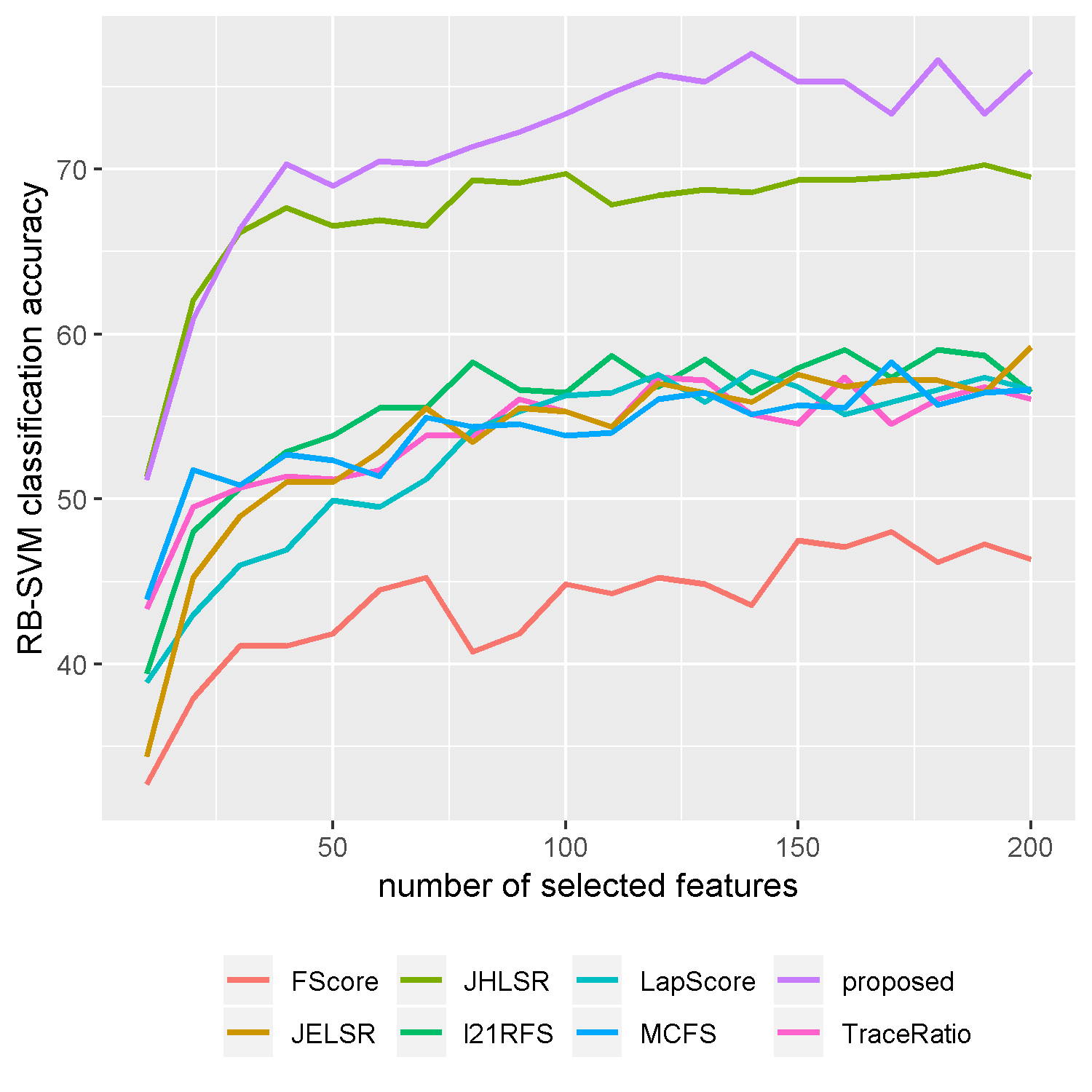}
\caption{Mouse-Type}
\end{subfigure}
\begin{subfigure}[b]{0.325\textwidth}
\includegraphics[scale=0.42]{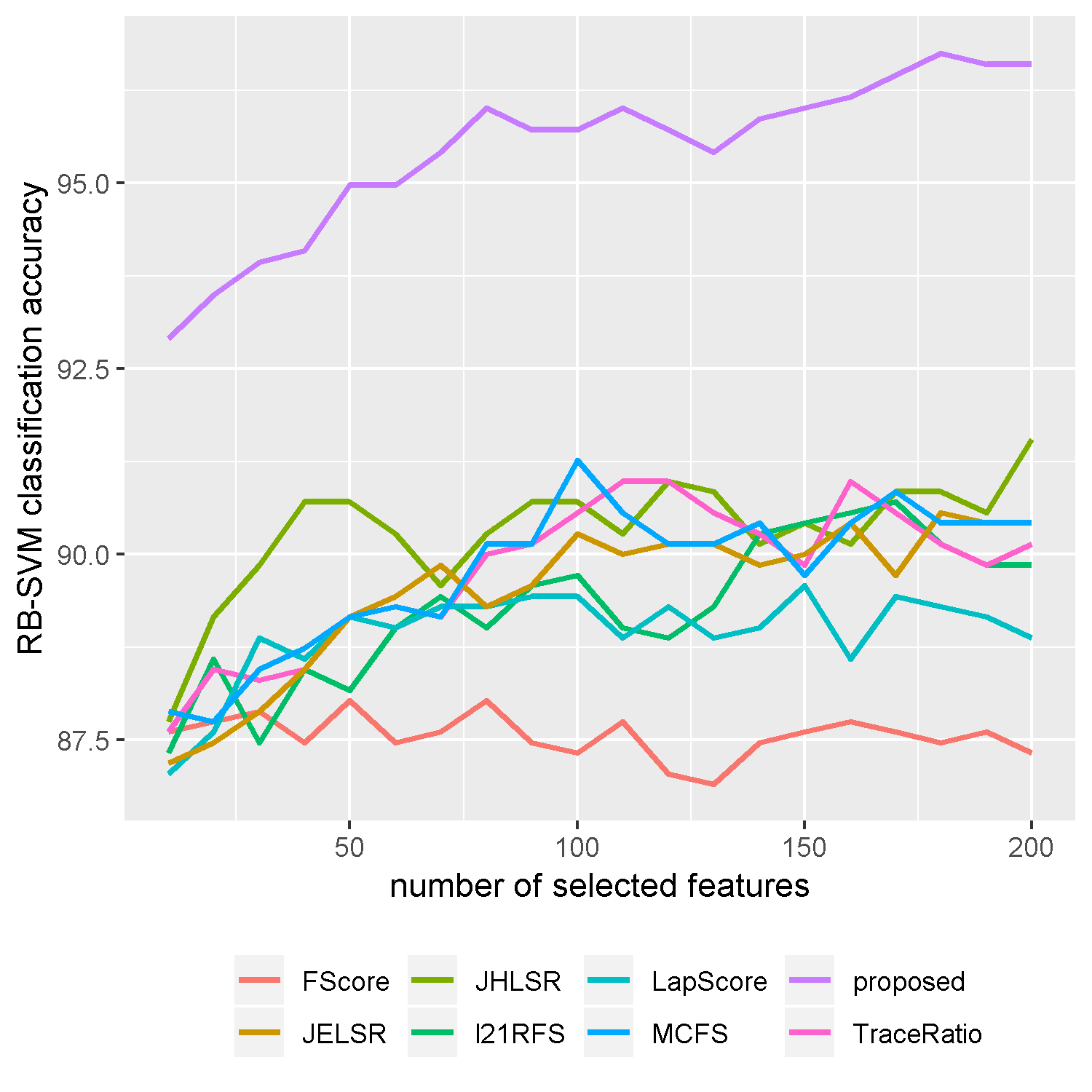}
\caption{Hepatitis-C}
\end{subfigure}
\caption{RB-SVM classification accuracy vs. number of selected features on benchmark data sets.}
\label{figure:results_rbsvm}
\end{figure}

\begin{figure}[!htbp]
\centering
\begin{subfigure}[b]{0.325\textwidth}
\includegraphics[scale=0.42]{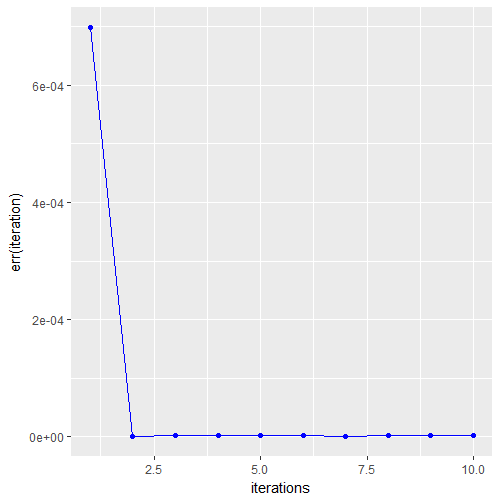}
\caption{Gene-Expression}
\end{subfigure}
\begin{subfigure}[b]{0.325\textwidth}
\includegraphics[scale=0.42]{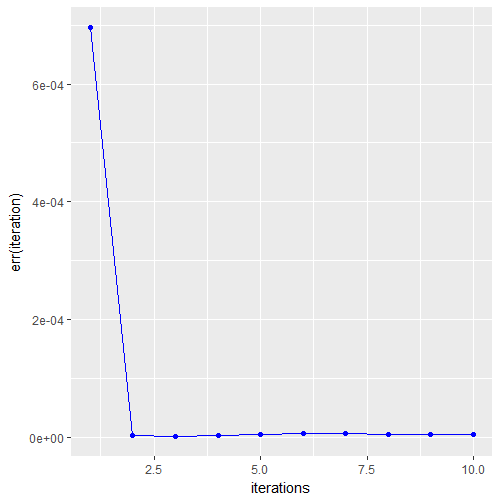}
\caption{Smoke-Cancer}
\end{subfigure}
\begin{subfigure}[b]{0.325\textwidth}
\includegraphics[scale=0.42]{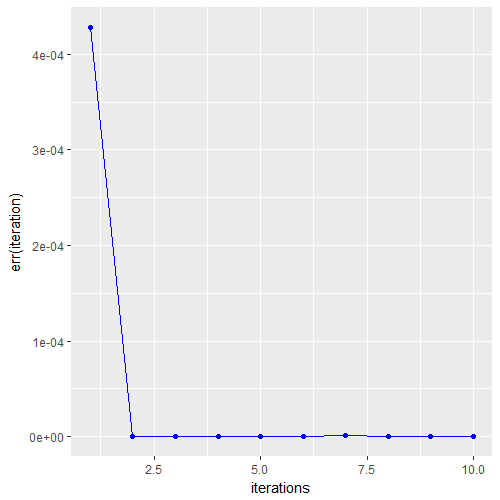}
\caption{Various-Cancers}
\end{subfigure}
\end{figure}
\begin{figure}[!htbp]
\ContinuedFloat
\centering
\begin{subfigure}[b]{0.325\textwidth}
\includegraphics[scale=0.42]{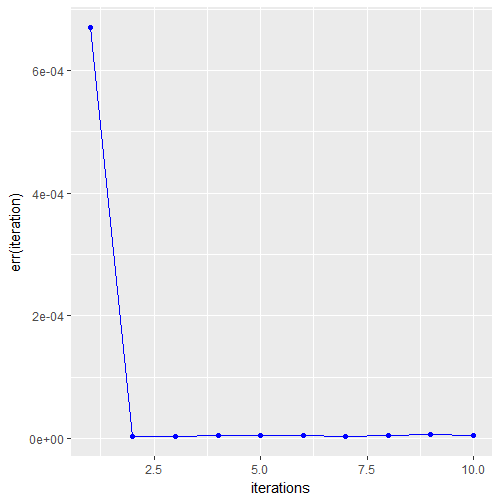}
\caption{Burkitt-Lymphoma}
\end{subfigure}
\begin{subfigure}[b]{0.325\textwidth}
\includegraphics[scale=0.42]{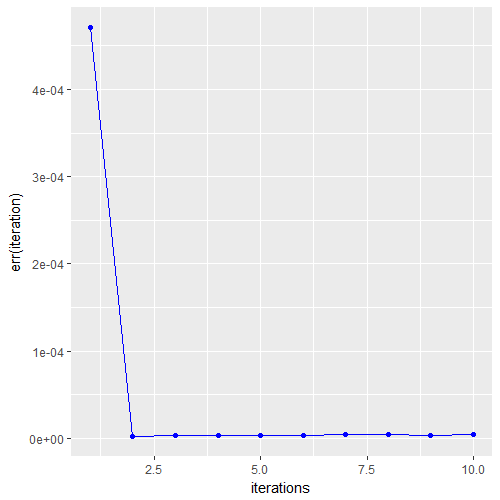}
\caption{Mouse-Type}
\end{subfigure}
\begin{subfigure}[b]{0.325\textwidth}
\includegraphics[scale=0.42]{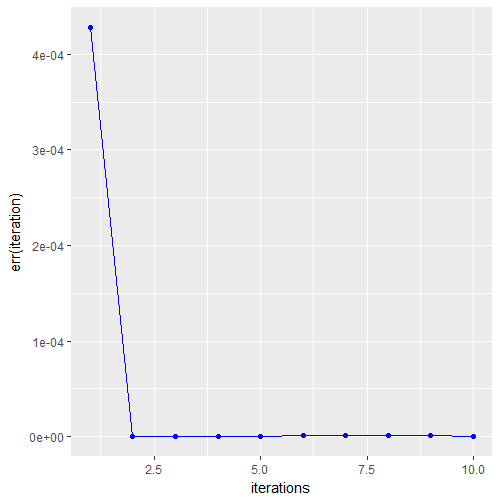}
\caption{Hepatitis-C}
\end{subfigure}
\caption{Convergence of matrix T on different data sets.}
\label{figure:convergence_T}
\end{figure}

\subsection{Convergence Analysis}\label{convergence analysis}
Since features are finally selected based on matrix T, we analyze changes of matrix T, along with the value of the objective function through iterations in order to study the convergence of algorithm \ref{algorithm:main}. Let 

\begin{equation}\label{eq:err}
    err(i) = ||T^{(i)} - T^{(i-1)}||_F^2~/~(d \times k)
\end{equation}

Where $T^{(i)}$ is the value of T after outer iteration number i and $(d \times k)$ is the number of elements of T. err(i) calculates the normalized change of T after an iteration. Generally, both T and the objective function converge to a global minimum after two iterations. See figures \ref{figure:convergence_T}, \ref{figure:convergence_obj}.

\begin{figure}[!htbp]
\centering
\begin{subfigure}[b]{0.325\textwidth}
\includegraphics[scale=0.42]{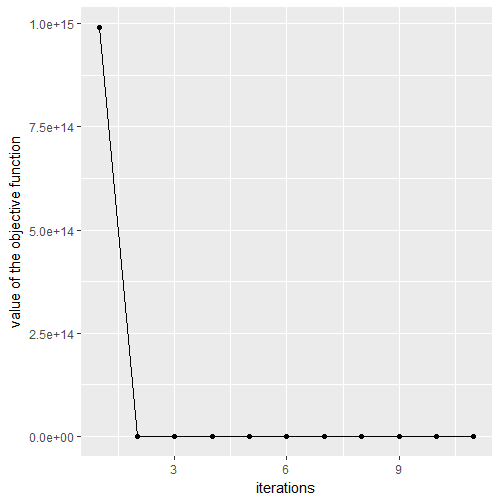}
\caption{Gene-Expression}
\end{subfigure}
\begin{subfigure}[b]{0.325\textwidth}
\includegraphics[scale=0.42]{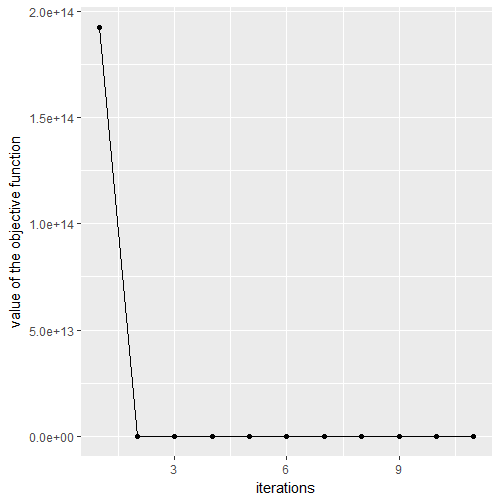}
\caption{Smoke-Cancer}
\end{subfigure}
\begin{subfigure}[b]{0.325\textwidth}
\includegraphics[scale=0.42]{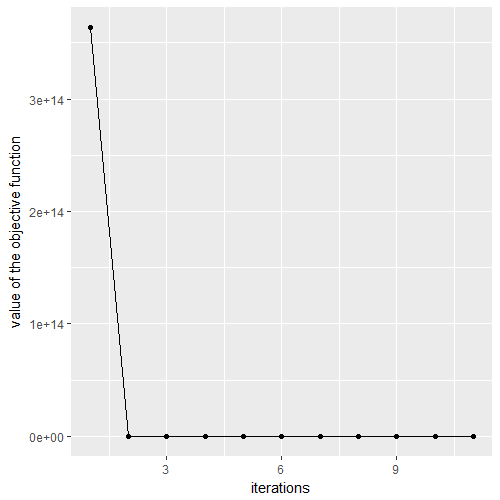}
\caption{Various-Cancers}
\end{subfigure}
\end{figure}
\begin{figure}[!htbp]
\ContinuedFloat
\centering
\begin{subfigure}[b]{0.325\textwidth}
\includegraphics[scale=0.42]{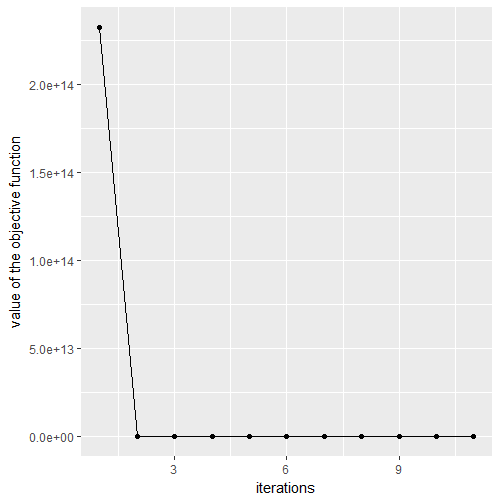}
\caption{Burkitt-Lymphoma}
\end{subfigure}
\begin{subfigure}[b]{0.325\textwidth}
\includegraphics[scale=0.42]{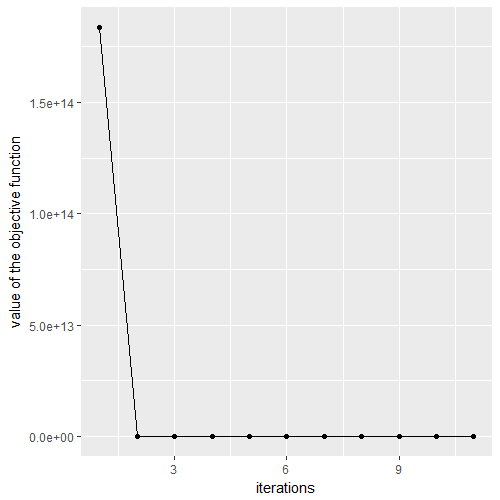}
\caption{Mouse-Type}
\end{subfigure}
\begin{subfigure}[b]{0.325\textwidth}
\includegraphics[scale=0.42]{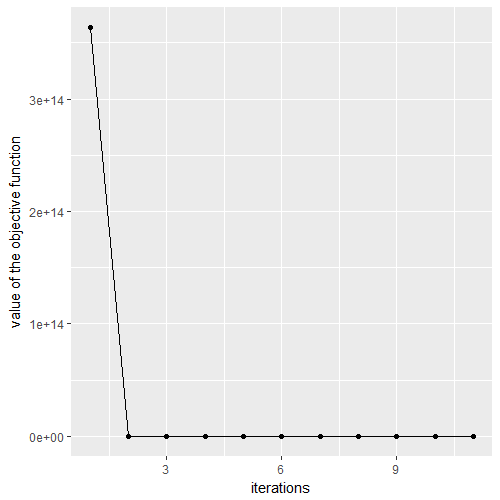}
\caption{Hepatitis-C}
\end{subfigure}
\caption{Convergence of the proposed objective function on different data sets.}
\label{figure:convergence_obj}
\end{figure}

\subsection{Effect of weighting points}\label{effect of weighting points}
\begin{figure}[!htbp]
\centering
\begin{subfigure}[b]{0.325\textwidth}
\includegraphics[scale=0.42]{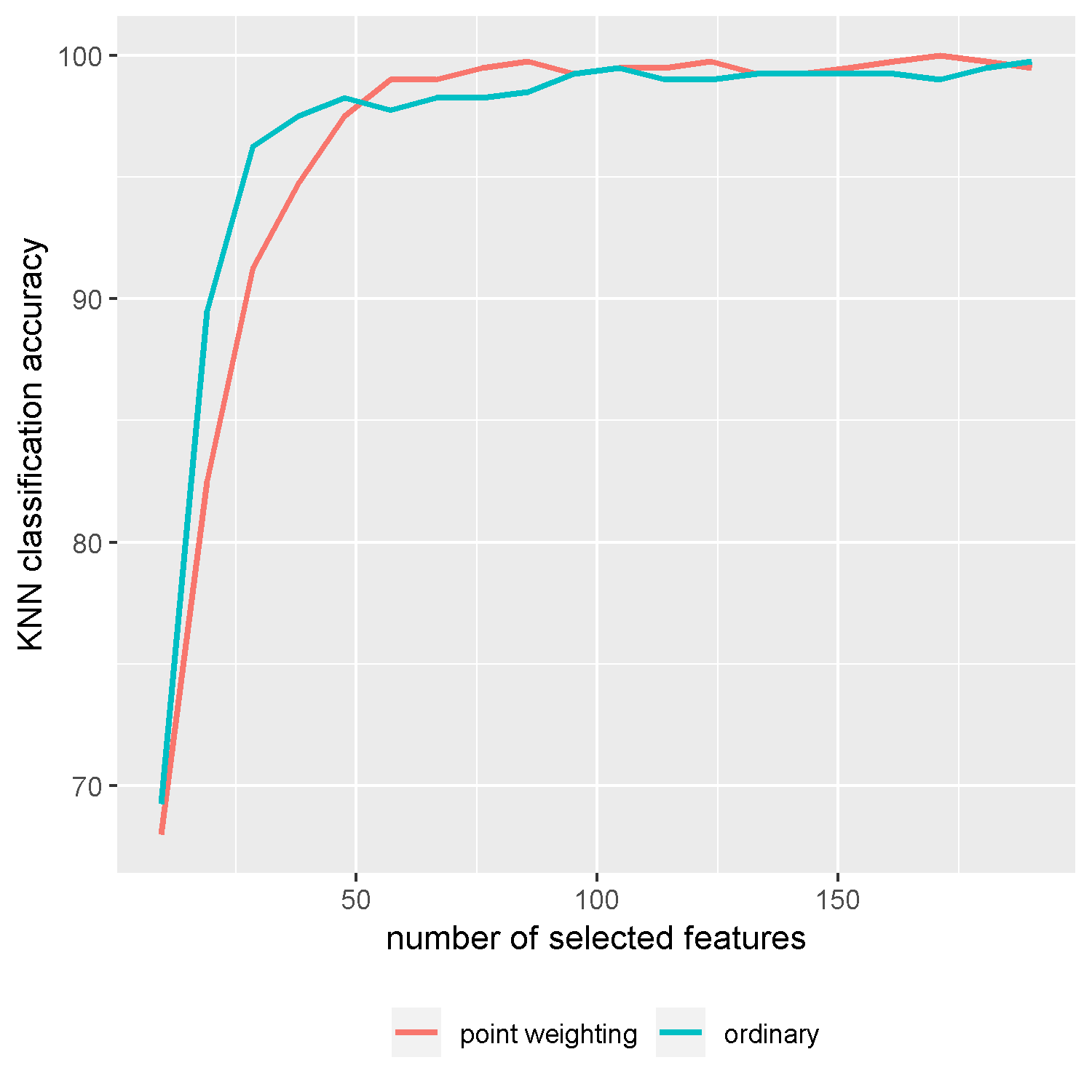}
\caption{Gene-Expression}
\end{subfigure}
\begin{subfigure}[b]{0.325\textwidth}
\includegraphics[scale=0.42]{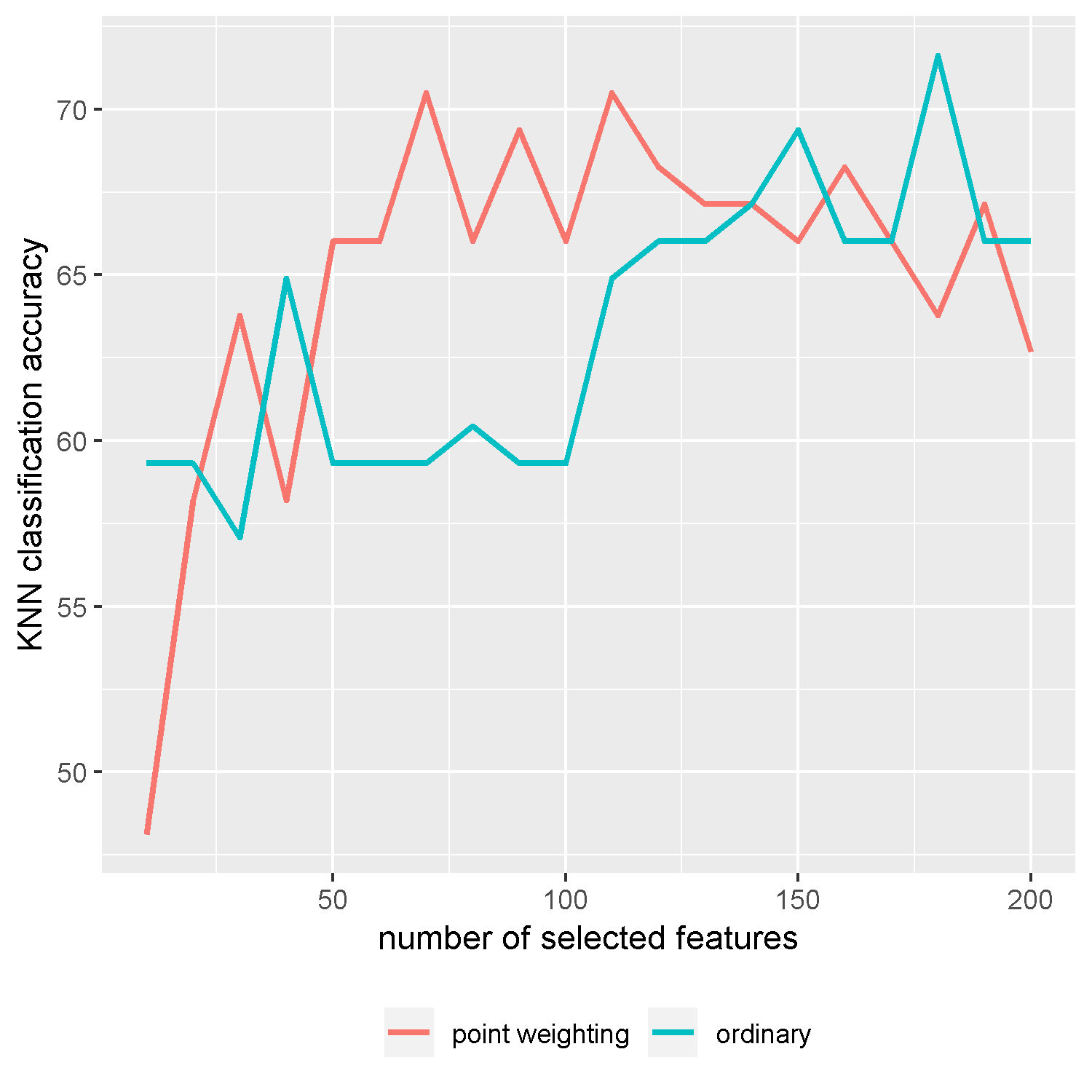}
\caption{Smoke-Cancer}
\end{subfigure}
\begin{subfigure}[b]{0.325\textwidth}
\includegraphics[scale=0.42]{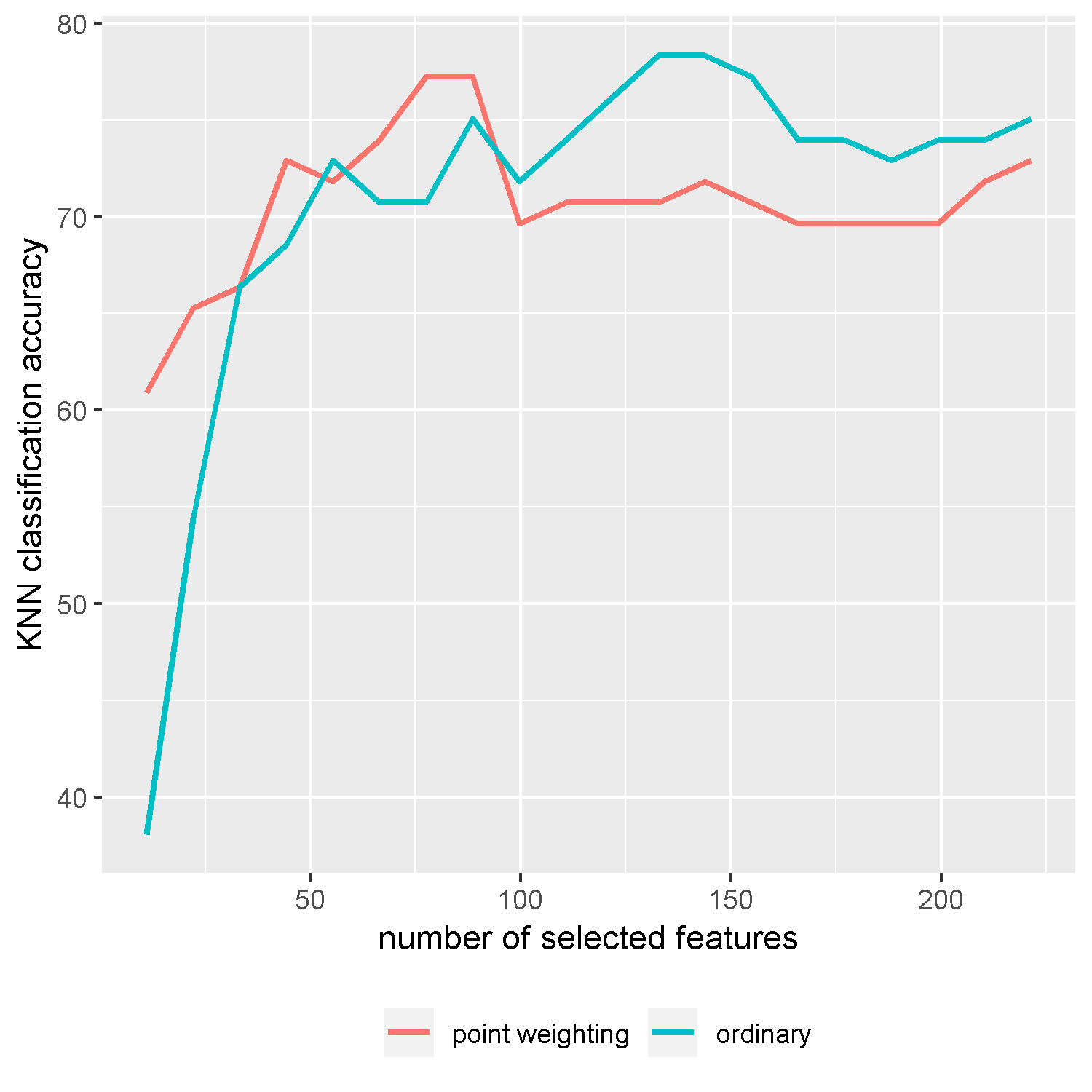}
\caption{Mouse-Type}
\end{subfigure}
\caption{KNN classification accuracies of different methods on benchmark data sets.}
\label{figure:non-point-weighting-knn}
\end{figure}

\begin{figure}[!htbp]
\centering
\begin{subfigure}[b]{0.325\textwidth}
\includegraphics[scale=0.42]{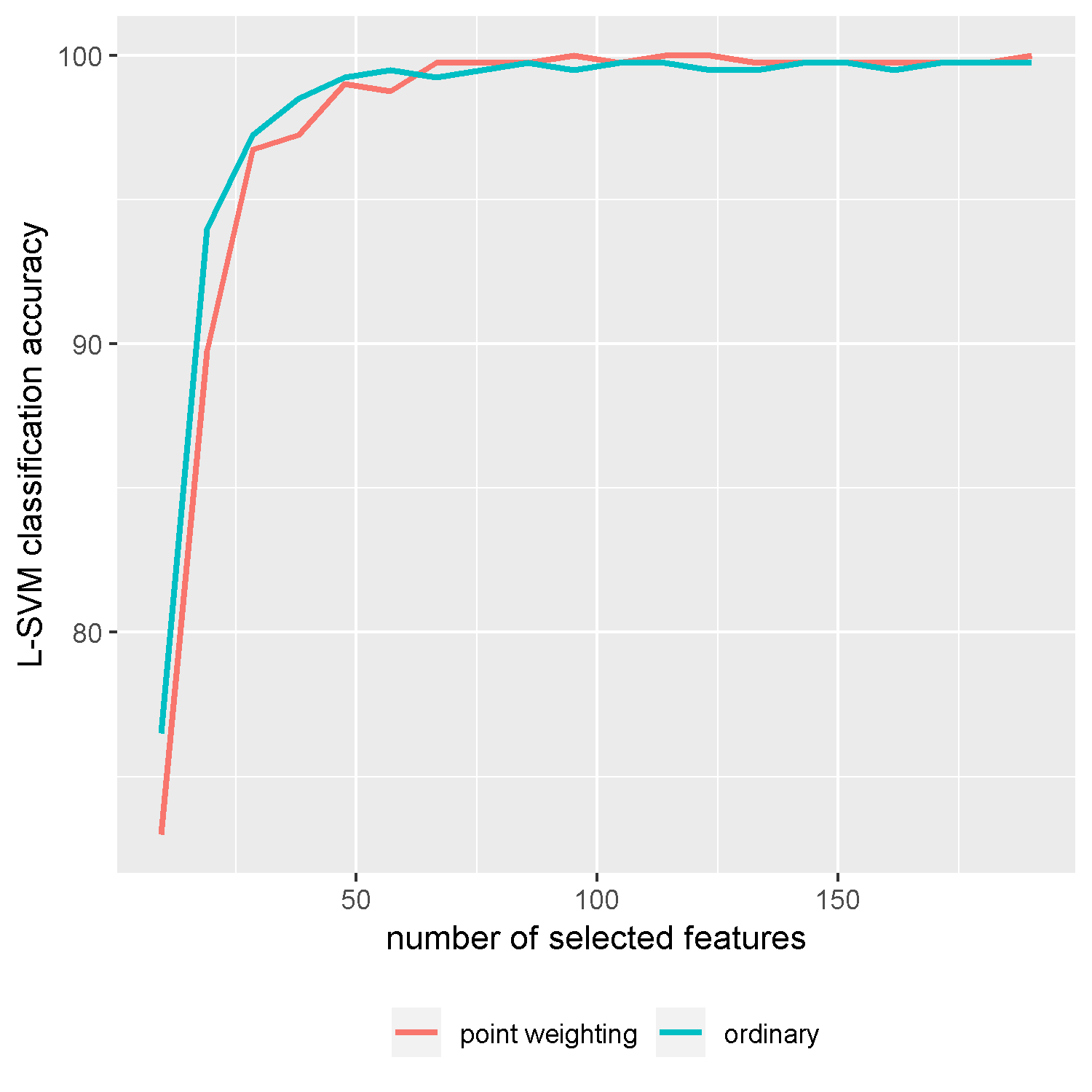}
\caption{Gene-Expression}
\end{subfigure}
\begin{subfigure}[b]{0.325\textwidth}
\includegraphics[scale=0.42]{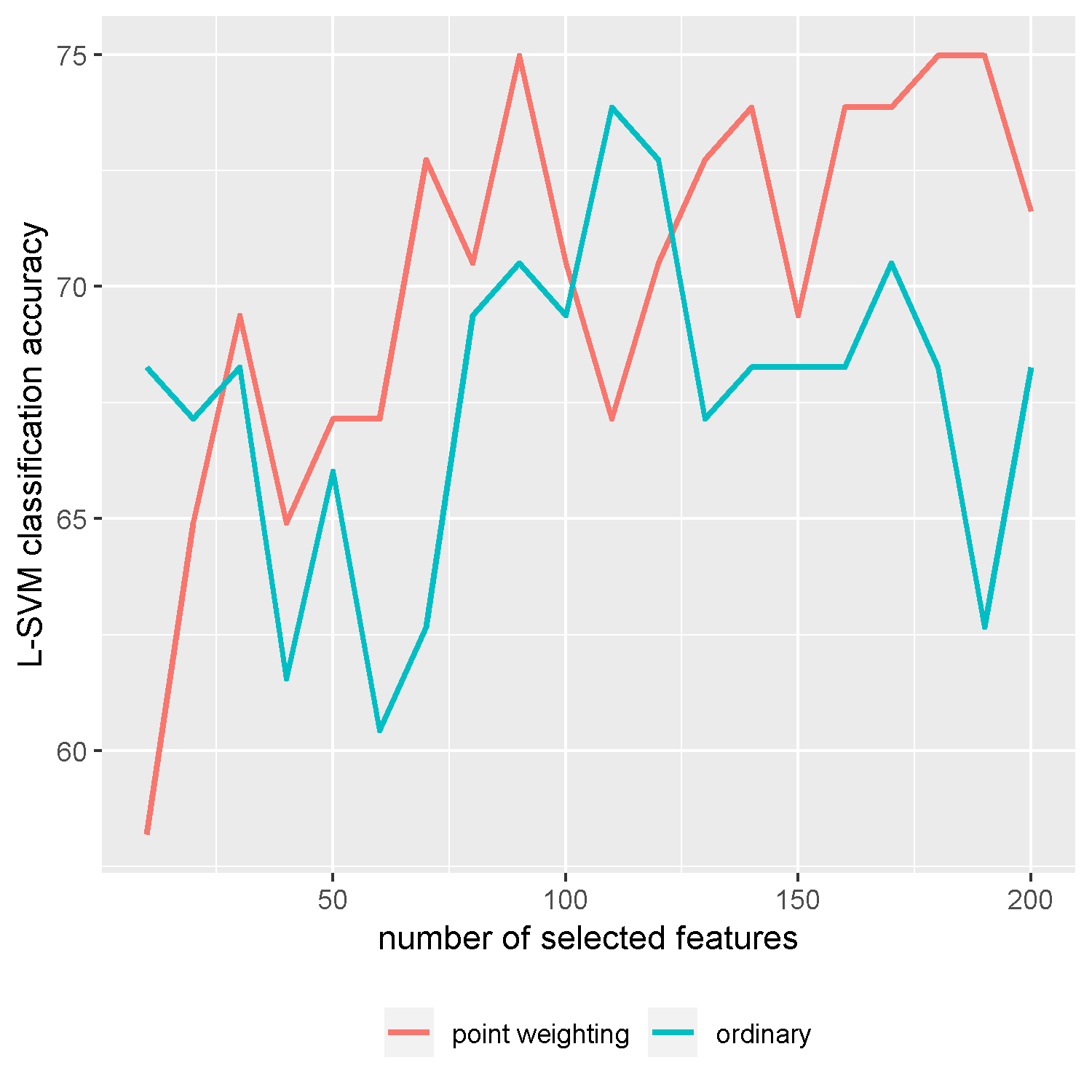}
\caption{Smoke-Cancer}
\end{subfigure}
\begin{subfigure}[b]{0.325\textwidth}
\includegraphics[scale=0.42]{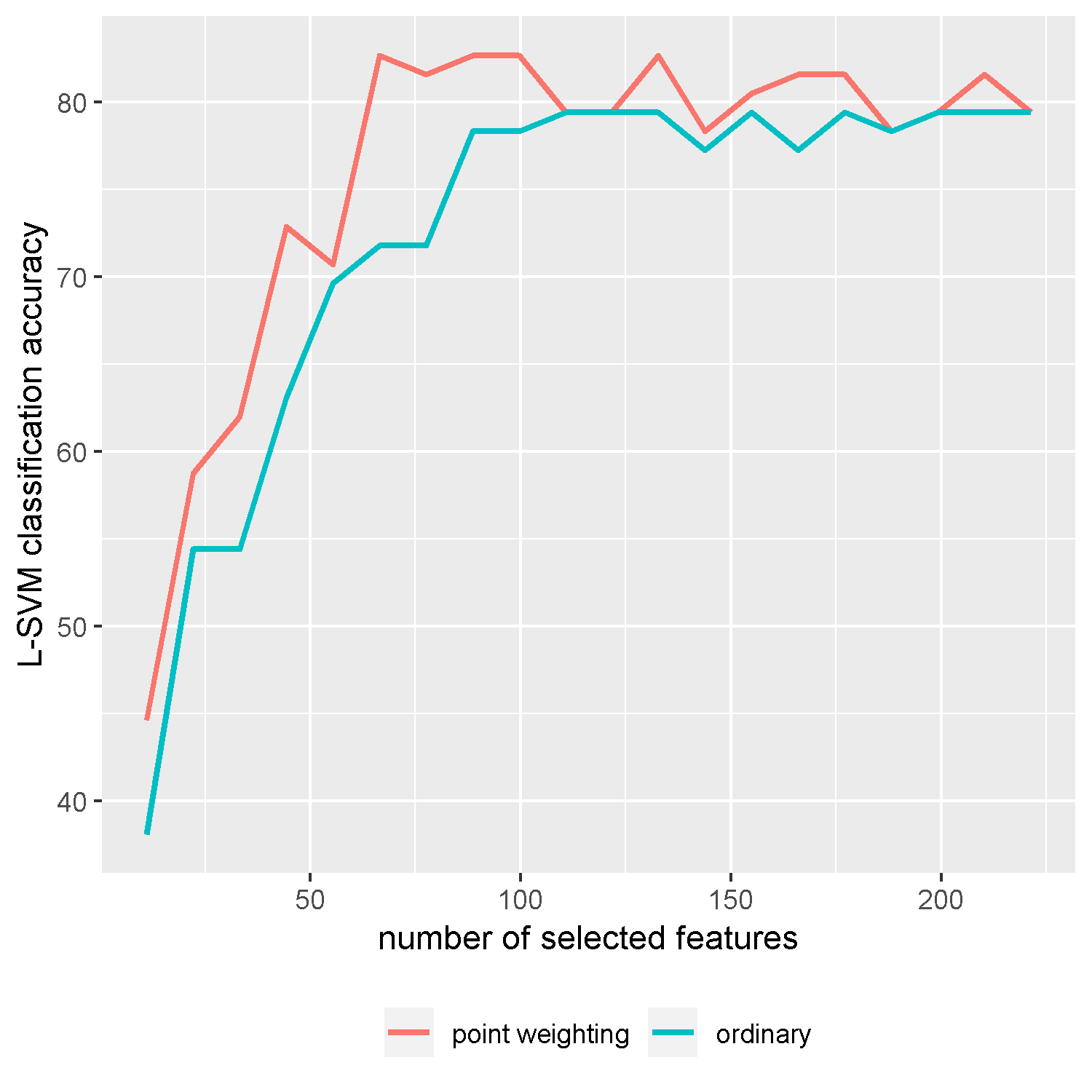}
\caption{Mouse-Type}
\end{subfigure}
\caption{L-SVM classification accuracies of different methods on benchmark data sets.}
\label{figure:non-point-weighting-lsvm}
\end{figure}

\begin{figure}[!htbp]
\centering
\begin{subfigure}[b]{0.325\textwidth}
\includegraphics[scale=0.42]{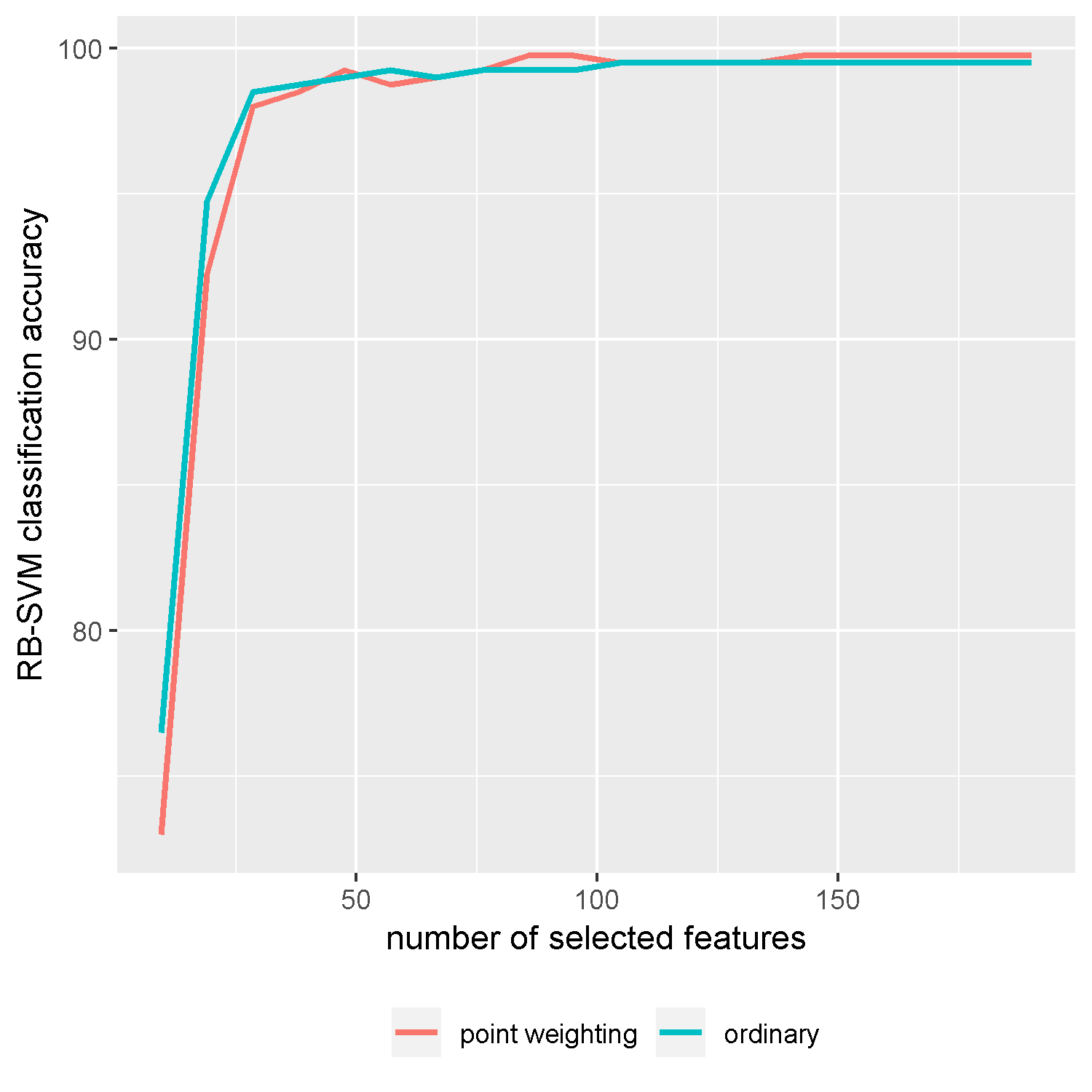}
\caption{Gene-Expression}
\end{subfigure}
\begin{subfigure}[b]{0.325\textwidth}
\includegraphics[scale=0.42]{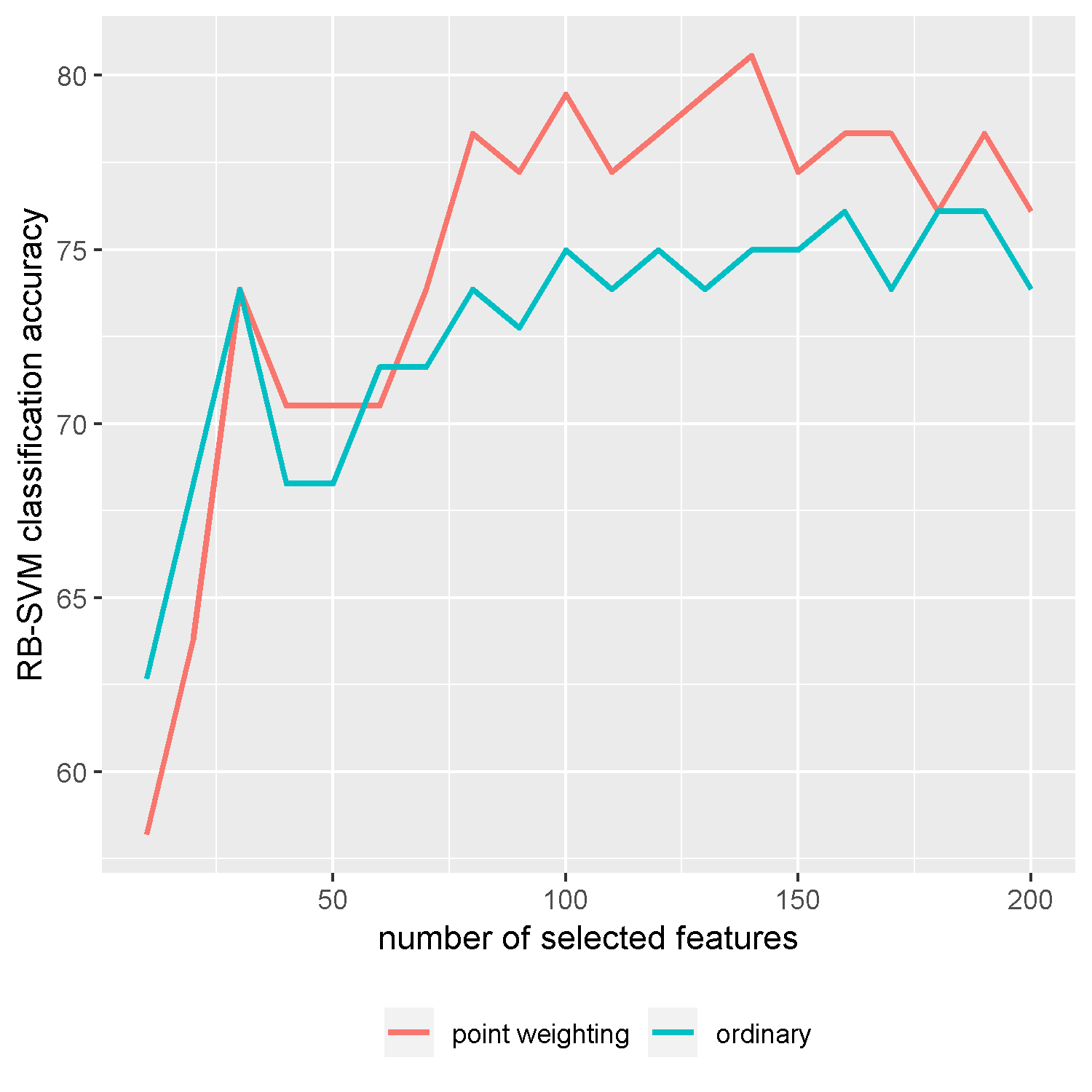}
\caption{Smoke-Cancer}
\end{subfigure}
\begin{subfigure}[b]{0.325\textwidth}
\includegraphics[scale=0.42]{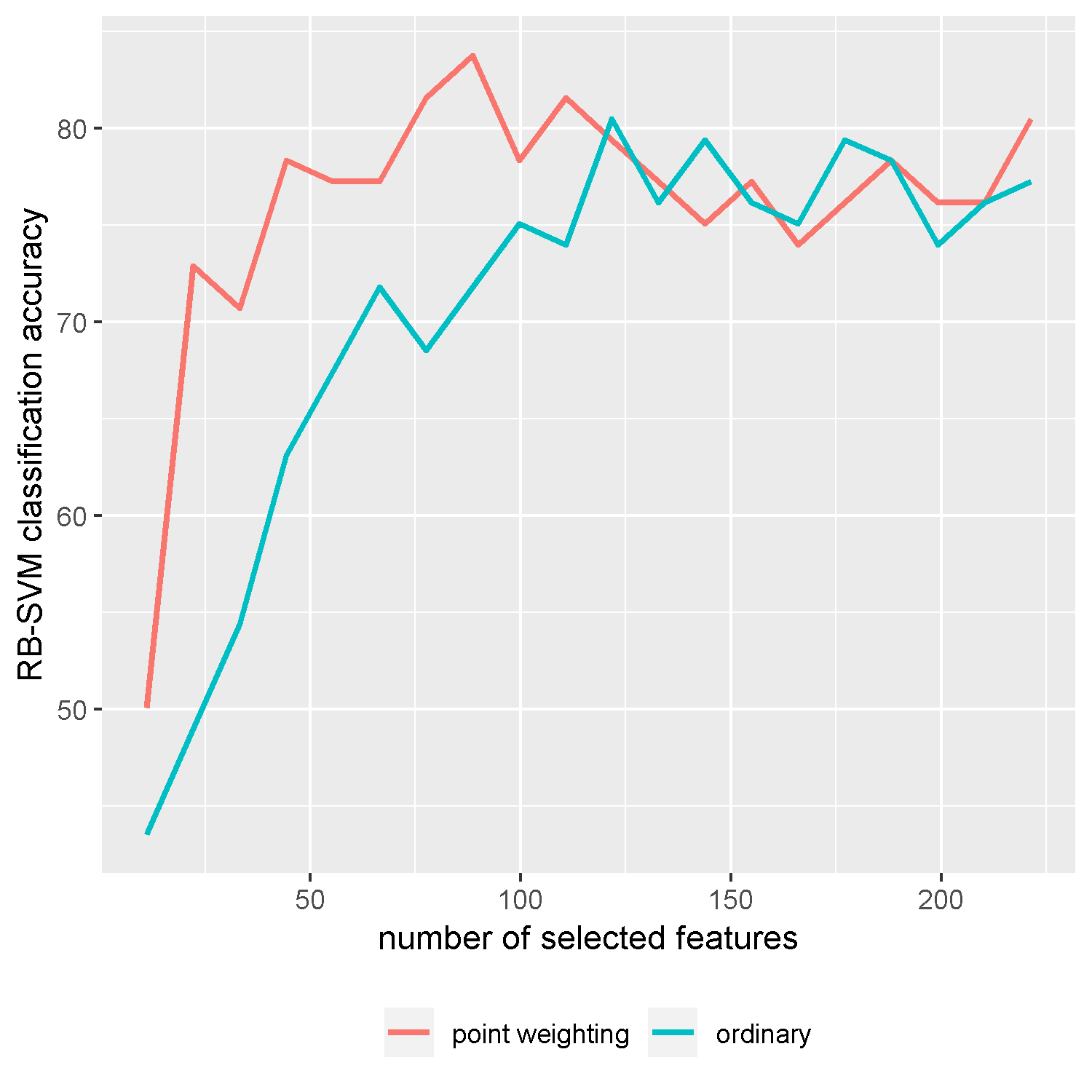}
\caption{Mouse-Type}
\end{subfigure}
\caption{RB-SVM classification accuracies of different methods on benchmark data sets.}
\label{figure:non-point-weighting-rbsvm}
\end{figure}

\begin{figure}[!htbp]
\centering
\begin{subfigure}[b]{0.325\textwidth}
\includegraphics[scale=0.42]{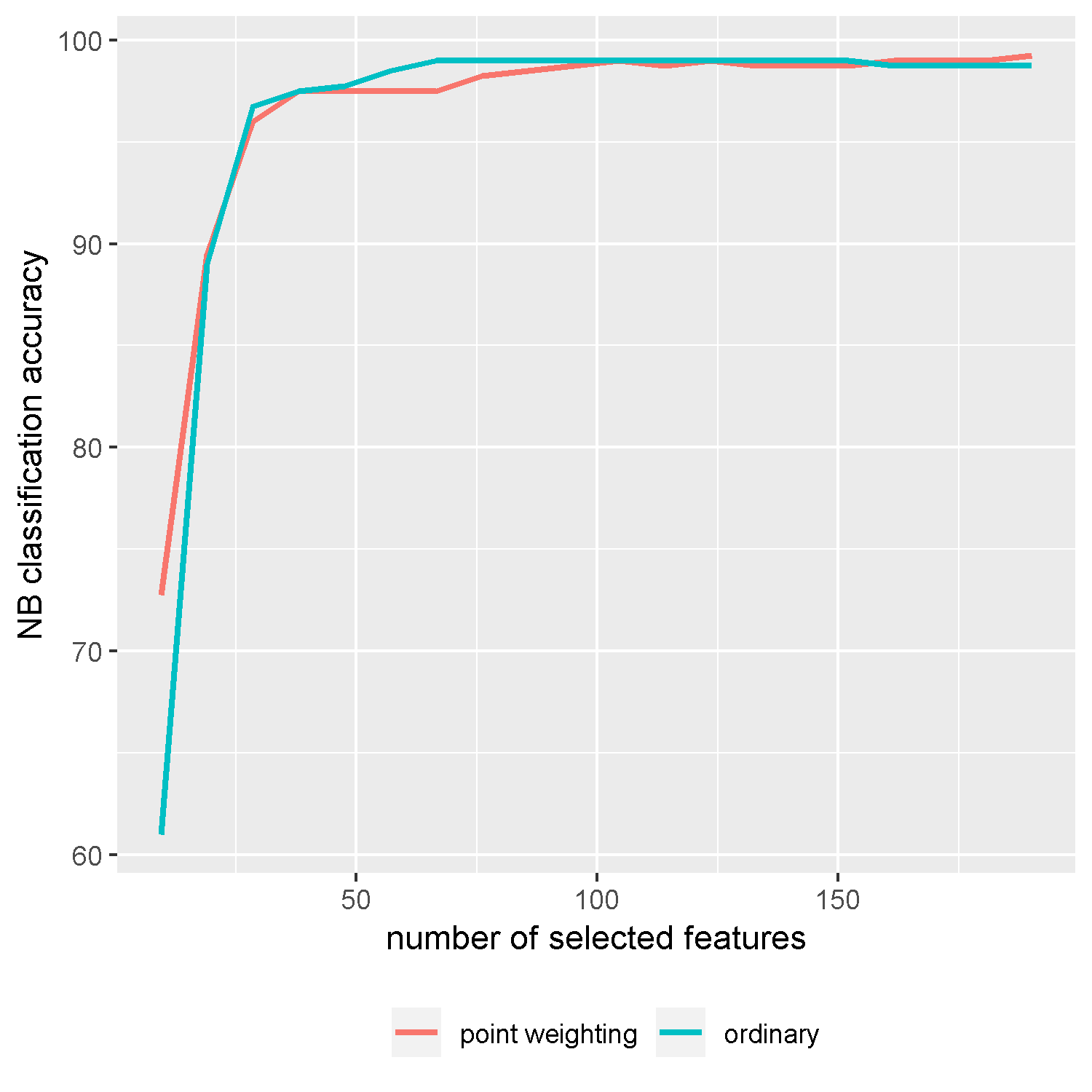}
\caption{Gene-Expression}
\end{subfigure}
\begin{subfigure}[b]{0.325\textwidth}
\includegraphics[scale=0.42]{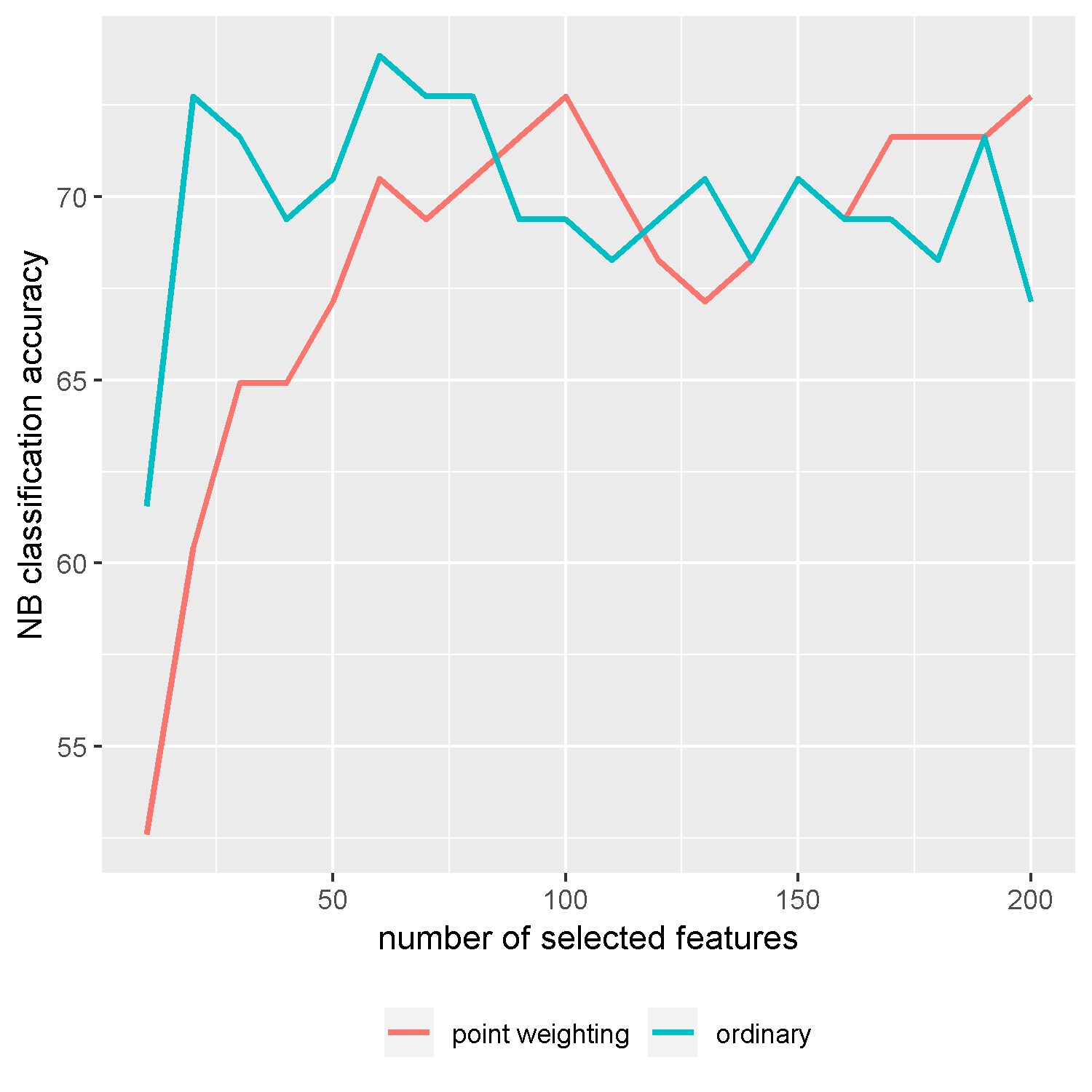}
\caption{Smoke-Cancer}
\end{subfigure}
\begin{subfigure}[b]{0.325\textwidth}
\includegraphics[scale=0.42]{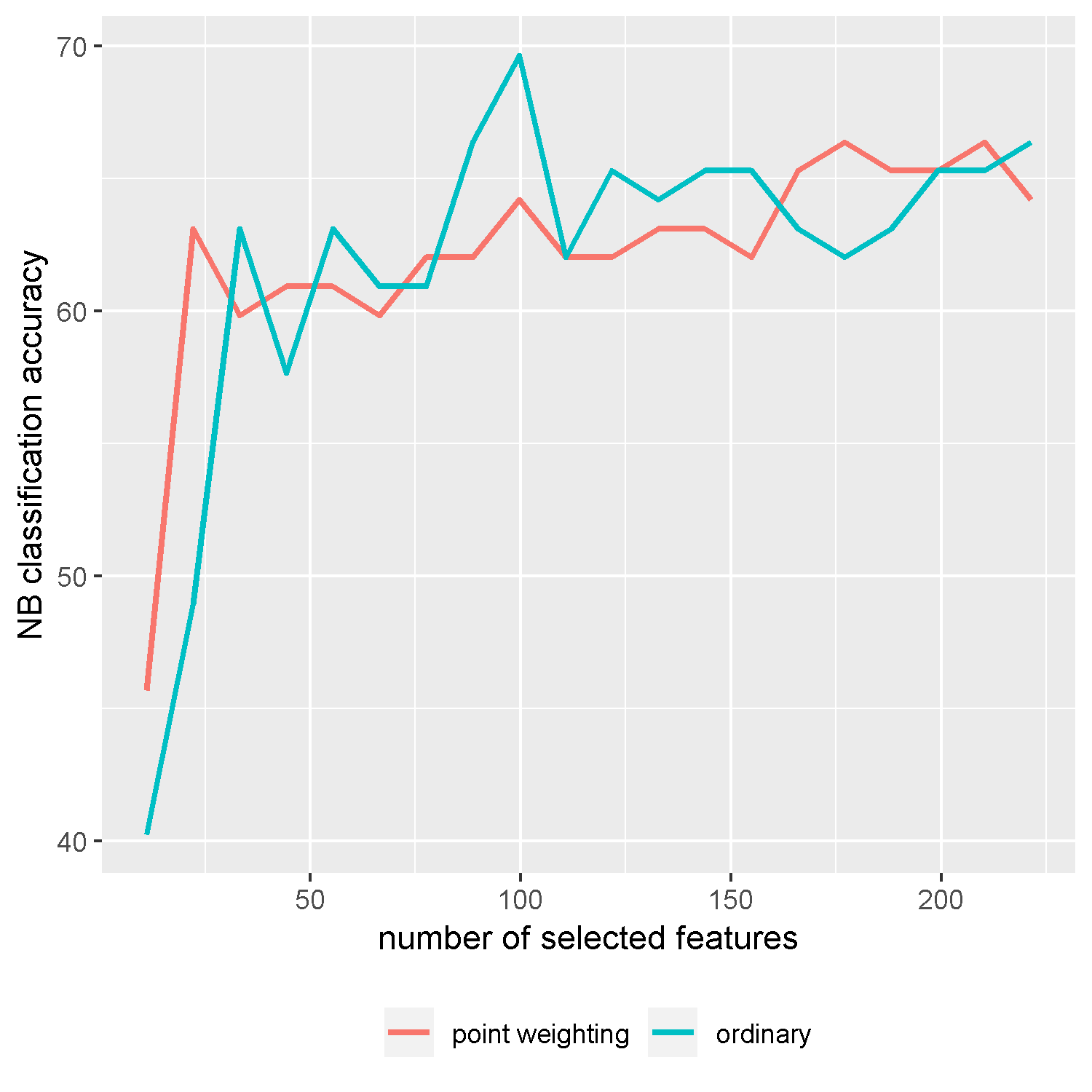}
\caption{Mouse-Type}
\end{subfigure}
\caption{NB classification accuracies of different methods on benchmark data sets.}
\label{figure:non-point-weighting-nb}
\end{figure}

To show the importance of weighting points of the proposed framework we conducted some experiments. We implemented the ordinary version of our point-weighting framework. In this new framework, all data points are treated equally. For this purpose, we replaced matrix D with the identity matrix. We also changed the soft hypergraph construction method. The new hypergraph was constructed by connecting each data point to its $l$ nearest neighbors. This soft hypergraph is widely used in recent studies \cite{HSIR, HLasso}.
This hypergraph contains n vertices and n hyperedges; While the hypergraph of our proposed method contains only m vertices and m hyperedges; Where $m \ll n$. Although our point-weighting framework has much lower computational complexity than the ordinary version, conducted experiments support its effectiveness by achieving higher classification accuracy on different benchmark datasets using different evaluation methods, in most cases. See figures \ref{figure:non-point-weighting-knn}, \ref{figure:non-point-weighting-lsvm}, \ref{figure:non-point-weighting-rbsvm}, and \ref{figure:non-point-weighting-nb}. These experiments exhibit the importance of behaving different data points based on their representation power and role in defining the data structure.

\subsection{Effect of soft hypergraph}\label{effect of soft hypergraph}
\begin{figure}[!htbp]
\centering
\begin{subfigure}[b]{0.325\textwidth}
\includegraphics[scale=0.42]{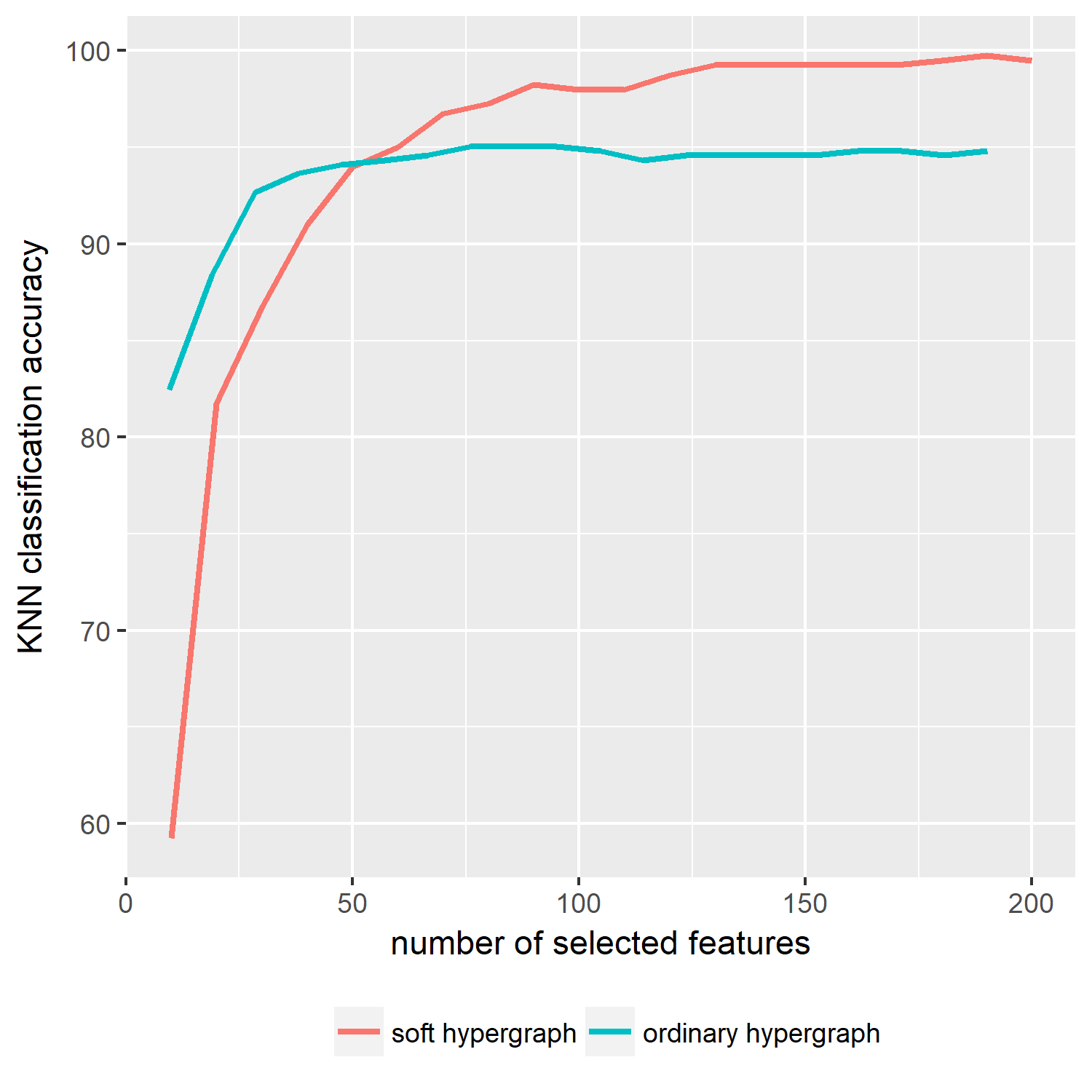}
\caption{Gene-Expression}
\end{subfigure}
\begin{subfigure}[b]{0.325\textwidth}
\includegraphics[scale=0.42]{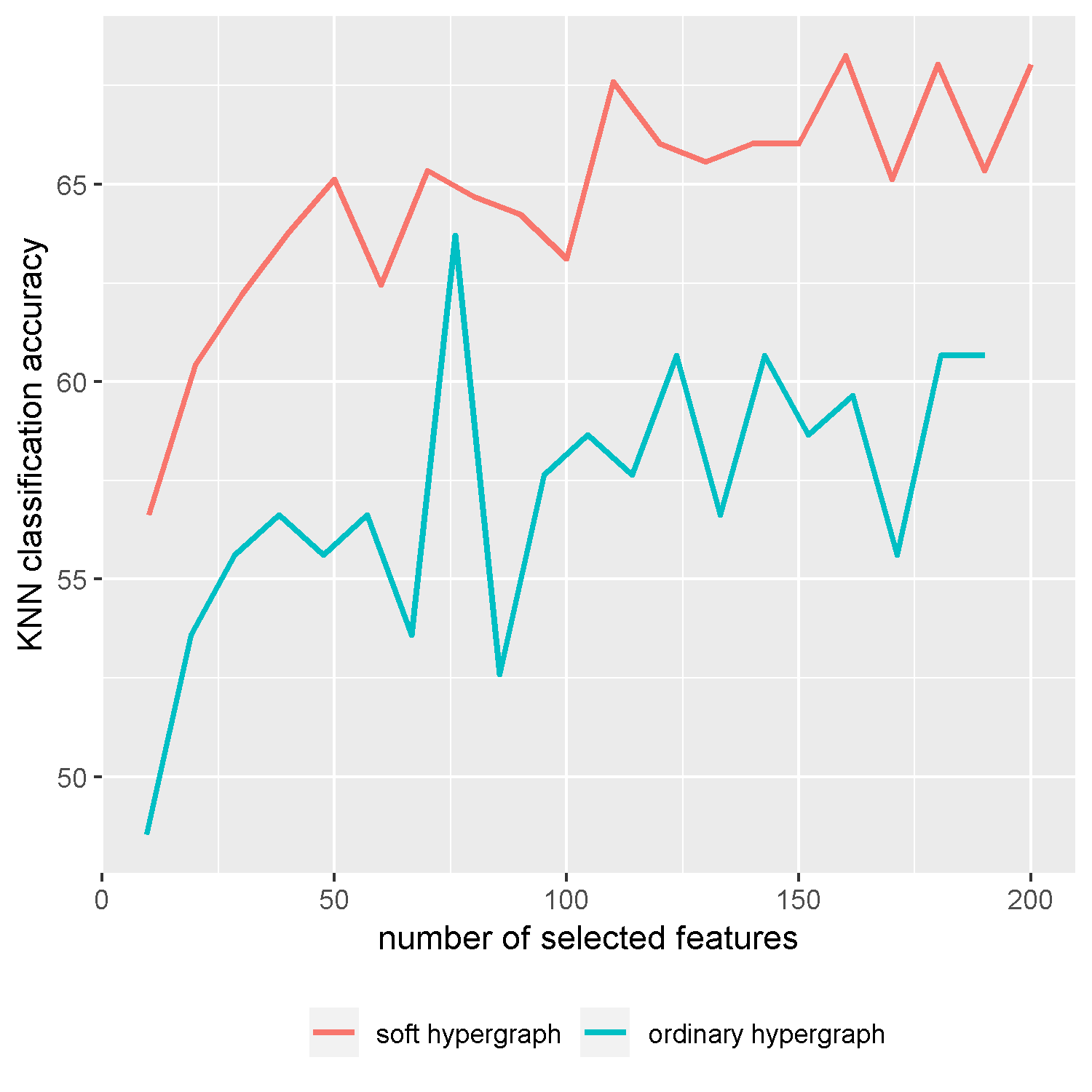}
\caption{Smoke-Cancer}
\end{subfigure}
\begin{subfigure}[b]{0.325\textwidth}
\includegraphics[scale=0.42]{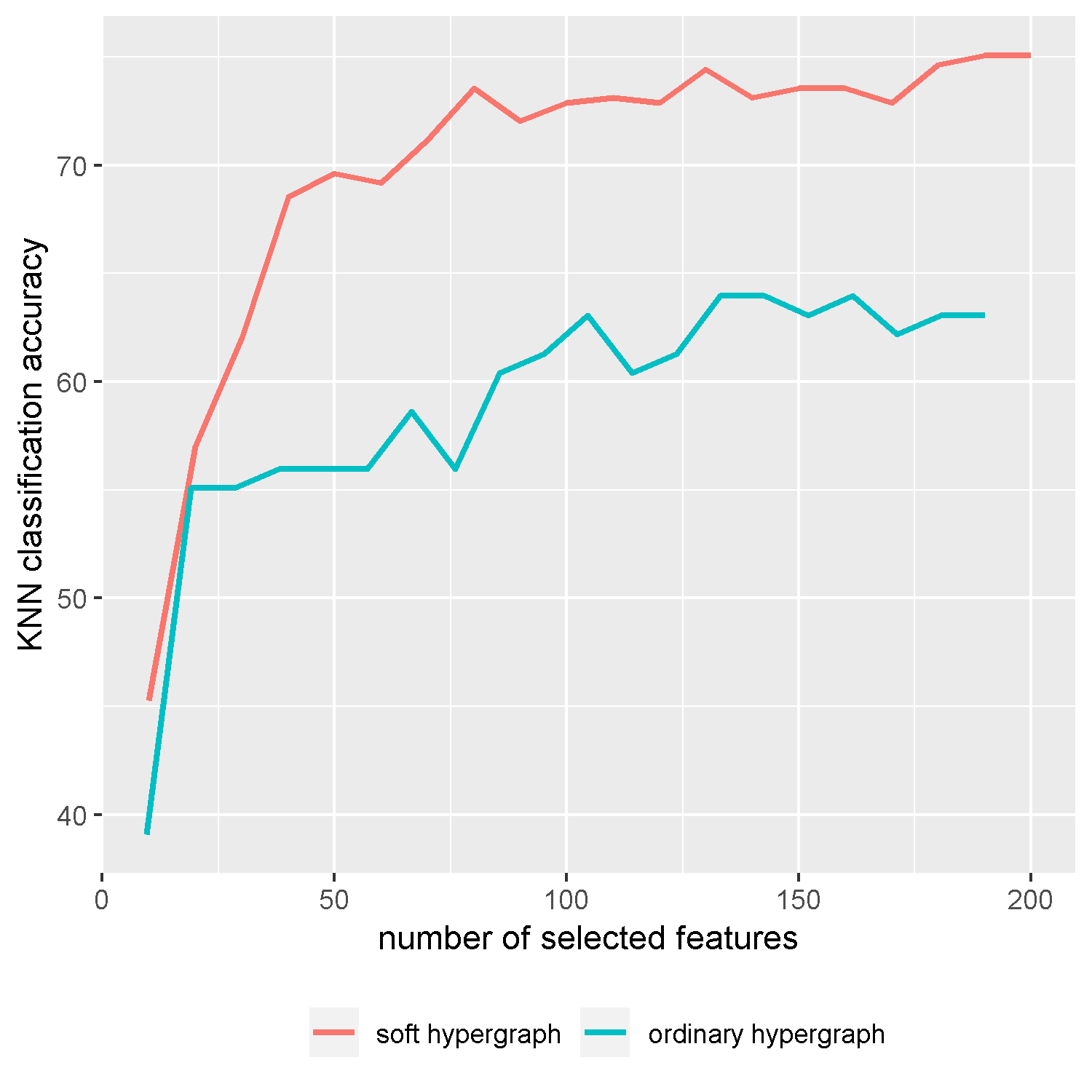}
\caption{Mouse-Type}
\end{subfigure}
\caption{KNN classification accuracies of different methods on benchmark data sets.}
\label{figure:ordinary-hypergraph-knn}
\end{figure}

\begin{figure}[!htbp]
\centering
\begin{subfigure}[b]{0.325\textwidth}
\includegraphics[scale=0.42]{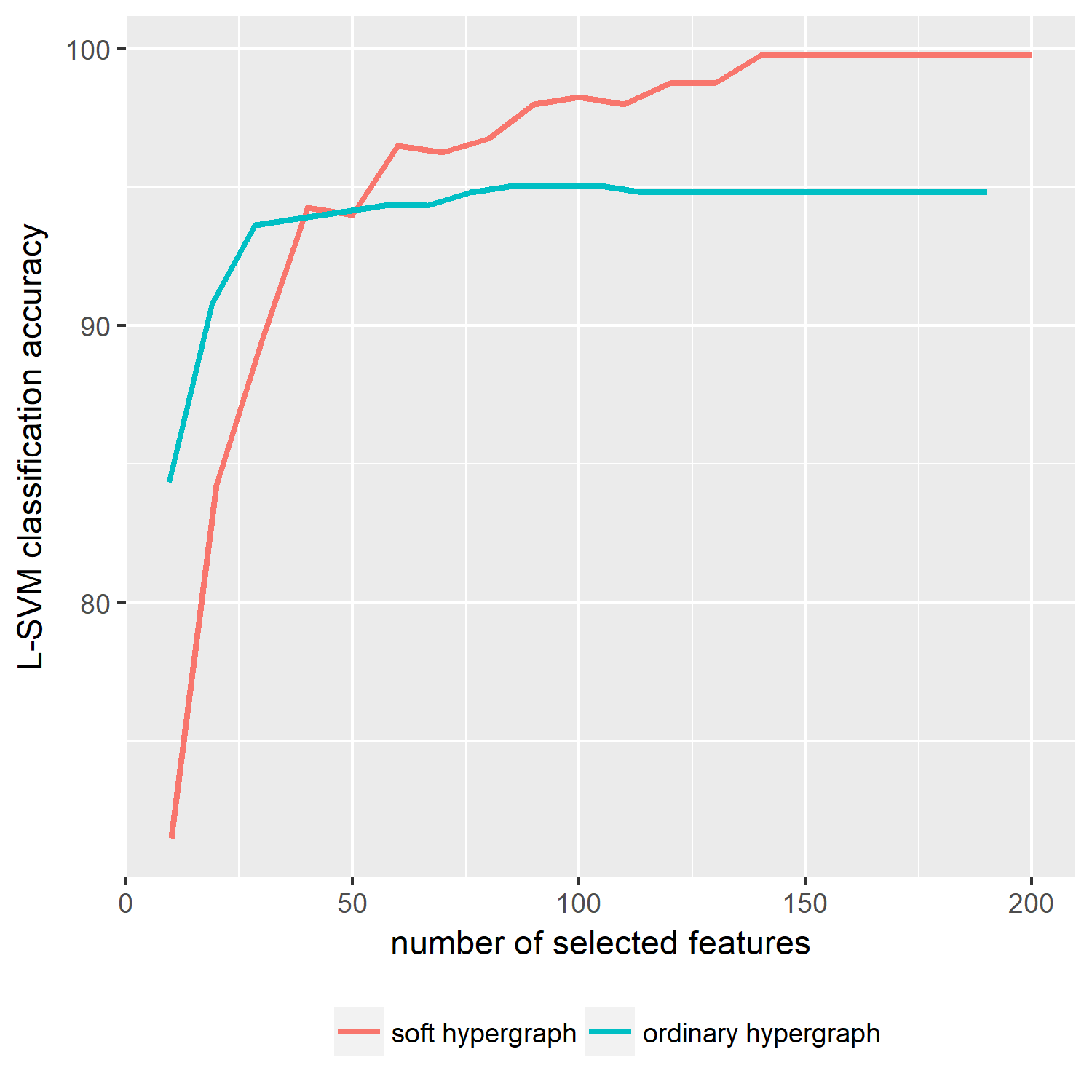}
\caption{Gene-Expression}
\end{subfigure}
\begin{subfigure}[b]{0.325\textwidth}
\includegraphics[scale=0.42]{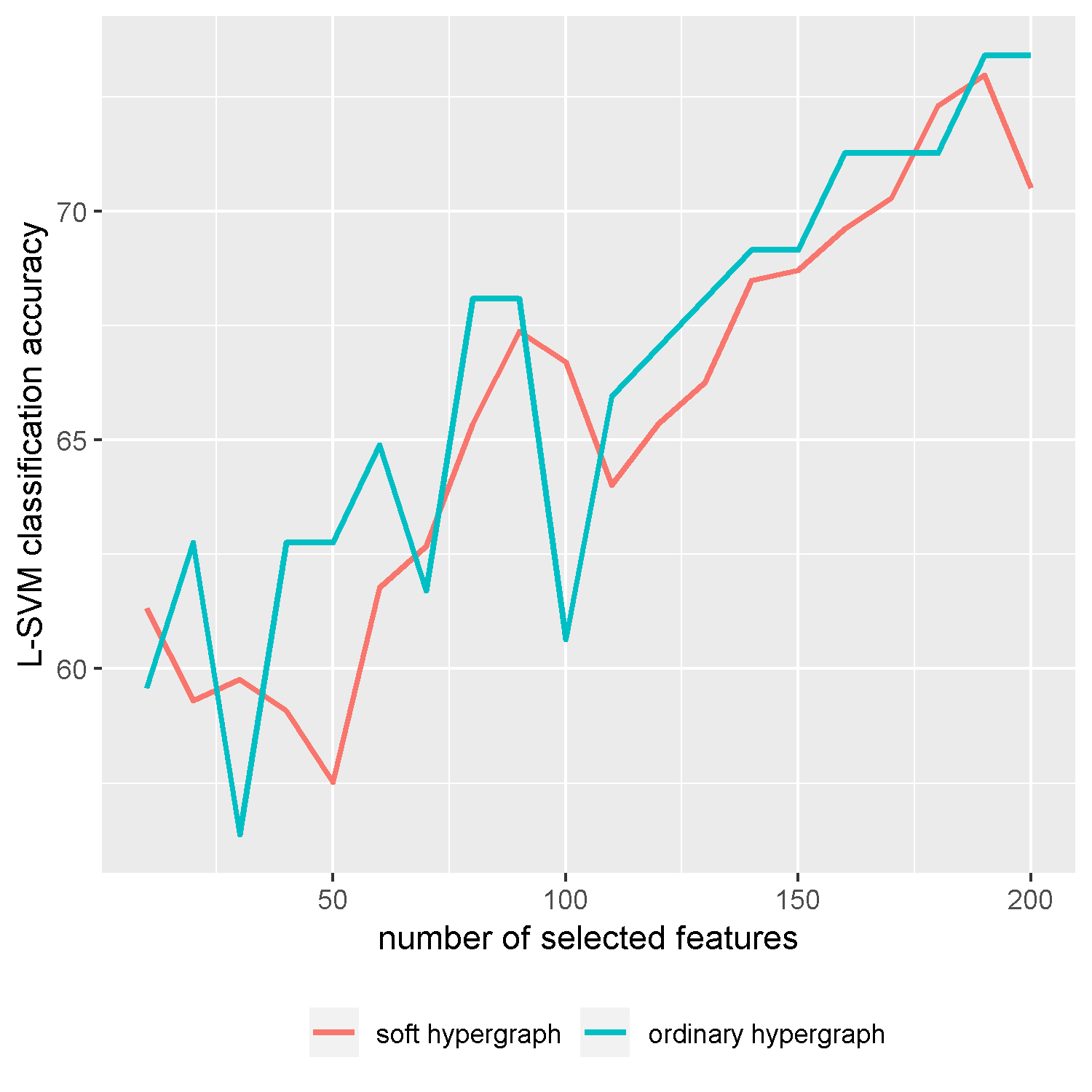}
\caption{Smoke-Cancer}
\end{subfigure}
\begin{subfigure}[b]{0.325\textwidth}
\includegraphics[scale=0.42]{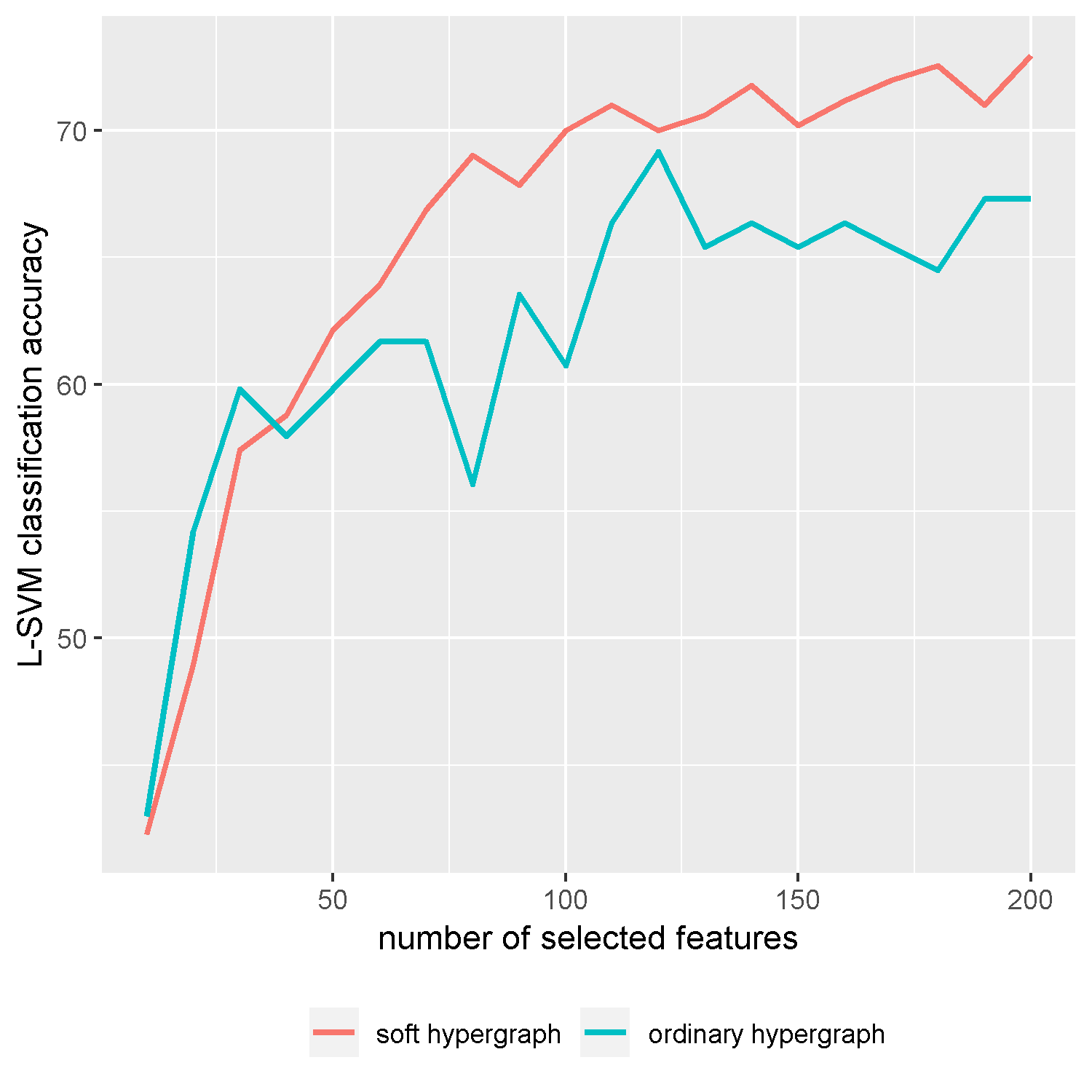}
\caption{Mouse-Type}
\end{subfigure}
\caption{L-SVM classification accuracies of different methods on benchmark data sets.}
\label{figure:ordinary-hypergraph-lsvm}
\end{figure}

\begin{figure}[!htbp]
\centering
\begin{subfigure}[b]{0.325\textwidth}
\includegraphics[scale=0.42]{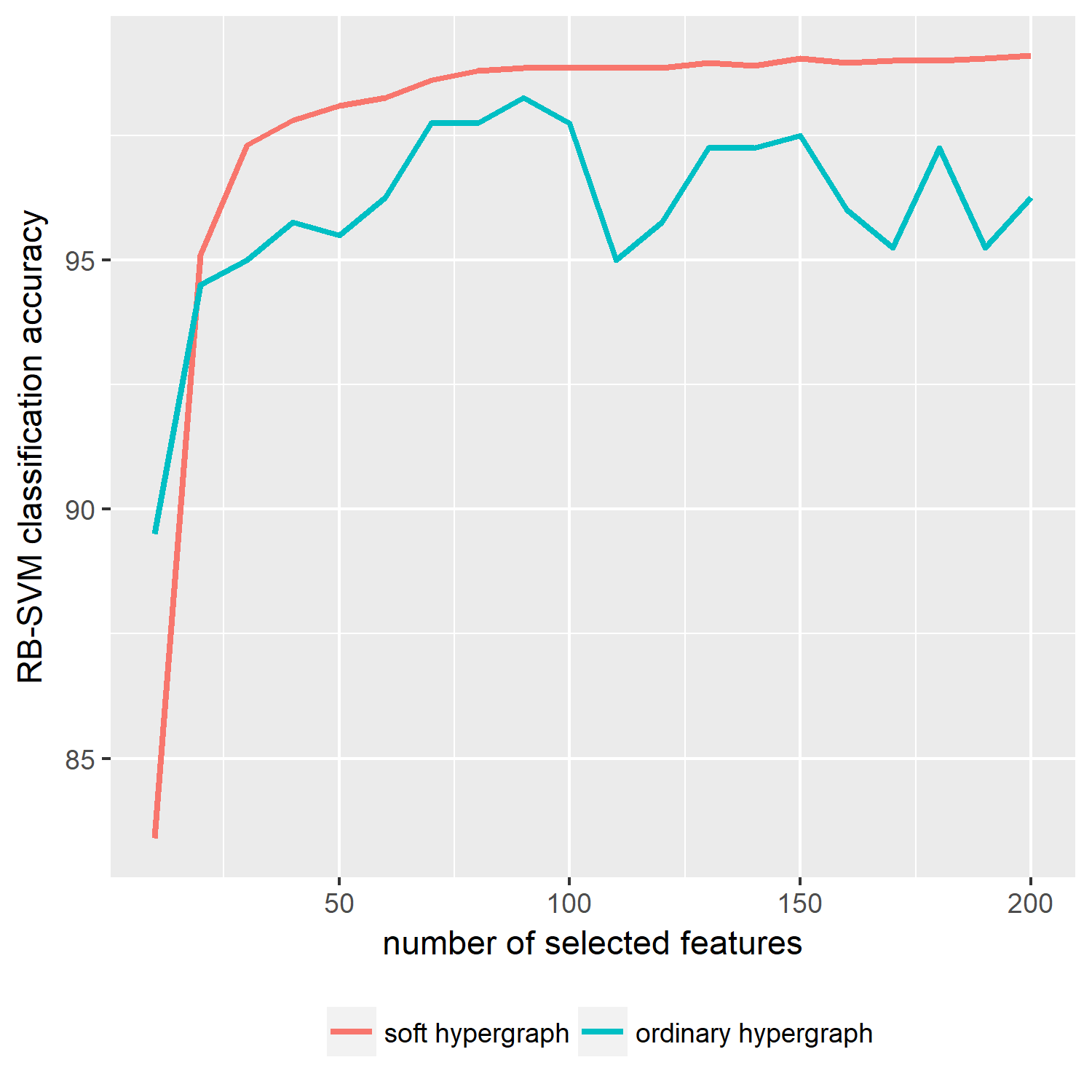}
\caption{Gene-Expression}
\end{subfigure}
\begin{subfigure}[b]{0.325\textwidth}
\includegraphics[scale=0.42]{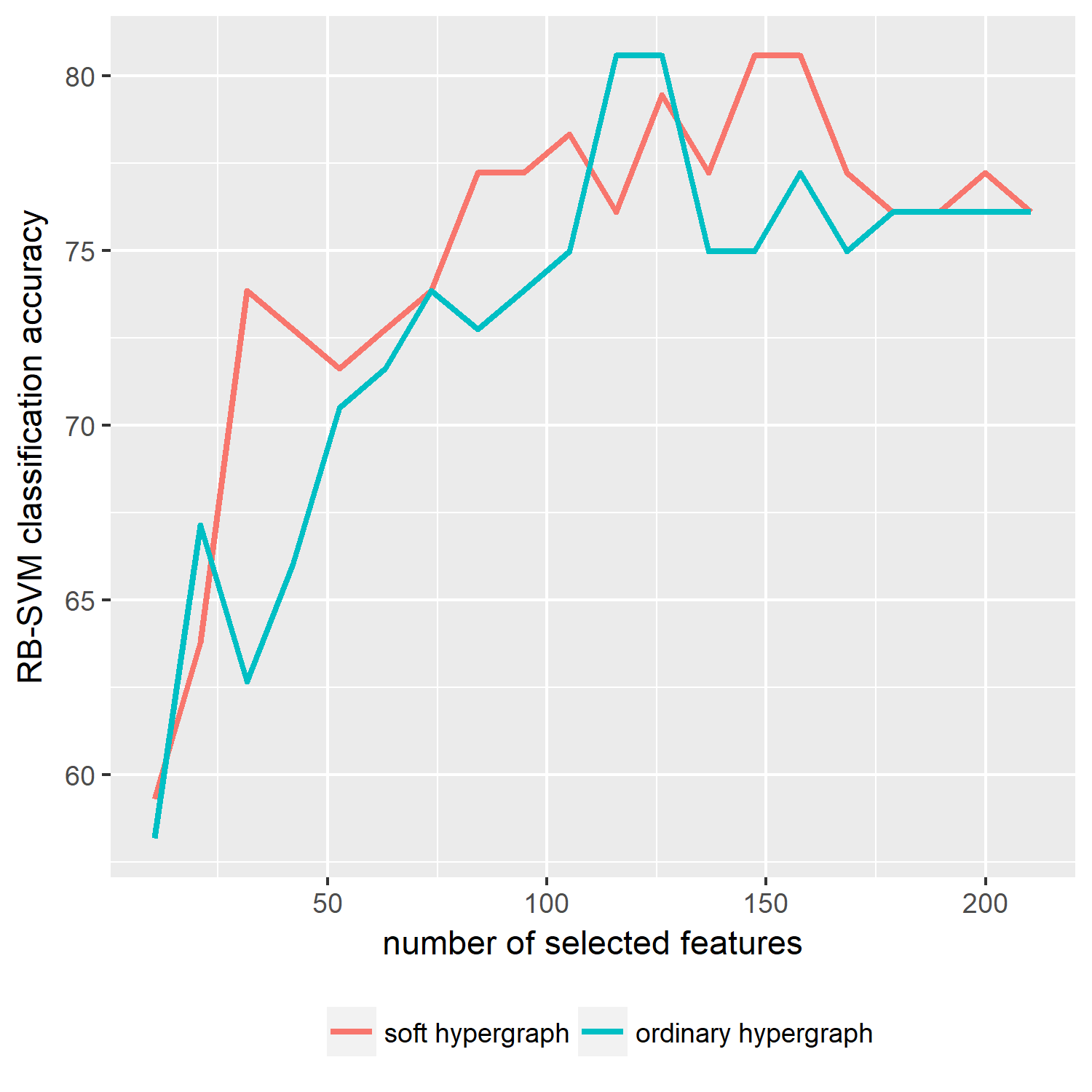}
\caption{Smoke-Cancer}
\end{subfigure}
\begin{subfigure}[b]{0.325\textwidth}
\includegraphics[scale=0.42]{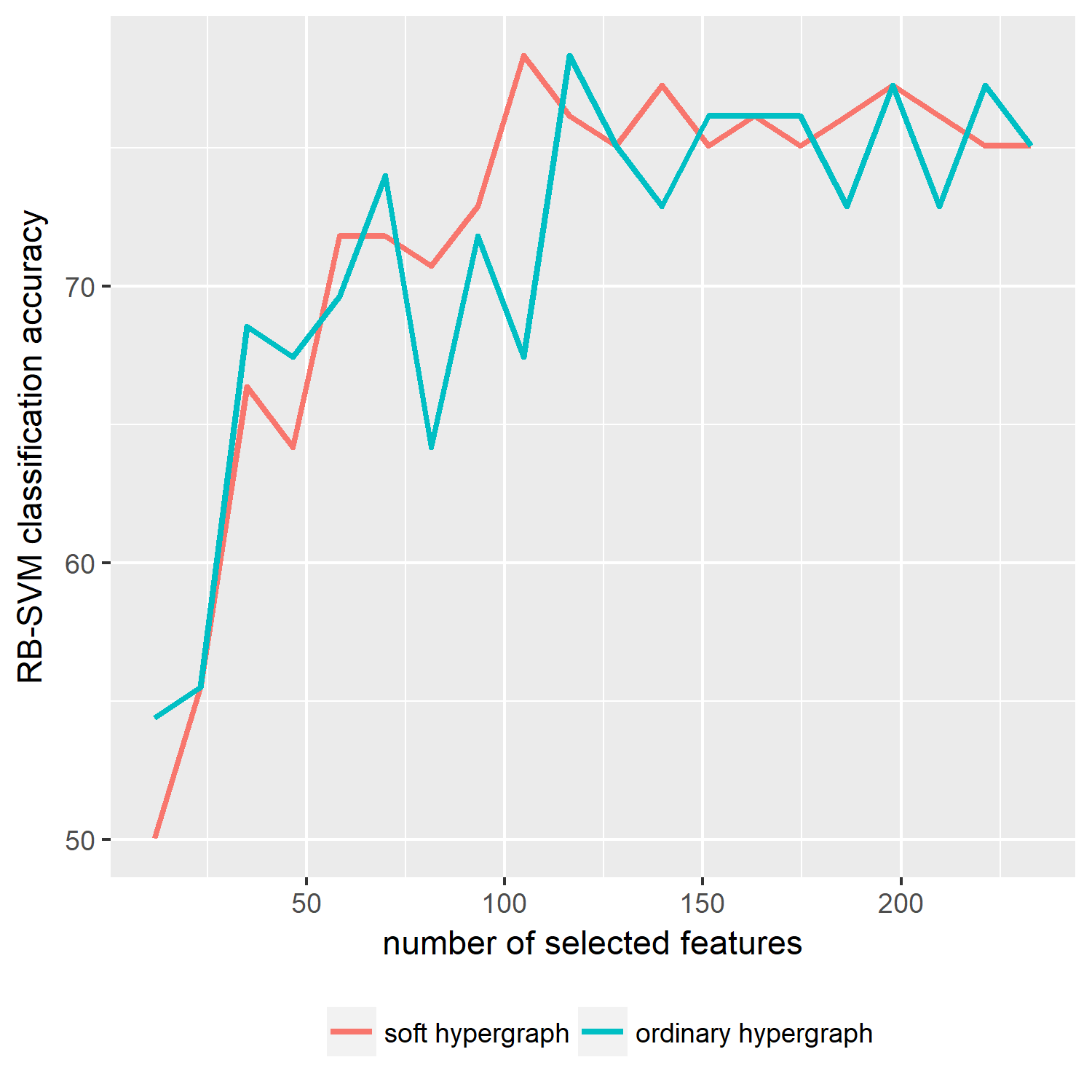}
\caption{Mouse-Type}
\end{subfigure}
\caption{RB-SVM classification accuracies of different methods on benchmark data sets.}
\label{figure:ordinary-hypergraph-rbsvm}
\end{figure}

\begin{figure}[!htbp]
\centering
\begin{subfigure}[b]{0.325\textwidth}
\includegraphics[scale=0.42]{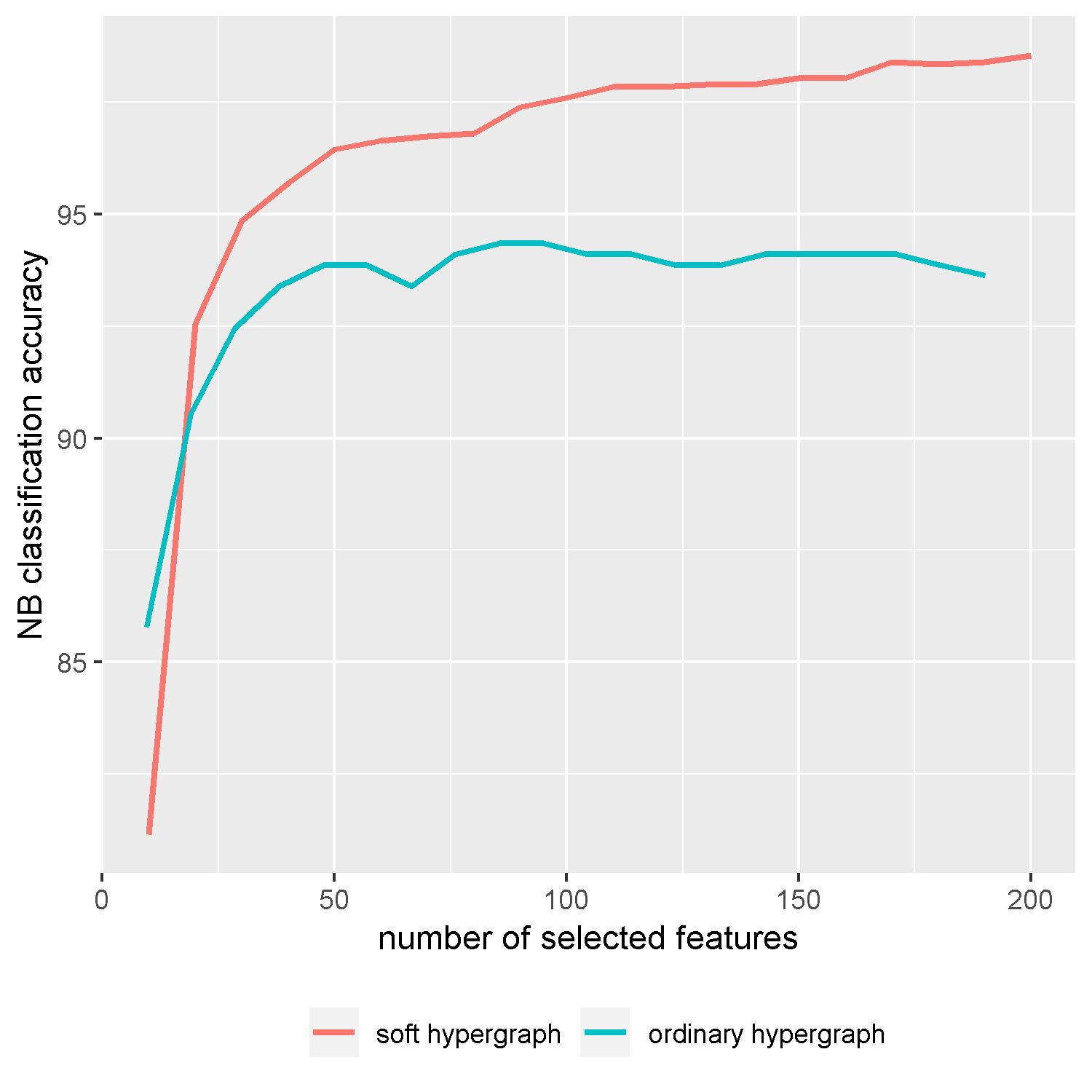}
\caption{Gene-Expression}
\end{subfigure}
\begin{subfigure}[b]{0.325\textwidth}
\includegraphics[scale=0.42]{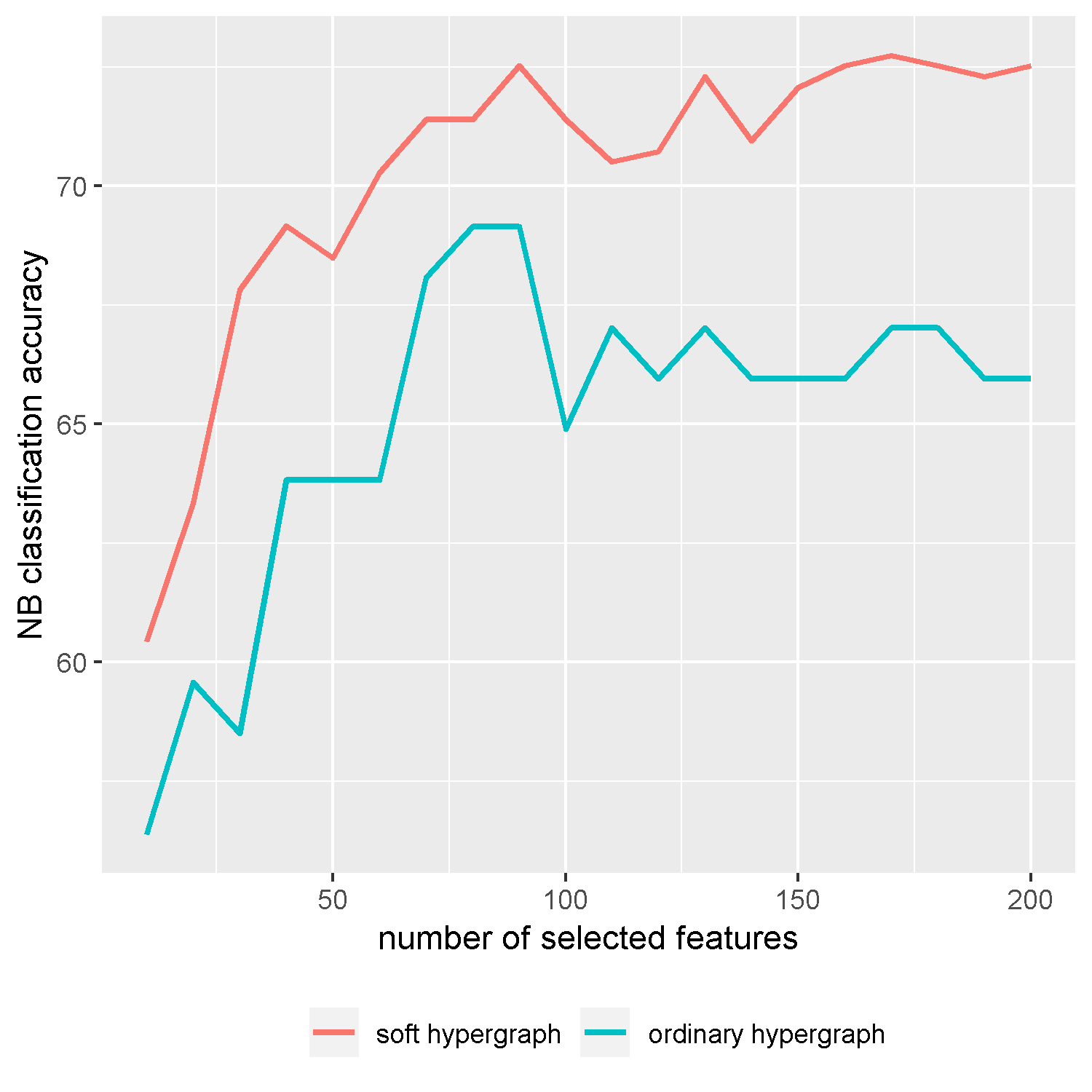}
\caption{Smoke-Cancer}
\end{subfigure}
\begin{subfigure}[b]{0.325\textwidth}
\includegraphics[scale=0.42]{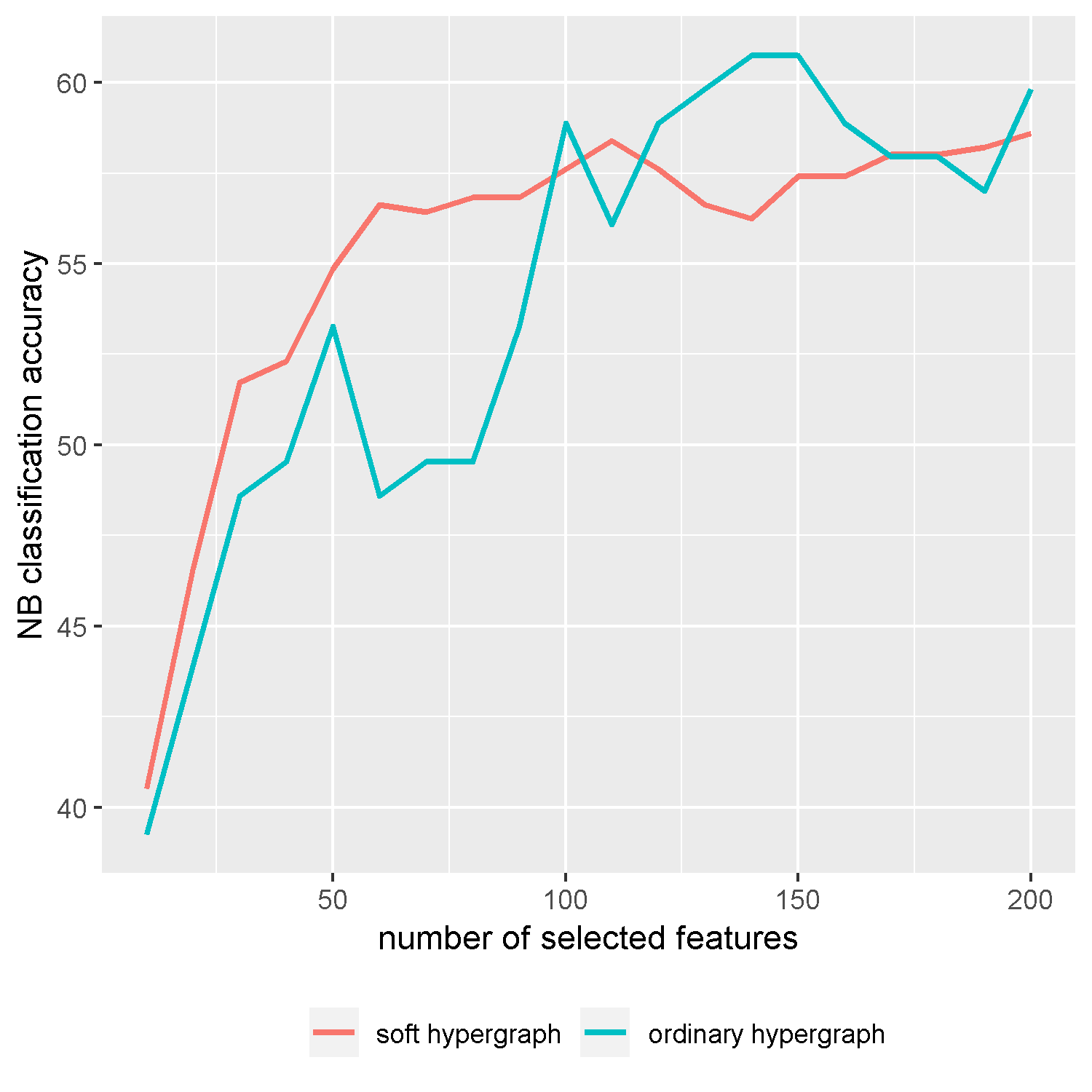}
\caption{Mouse-Type}
\end{subfigure}
\caption{NB classification accuracies of different methods on benchmark data sets.}
\label{figure:ordinary-hypergraph-nb}
\end{figure}

In soft hypergraphs, different vertices can have different participation in a hyperedge. This participation can be based on their importance and role in that hyperedge. This helps the framework to model data structure more precisely. We carried out some experiments to show the superiority of soft hypergraphs to ordinary hypergraphs. To do so, we converted the soft hypergraph of our framework to an ordinary hypergraph and tested it on some benchmark data sets using diverse evaluation methods. As it can be seen in figures \ref{figure:ordinary-hypergraph-knn}, \ref{figure:ordinary-hypergraph-lsvm}, \ref{figure:ordinary-hypergraph-rbsvm}, and \ref{figure:ordinary-hypergraph-nb}, using soft hypergraph results more classification accuracy in most cases.

\begin{figure}[!htbp]
\centering
\begin{subfigure}[b]{0.325\textwidth}
\includegraphics[scale=0.42]{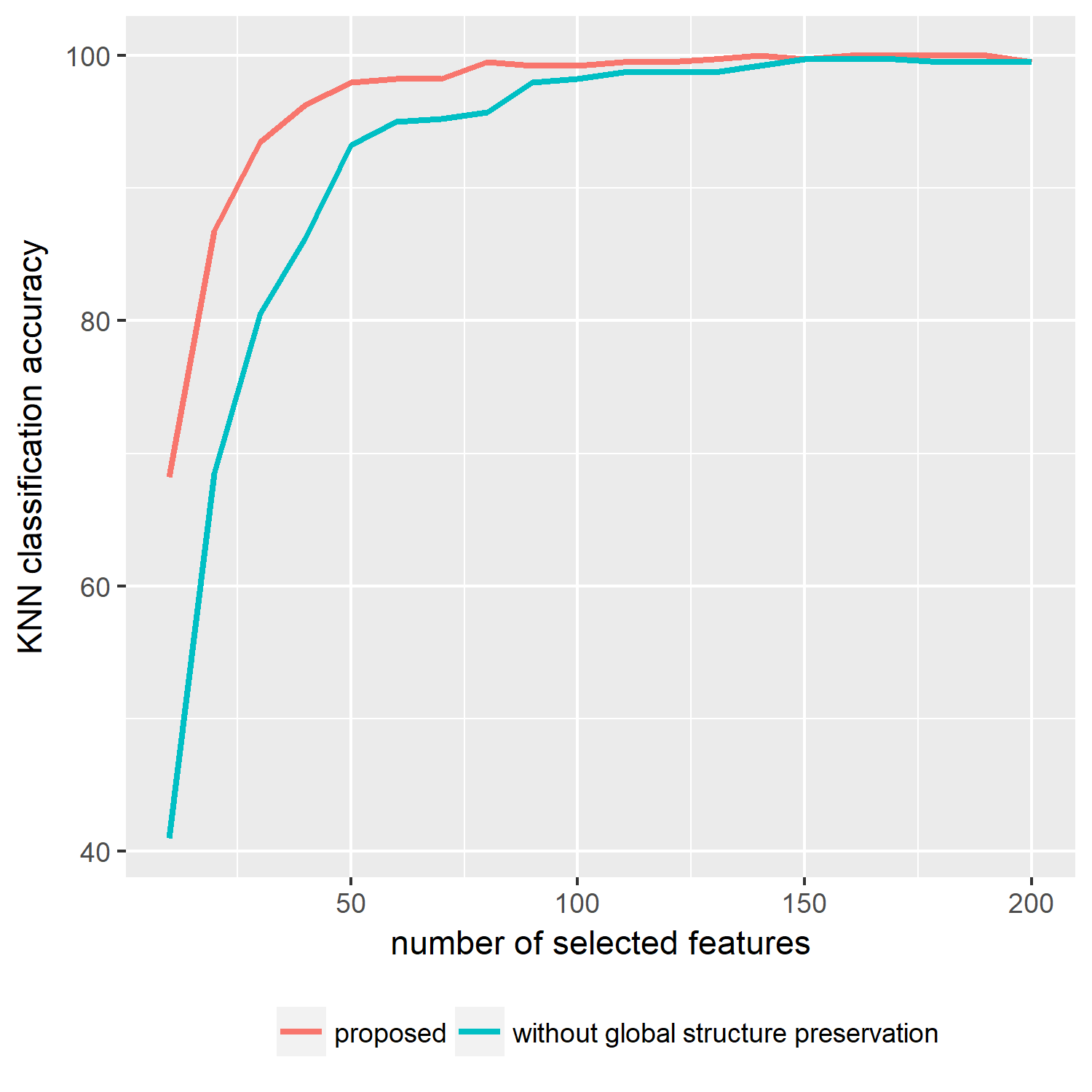}
\caption{Gene-Expression}
\end{subfigure}
\begin{subfigure}[b]{0.325\textwidth}
\includegraphics[scale=0.42]{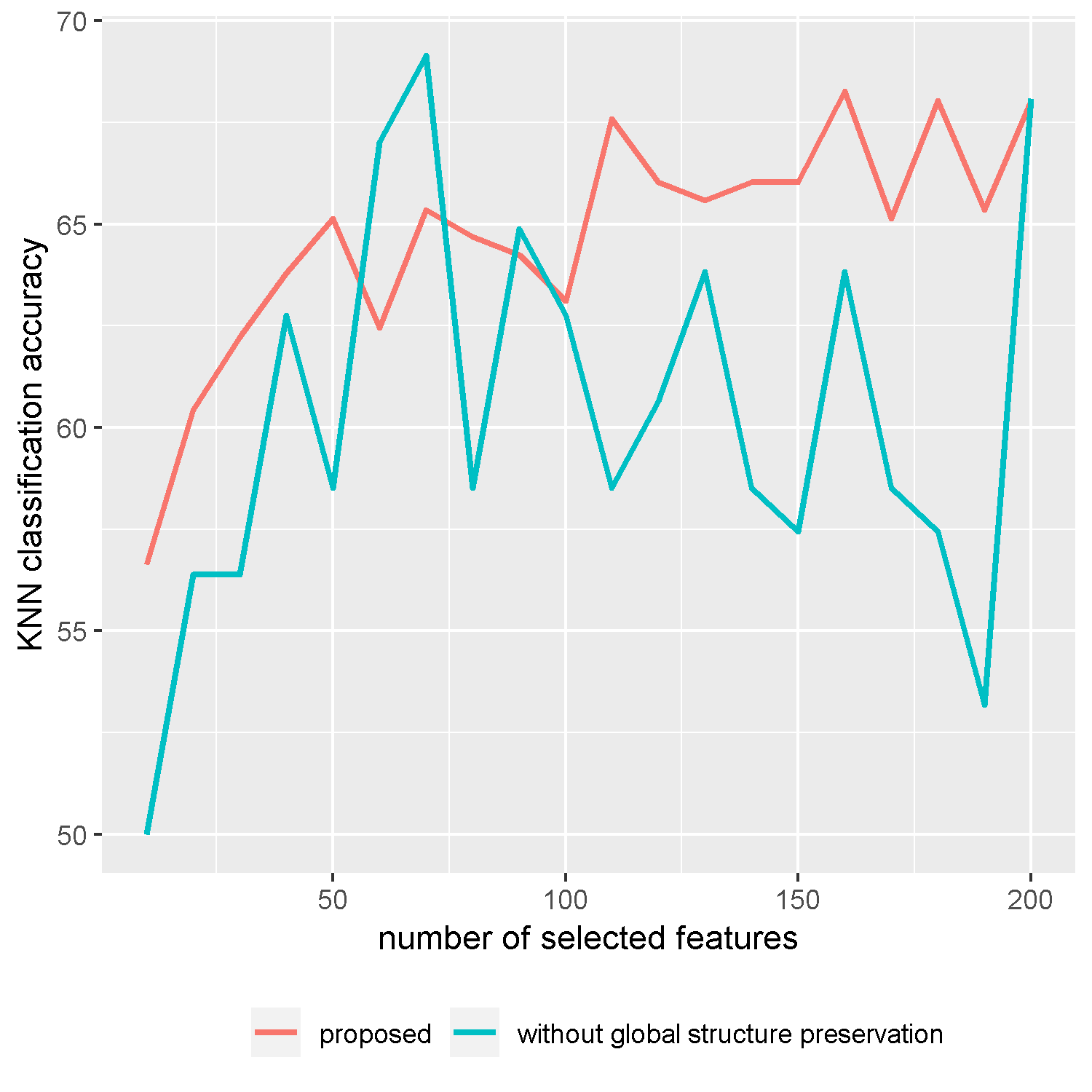}
\caption{Smoke-Cancer}
\end{subfigure}
\begin{subfigure}[b]{0.325\textwidth}
\includegraphics[scale=0.42]{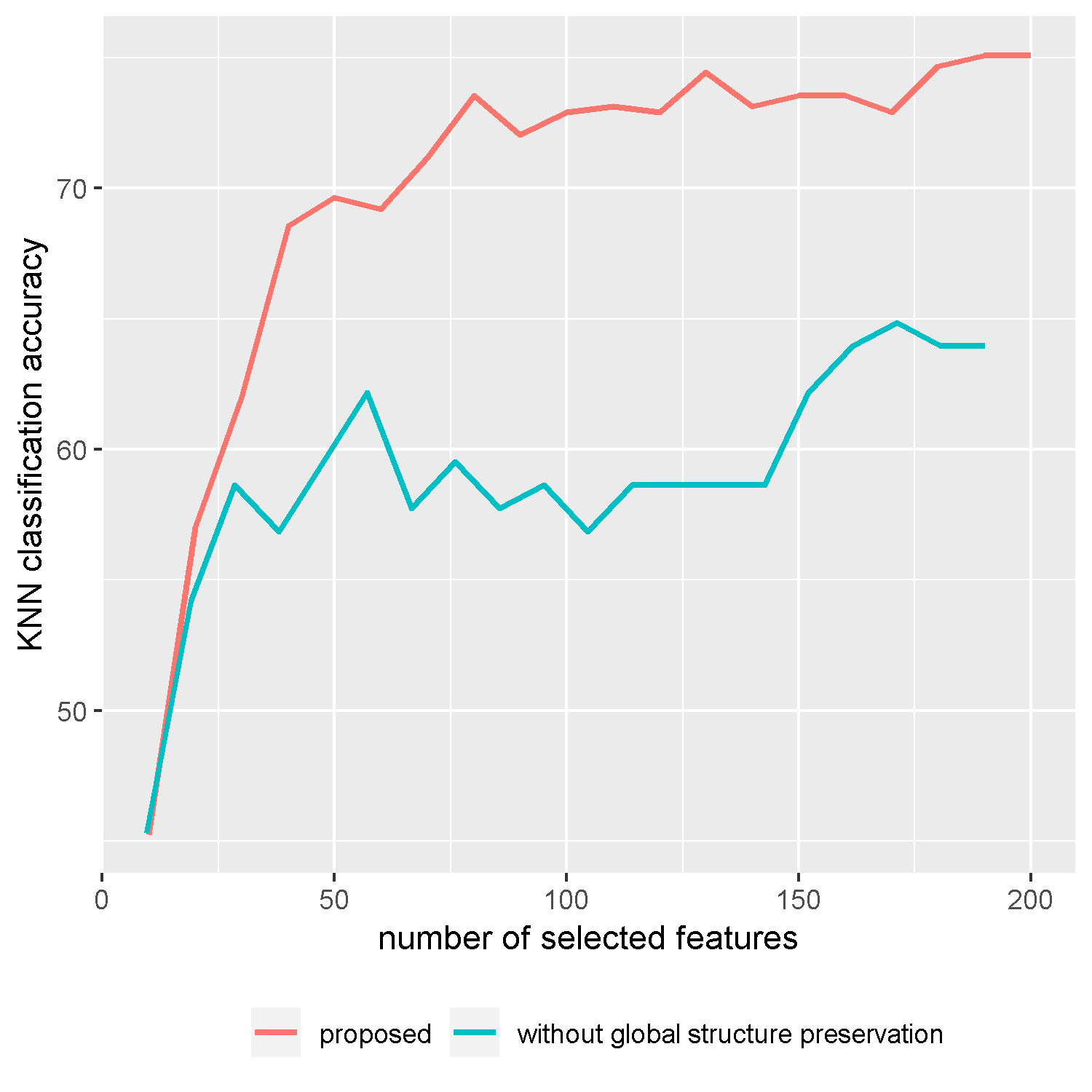}
\caption{Mouse-Type}
\end{subfigure}
\caption{KNN classification accuracies of different methods on benchmark data sets.}
\label{figure:global-str-knn}
\end{figure}

\subsection{Effect of global structure preservation}\label{Effect of global similarity preservation}
\begin{figure}[!htbp]
\centering
\begin{subfigure}[b]{0.325\textwidth}
\includegraphics[scale=0.42]{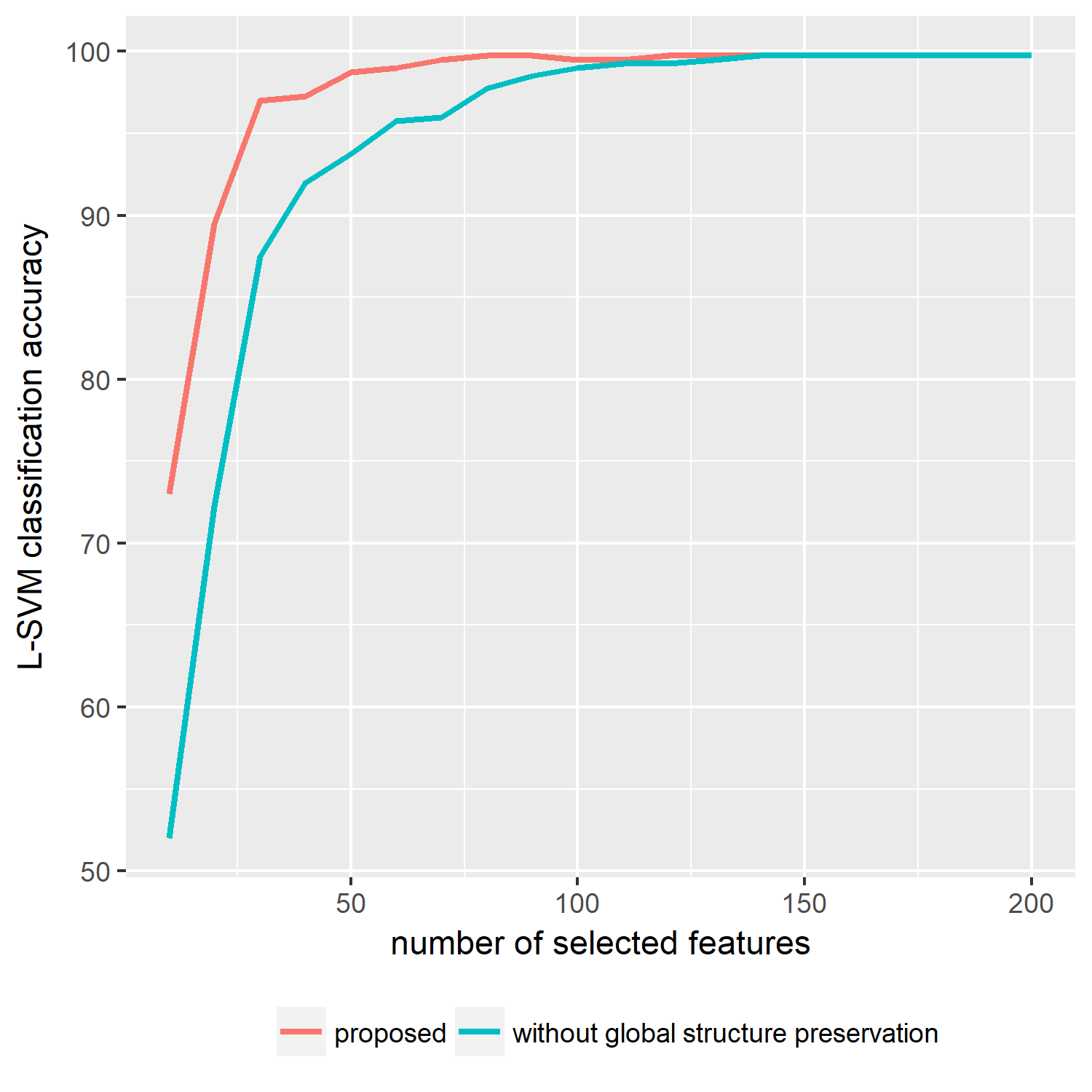}
\caption{Gene-Expression}
\end{subfigure}
\begin{subfigure}[b]{0.325\textwidth}
\includegraphics[scale=0.42]{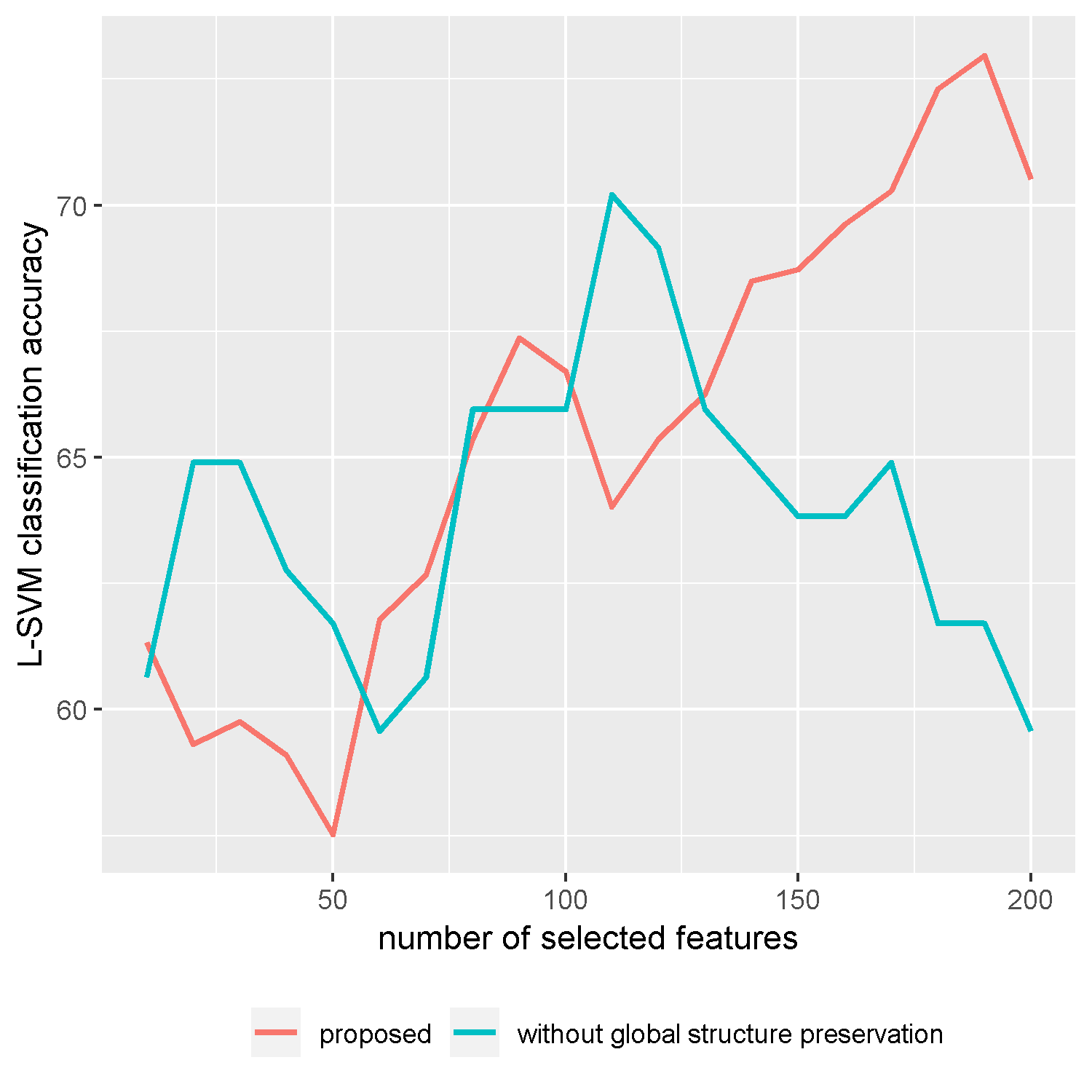}
\caption{Smoke-Cancer}
\end{subfigure}
\begin{subfigure}[b]{0.325\textwidth}
\includegraphics[scale=0.42]{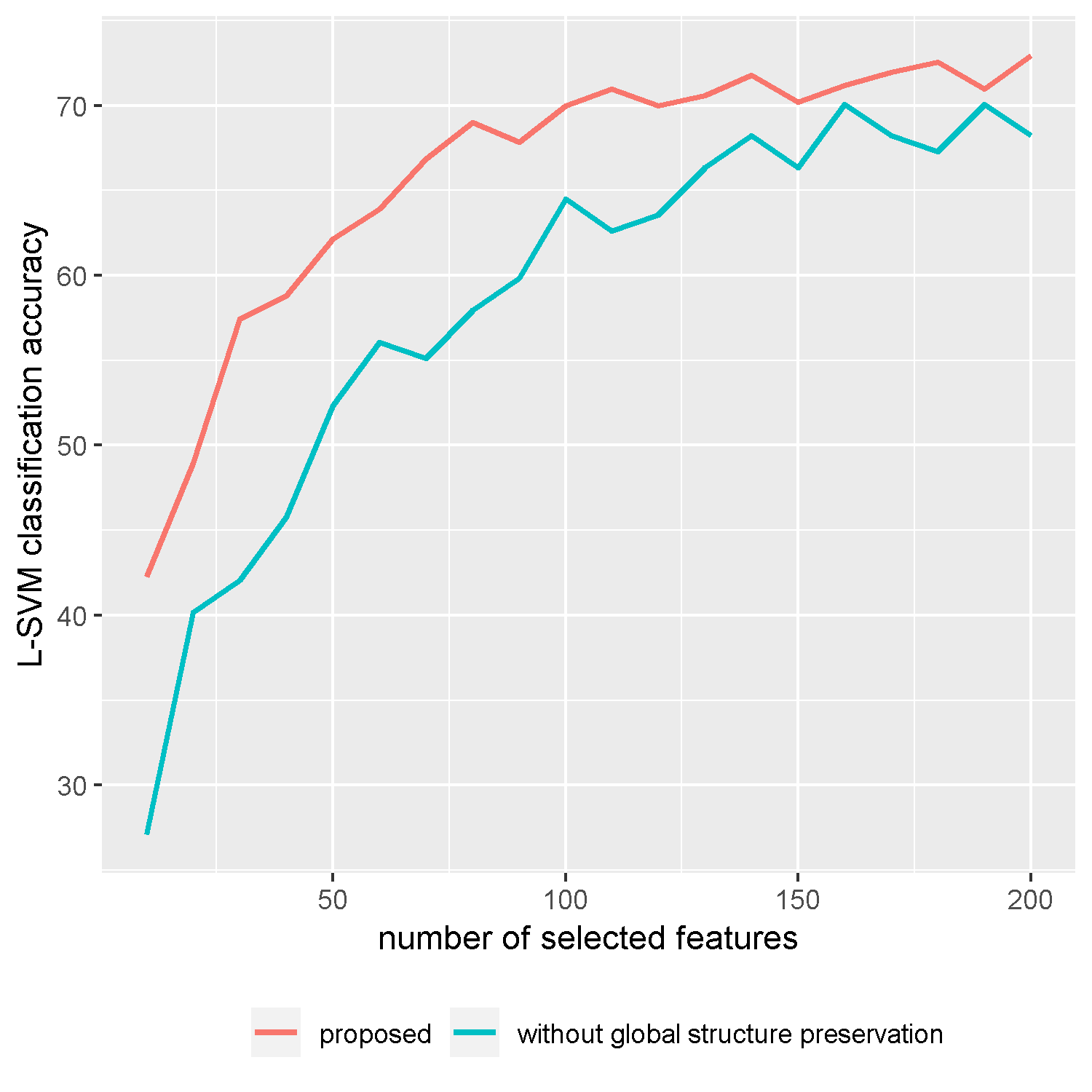}
\caption{Mouse-Type}
\end{subfigure}
\caption{L-SVM classification accuracies of different methods on benchmark data sets.}
\label{figure:global-str-lsvm}
\end{figure}

\begin{figure}[!htbp]
\centering
\begin{subfigure}[b]{0.325\textwidth}
\includegraphics[scale=0.42]{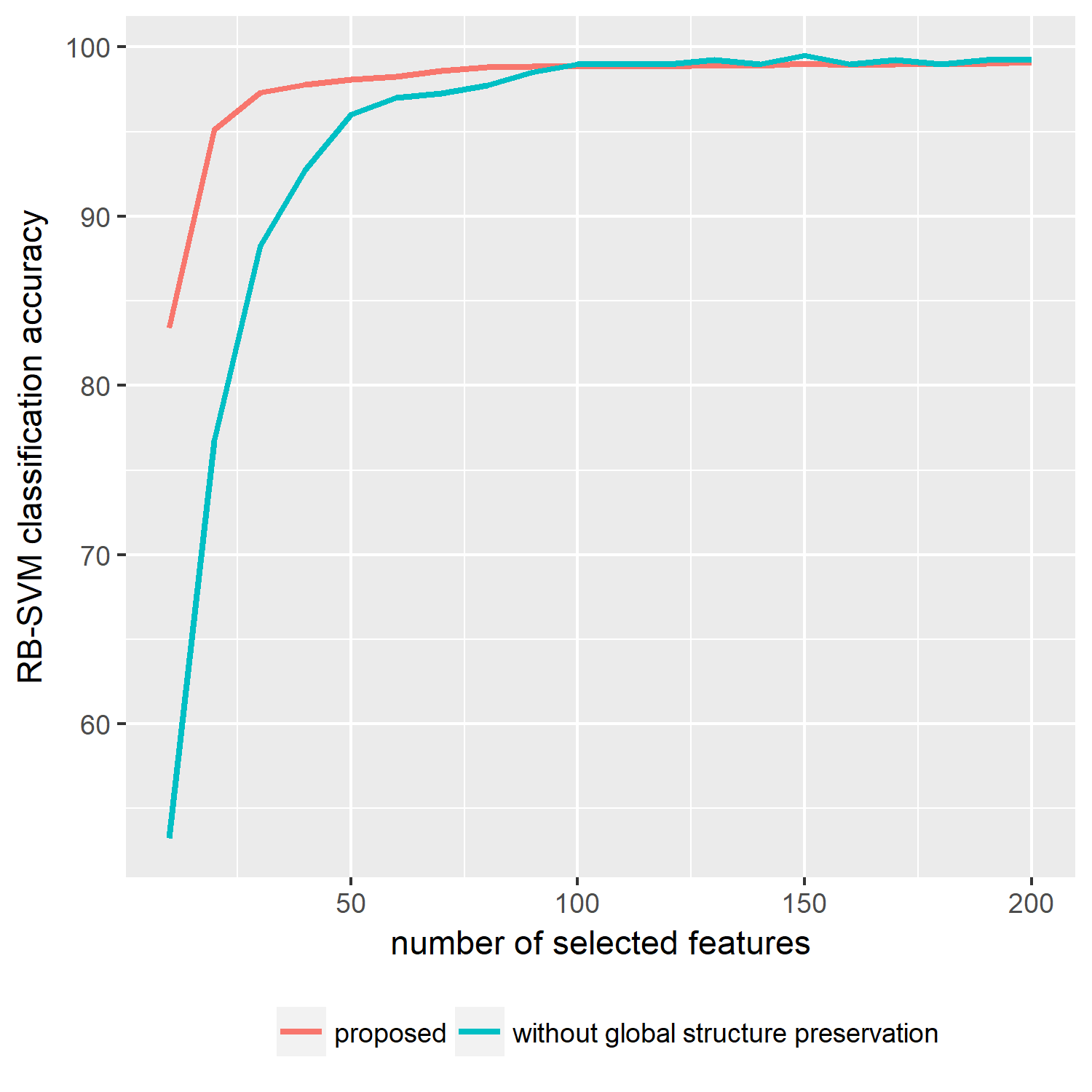}
\caption{Gene-Expression}
\end{subfigure}
\begin{subfigure}[b]{0.325\textwidth}
\includegraphics[scale=0.42]{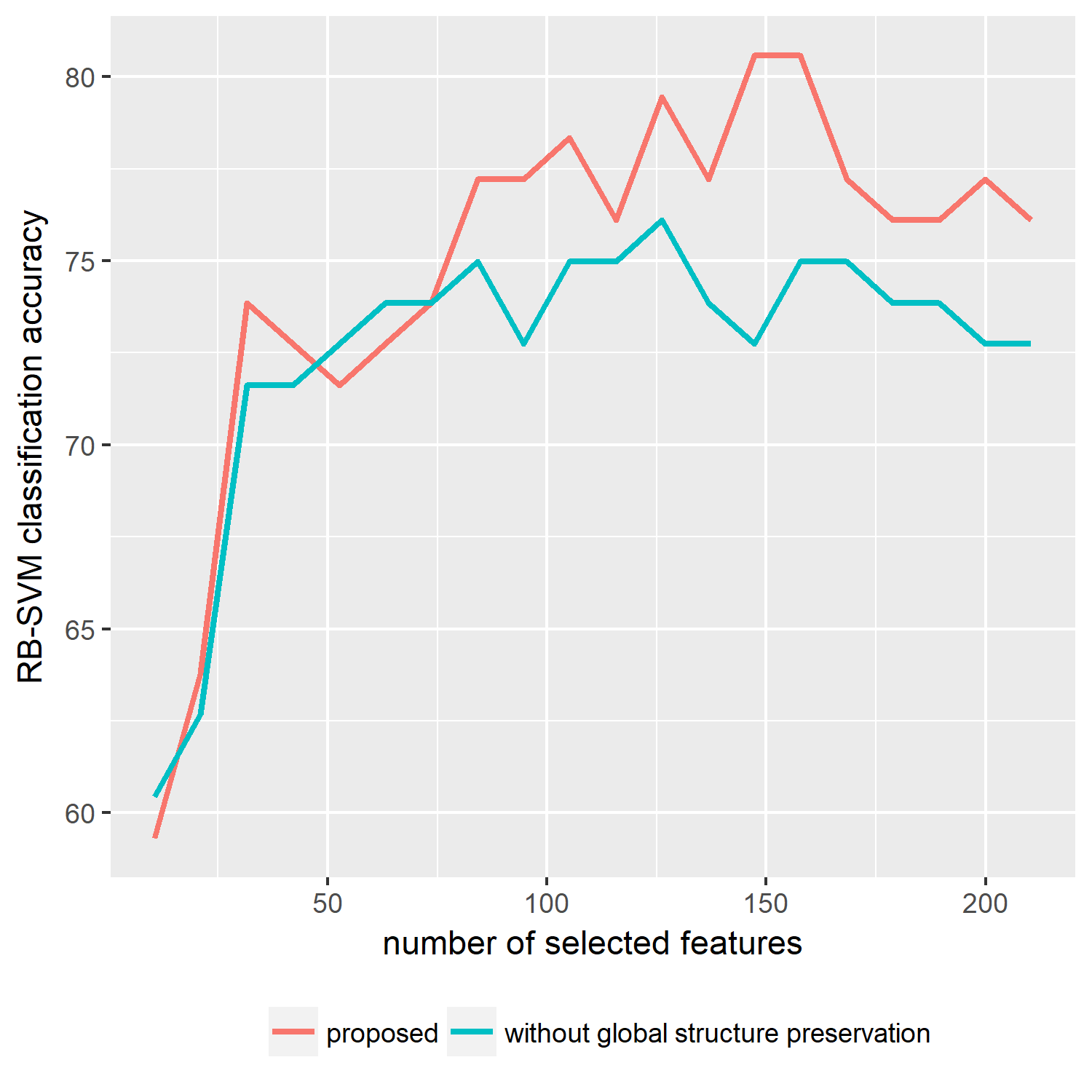}
\caption{Smoke-Cancer}
\end{subfigure}
\begin{subfigure}[b]{0.325\textwidth}
\includegraphics[scale=0.42]{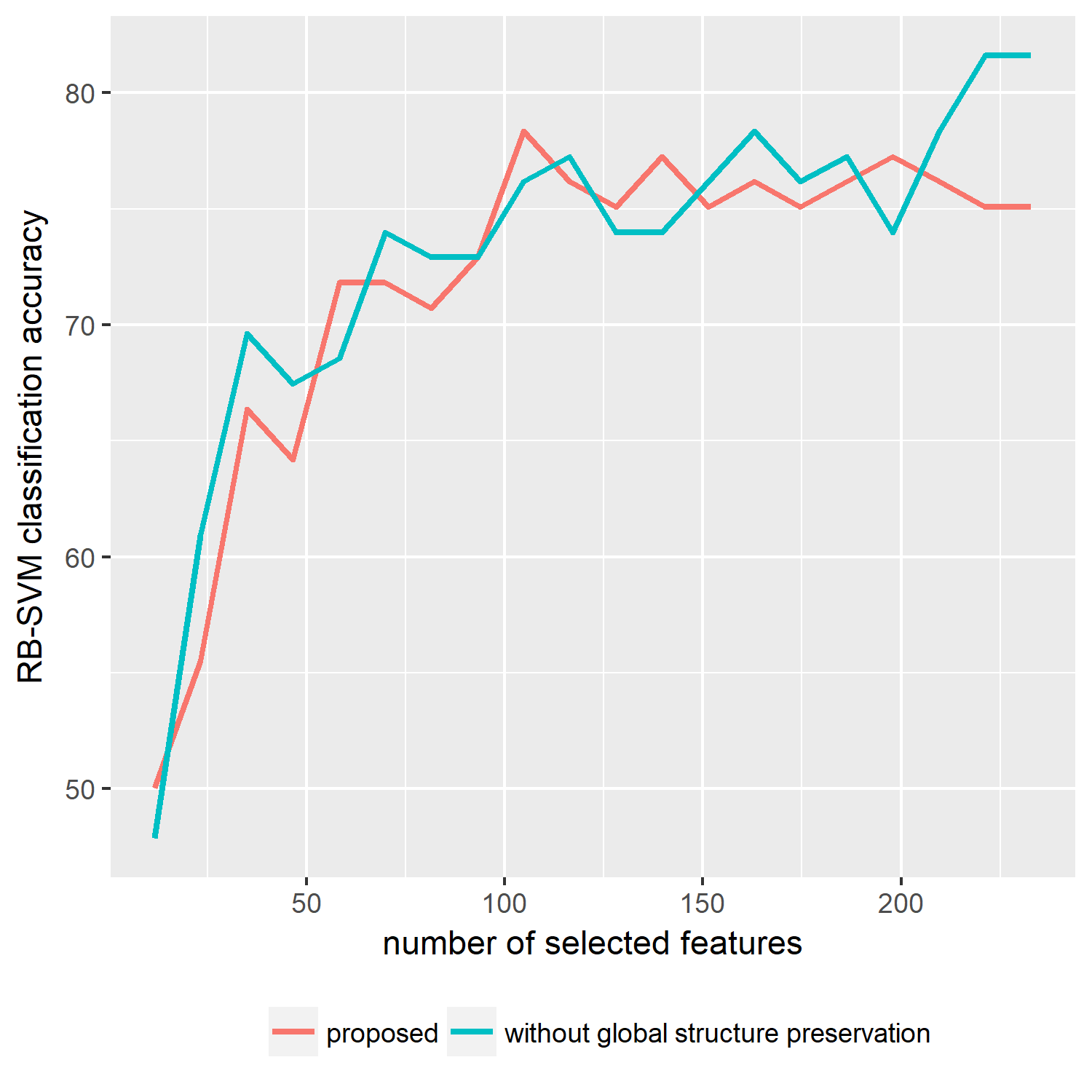}
\caption{Mouse-Type}
\end{subfigure}
\caption{RB-SVM classification accuracies of different methods on benchmark data sets.}
\label{figure:global-str-rbsvm}
\end{figure}

\begin{figure}[!htbp]
\centering
\begin{subfigure}[b]{0.325\textwidth}
\includegraphics[scale=0.42]{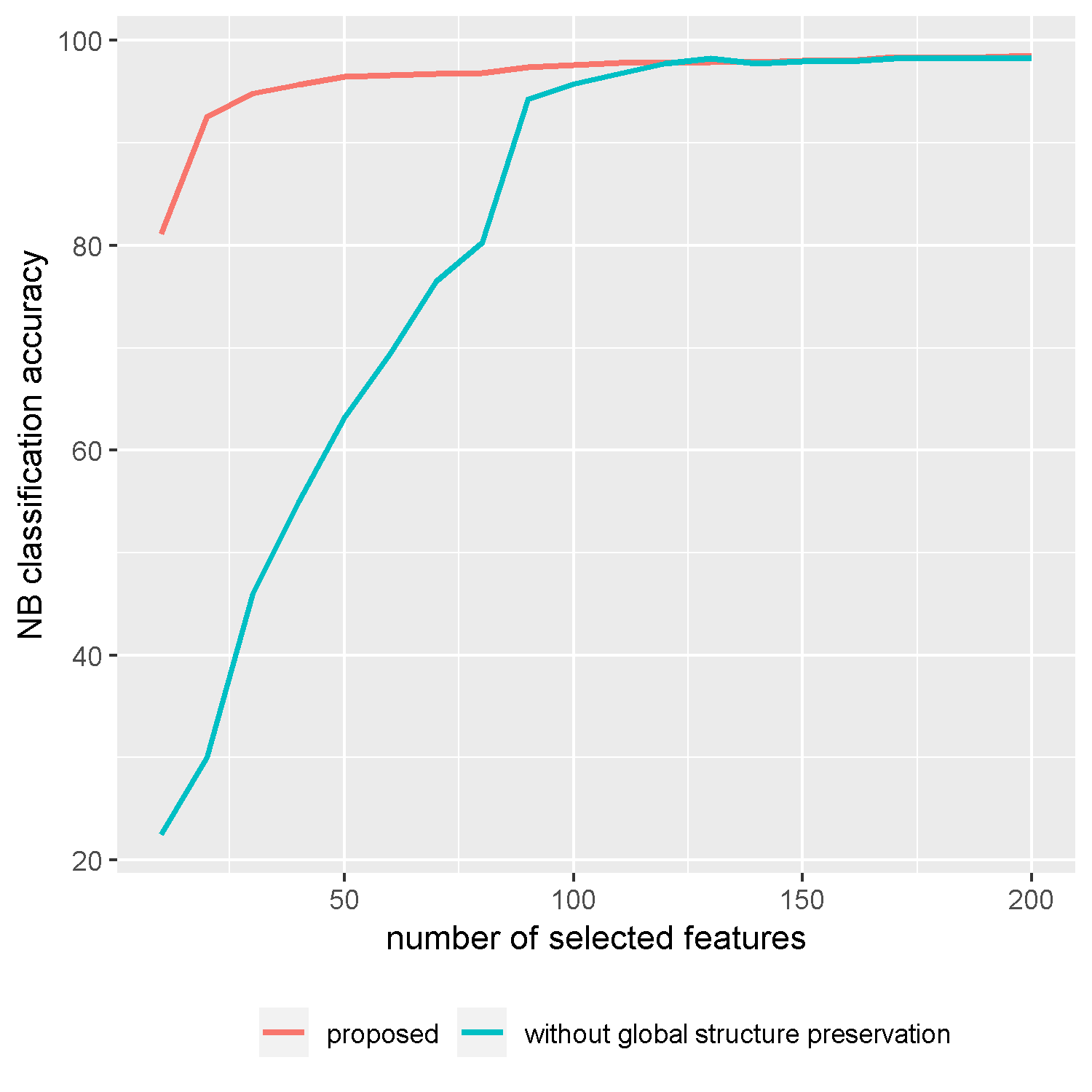}
\caption{Gene-Expression}
\end{subfigure}
\begin{subfigure}[b]{0.325\textwidth}
\includegraphics[scale=0.42]{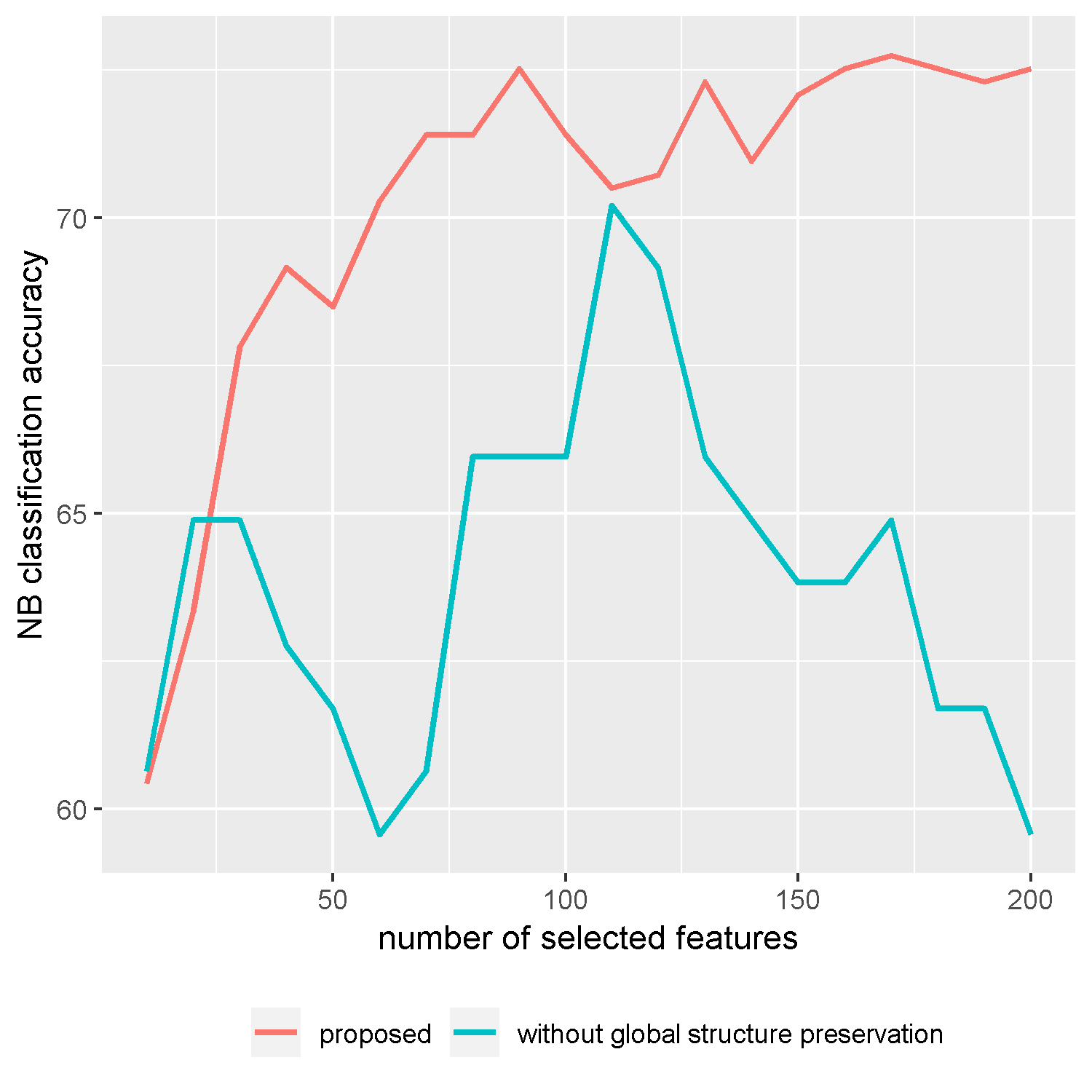}
\caption{Smoke-Cancer}
\end{subfigure}
\begin{subfigure}[b]{0.325\textwidth}
\includegraphics[scale=0.42]{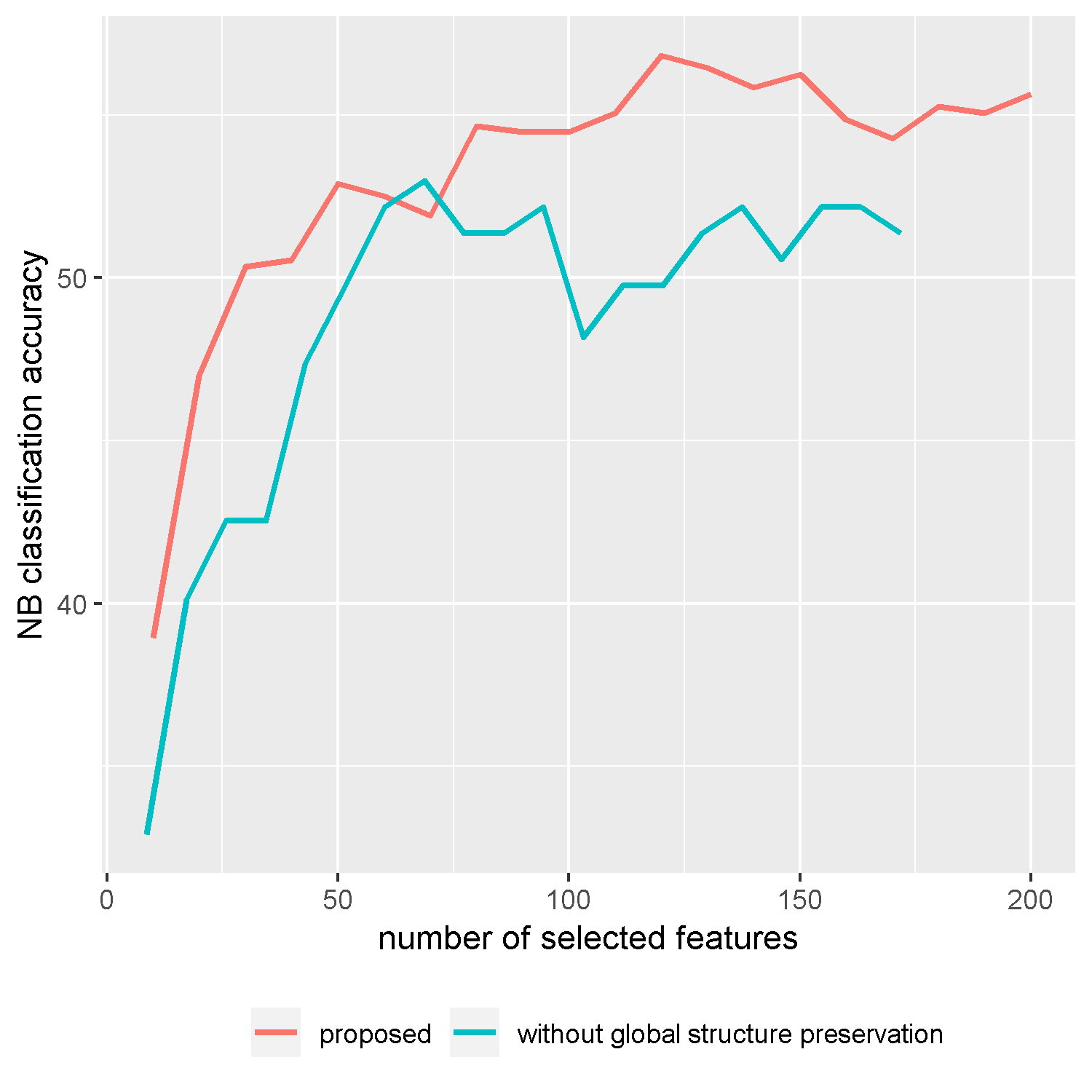}
\caption{Mouse-Type}
\end{subfigure}
\caption{NB classification accuracies of different methods on benchmark data sets.}
\label{figure:global-str-nb}
\end{figure}

In the proposed framework, we preserve the global structure of data by maintaining the correlation between data points. Since in unsupervised frameworks data labels are unavailable and data points from the same class generally have high linear dependency and correlation \cite{RFRSR}, this approach performs as an alternative to employment of data labels in supervised learning for structure preservation; Moreover, it helps to repair the inevitable loss of information caused by our low computational hypergraph which is constructed only based on cluster centroids. In order to test the importance of this term, we performed some experiments on different datasets using various evaluation methods. Although adding global structure preservation term has minor effect in computational cost of the proposed algorithm, as it can be seen in figures \ref{figure:global-str-knn}, \ref{figure:global-str-lsvm}, \ref{figure:global-str-rbsvm}, and \ref{figure:global-str-nb}, it leads to much more effective selection of features in most cases.

\section{Conclusion}\label{conclusion}
In this paper, a novel point-weighting framework for hypergraph feature selection was proposed which adopts low-rank representation to handle noise and outlier, and captures the importance of different data points. Focusing on centroids helps the proposed method to have lower computational complexity with respect to many state-of-the-art feature selection methods, in most cases. We also behave different data points based on their representation power and role in defining the data structure. We conducted experiments to show the effectiveness of the innovative ideas of this paper. Exhaustive experiments exhibit the effectiveness of our proposed method in comparison with other state-of-the-art feature selection methods.

\clearpage
\bibliographystyle{elsarticle-num}
\bibliography{ref}
\end{document}